# Sentences as connection paths:
# A neural language architecture of sentence structure in the brain

*Frank van der Velde*

University of Twente, Cognition, Data and Education, BMS
Drienerlolaan 5, 7522NB Enschede, The Netherlands
f.vandervelde@utwente.nl; veldefvander@outlook.com

**Abstract**

This article presents a neural language architecture of sentence structure in the brain, in which sentences are temporal connection paths that interconnect neural structures underlying their words. Words remain 'in-situ', hence they are always content-addressable. Arbitrary and novel sentences (with novel words) can be created with 'neural blackboards' for words and sentences. Hence, the unlimited productivity of natural language can be achieved with a 'fixed' small world like network structure. The article focuses on the neural blackboard for sentences. The architecture uses only one 'connection matrix' for binding all structural relations between words in sentences. Its ability to represent arbitrary (English) sentences is discussed in detail, based on a comprehensive analysis of them. The architecture simulates intra-cranial brain activity observed during sentence processing and fMRI observations related to sentence complexity and ambiguity. The simulations indicate that the observed effects relate to global control over the architecture, not to the sentence structures involved, which predicts higher activity differences related to complexity and ambiguity with higher comprehension capacity. Other aspects discussed are the 'intrinsic' sentence structures provided by connection paths and their relation to scope and inflection, the use of a dependency parser for control of binding, long-distance dependencies and gaps, question answering, ambiguity resolution based on backward processing without explicit backtracking, garden paths, and performance difficulties related to embeddings.

## 1. Introduction

Language is a key element of human life, culture and history, giving an unlimited ability to communicate on a wide variety of topics. We can construct and understand a virtually unlimited set of sentences, based on a vast but limited set of words.

We can understand novel sentences based on familiar words. At least we can recognize what is said about what, even when novel words are used (Feldman, 2013). A general view (e.g., Jackendoff, 2002) on language use is that words are stored in memory, but sentences are not (except idioms, e.g., *kick the bucket*). Instead, sentence structures are constructed during sentence comprehension. A central issue in the study of human cognition (e.g., Newell, 1990) and the brain (e.g., Pylkkänen, 2019) is how this can be achieved.

It is informative to distinguish between the abilities of constructing sentence structures and combining them with words, because each imposes specific demands on the underlying architecture. For example, digital-like architectures could represent *Sue likes pizza* by using memory registers for subject, verb and object, filled with (neural) codes related to the words (e.g., Kriete et al., 2013). But any noun or (transitive) verb could be used here. Hence, this approach basically requires the ability to store arbitrary codes in arbitrary registers.

This can be seen as a 'logistics' problem, because it concerns the unlimited access needed for combinatorial processing. In digital computing, it can be solved by connecting all memory locations to a data-bus. The digital codes or 'symbols' representing words can then be copied from their memory locations and transported to the memory locations used to process a sentence (e.g., Fodor & Pylyshyn, 1988; Newell, 1990).

Although aspects of digital computing such as neural word or address codes have been used to model basic aspects of neural language processing (Kriete et al., 2013; Eliasmith, 2015; Müller et al., 2020), a neural language model of this kind would require a neural architecture for unlimited access as given by a data-bus to account for combinatorial language processing.

Here, the Head-Dependent Neural Blackboard Architecture (HD-NBA[1]) is presented and simulated to account for the ability to compose arbitrary sentence structures without relying on aspects of digital computing. That is, the architecture does not rely on 'neural codes' to represent words, relations, or memory locations. Furthermore, it specifically does not rely on a data-bus like architecture for unlimited access. Instead, it posits that sentence structures are temporal connection paths between neural word structures that remain 'in-situ'.

Connection paths can be constructed for all phrase and sentence structures in English, as given in a comprehensive overview of them (Huddleston and Pullum, 2002), referred to here as *Cambridge Grammar.* Simulations of the architecture, e.g., aim to account for a number of observations related to brain activity observed with human language processing and performance.

## 2. Head-Dependent Neural Blackboard Architecture

The HD-NBA posits that sentence structures in the brain are connection paths (James, 1890; Shanahan, 2010) in a network structure that interconnects the neural structures underlying the words involved. The connection paths are temporal, and activated in line with the sentence structure. Arbitrary and novel sentences can be constructed in this way, because the network structure (or 'workspace', Baars, 1988; Wiggins, 2012) that interconnects neural word structures resembles a 'small-world' network (Watts & Strogatz, 1998; Shanahan, 2010). The neural word structures themselves remain 'in-situ' during this process. Hence, words are content-addressable, also when they are part of sentence structures.

The in-situ neural word structures in the architecture are based on the 'hub-and-spoke' model of semantic representation (Lambon-Ralph et al., 2017). Spokes are modality specific, and distributed over and within different cortical areas (Huth et al., 2016). They are interconnected via transmodal hubs in the anterior temporal lobes (Lambon-Ralph et al., 2017).

Figure 1 illustrates this for *pizza* and *likes*. Both word structures would consist of neural representations in modal specific spokes and amodal hubs. The spokes would relate to how we perceive and produce the words, or perceive, use and validate the object pizza, or engage in the act of liking.

Hub representations play an important role in integrating spoke representations, as those of *pizza*. But they do not code or stand in for them. The meaning of *pizza* is not given by its hub representation only, but by the entire hub-and-spoke combination. For example, the hub representation alone could not account for the fact that we can hear the word *pizza* or speak it in response to seeing the object. For words, these abilities are encoded as connection paths in the sensory-motor loop, provided by the overall hub-and-spoke neural word structure.

[1] All abbreviations used in the article are listed in section S11 in the Supplementary Information,

In the HD-NBA, connection paths of words are connected to the sentence structure. So, the word *pizza* in *Sue likes pizza* can be heard or spoken via its hub-and-spoke word structure. This does not rule out effects of sentence contexts on these processes, or effects of semantic interactions between words (e.g., Pylkkänen, 2019).

Also, words in questions, e.g., *Sue* and *like* in *what does Sue like?*, directly activate their parts in sentence structures, e.g., *Sue likes pizza*, because in-situ representations are part of questions and sentences simultaneously (see section[2] S7.6).

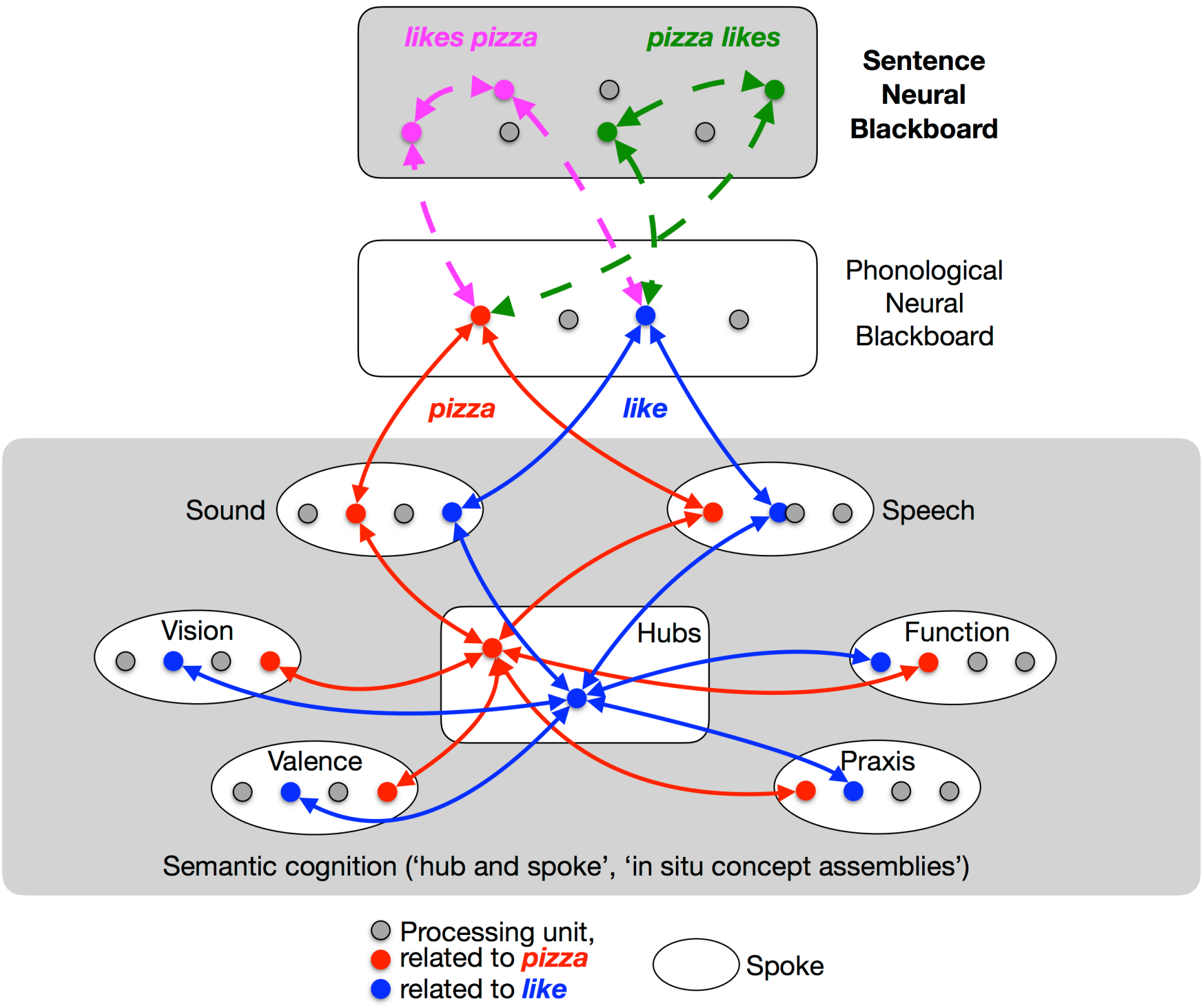

**Figure 1. Sentences as connection paths.** Bottom: in-situ word structures of *likes* and *pizza* in a hub-and-spoke model of semantic representation (based on Lambon-Ralph et al., 2017). Top: the phrases *likes pizza* and *pizza likes* in neural blackboards for phonological and sentence structures.

The HD-NBA aims to combine a 'phonological neural blackboard' with a 'sentence neural blackboard', in line with the two-tier combinatorial nature of language (Jackendoff, 2002). This combination would provide the small-world like network structure needed for the unlimited productivity of language and allows the use of novel words (S4.3). At present, the HD-NBA addresses the construction and processing of arbitrary sentence structures in the sentence blackboard. It is assumed that words are first encoded in a phonological blackboard, interconnected with the sentence neural blackboard.

Figure 1 illustrates that the sentence neural blackboard is needed to build a temporal connection path that reflects the structure of a phrase or sentence. In *likes pizza*, this connection path entails that *pizza* is the object of *likes*, whereas it is the subject in the path of *pizza likes*. The HD-NBA allows both possibilities to be formed and stored simultaneously. This reflects the ability to handle arbitrary sentence and phrase structures, with multiple occurrences of the same word in different sentences (S7.6).

[2] References starting with 'S' refer to the Supplementary Information, attached to this article.

A main demand on the structure of the sentence neural blackboard derives from the 'logistics' problem addressed above, i.e. the ability to combine arbitrary words in arbitrary sentence structures (referred to here as 'combinatorial productivity'). Figure 1 illustrates this ability with the phrase *pizza likes*. Clearly, this is not a state of affairs found in the real world. But we have no difficulty in understanding the expressed relation, or recognizing it in a cartoon movie. Indeed, we know that it is odd precisely because we understand the different relations between *pizza* and *likes* in *pizza likes* versus *likes pizza*. This ability lies at the heart of human creativity, as expressed in literature, movies, cartoons or games, which all critically depend on virtually unlimited combinatorial productivity.

Combinatorial productivity imposes demands on the ability to form connection paths in the sentence neural blackboard. For example, connection paths must be possible between all in-situ structures of nouns and all in-situ structures of verbs, to allow any noun to be any of the arguments of any verb (subject, object or recipient, depending on the verb). This includes 'weird' combinations such as *pizza likes*.

Another demand derives from the ability to process and represent sentence structures at a rate of 3 to 4 words per second, as found in human performance (Rayner & Clifton, 2009). This would preclude the development of new connections to combine words that have not been combined before. Hence, the unlimited productivity of natural language would depend on a 'fixed' connection structure (S2). This could develop over time, but growing new connections that would interconnect novel combinations of words during processing seems unfeasible.

In short, the sentence neural blackboard needs to provide the means to construct connection paths for arbitrary sentence structures, based on any set of words, and in the time of sentence processing as found in human performance. The connection paths link the in-situ structures of the words in a sentence, so that each of them is directly accessible. In this way, the meaning of a sentence can be composed by the meanings of the words and their relations in the sentence. So, *Sue* can be identified as the agent of *likes pizza*. The connection paths are temporal, so that new sentences can be processed all the time (but they can be transferred to long-term memory, van der Velde and de Kamps, 2006).

The role of in-situ word structures in connection paths can also be illustrated with the ability to answer questions as fast as possible. A closed question in particular contains words that are a part of the queried sentence structure, as well as cues about the role of the requested information in the sentence. For example, the question *what does Sue like?* presupposes that *Sue* and *likes* are part of the sentence queried and that the requested information has the role of object, as in *Sue likes X* or *John believes Sue likes X*.

From an evolutionary perspective, this close relation between questions and queried sentence structures could be more than a coincidence. It could reflect the fact that this relation provides the best or fastest way to retrieve the requested information, also given the time pressure on retrieving information in potentially dangerous environments.

In the HD-NBA it is clear why questions are structured like this. The in-situ word representations *Sue* and *likes* are the same in the question and the queried sentence structure, and directly accessible in both. Hence, their activation by the question results in the direct activation of the parts of the sentence structures to which they belong. In turn, the reactivated part of the sentence structure can be used to answer the question, also because the type of the requested of information (e.g., *what*) is typically given by the question (see S7.6).

In the course of evolution, this direct approach would have had benefits over more indirect and hence slower and more error-prone approaches, as given by a required de-encoding of the sentence structure to retrieve its constituents, or by protracted searches of symbols in memory. In turn, these benefits would have shaped the neural architecture of language representation and processing.

The current development of the HD-NBA is based on a numbers of core assumptions (S2.1) and choices (S2.2), in particular motivated by an extensive survey of sentence structures in *Cambridge Grammar* (S3.1). A main choice is the use of a single 'connection matrix' for binding words in sentence structures, as illustrated in the next section.

## 3. Sentence structures as connection paths

Figure 2 illustrates a connection path in the sentence neural blackboard of the HD-NBA for the sentence *Sue likes pizza*.

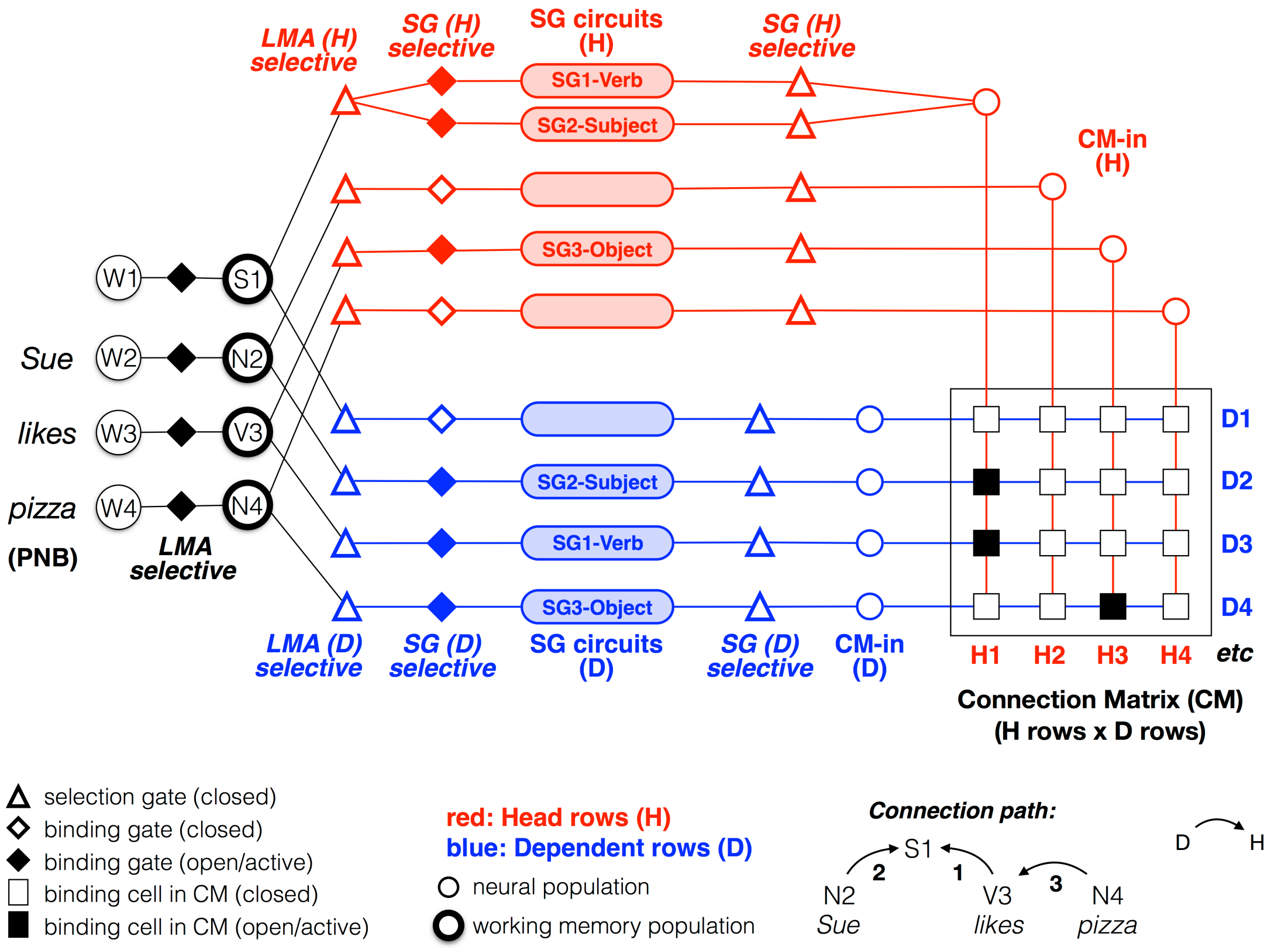

**Figure 2. Connection structure of the HD-NBA**. Illustration of the connection path of the sentence *Sue likes pizza*. The head part of a connection path (top) is depicted in red, the dependent part (bottom) in blue. D = dependent; H = head; LMA = lexical main assembly, N = noun, PNB = phonological neural blackboard, S = sentence, SG = structure group; V = verb; W = word.

Every local node of the blackboard is seen and modeled as a small set (or 'population') of locally interacting excitatory and inhibitory neurons (Wilson & Cowan, 1972; 2021) that operates as a unity (S7). There are two kinds of these. The first are just neural populations. They can be active for as long as they receive input activation. The second are 'working memory populations' (S4.1). Once activated, they sustain their activity for a while, in line with cortical working memory activity (e.g., Amit, 1989; Bastos et al., 2018). They form the basis of a connection path in the sentence neural blackboard and play a crucial role in sentence processing (S6).

The neural sentence blackboard is organized in terms of 'rows' (S4.2), and the number of rows determines its capacity (S7.3). Each row starts with a population referred to as a 'word' population (W). These are assumed to interconnect the phonological and sentence neural blackboards. They are numbered consecutively in Figure 2. During processing, only one word population is active at a time, with the assumption that a new consecutive row is used for each new word or phrase.

A connection path in the HD-NBA is based on the types of words and phrases involved, the roles they have in their relations and the types of these relations. In the HD-NBA, word and phrase types are represented with working memory populations, referred to as 'lexical main assemblies' or LMAs, listed in Table 1. All LMA types are represented in each row of the sentence blackboard (S4.2).

| Lexical word categories (CG p. 22 [4]) | Lexical main assemblies (HD-NBA) |
|---|---|
| noun | N |
| verb | V |
| adjective (e.g., *good, nice, big, easy*) | Adj |
| adverb (e.g., *easily, soon, so, too*) | Adv |
| preposition (e.g., *of, to, by, between, since*) | P |
| determinative (e.g., *the, this, some, all, each*) | Det |
| subordinator (e.g., *that, for, to, whether, if*) | C (clause) |
| coordinator (e.g., *and, or, but, not*) | Co |
| interjection (e.g., *ah, gosh, damn, wow*) | - |
| | Pr (pronoun) |
| | S (sentence) |

**Table 1**. Left: Lexical word categories from *Cambridge Grammar* (CG). Right: Lexical main assemblies (LMAs) in the HD-NBA.

A connection path as illustrated in Figure 2 is constructed by means of selective control of activation and 'binding'. Both operations depend on neural 'gating' circuits, referred to as 'selection gates' and 'binding gates' respectively (S4.1). They are based on cortical dis-inhibition circuits (e.g., Letzkus et al. 2015). A selection gate is opened by an external selection signal. When active, activation can flow through the selection gate (Figure S4A). A binding gate is opened initially by an external binding signal. After opening it remains active for a while due to a working memory population in the circuit (Figure S5A). While it is active, activation can flow through the binding gate, which constitutes the binding of the neural populations it interconnects. Binding is temporal because the reverberating activity decays after a while (e.g., Amit, 1989) or it can be inhibited (S4.2). This ensures that the sentence neural blackboard can be used for new sentences continuously.

Processing in the HD-NBA is incremental, in line with human performance (Tanenhaus et al., 1995; Rayner & Clifton, 2009). It starts with the first word (or sentence LMA) and gradually builds a sentence structure (S6). A word is first processed and bound to a word population in the phonological neural blackboard. This forms the entry to a row in the sentence neural blackboard. Each activated word population binds to one LMA in its row, corresponding to the type of word involved. So, *Sue* (W2) binds to a noun LMA (N2), *likes* (W3) to a verb LMA (V3) and *pizza* (W4) to another noun LMA (N4). In Figure 2, W1 was also activated and bound to a sentence LMA (S1), even though it does not correspond with a word. Gradually, these LMAs bind to each other in line with the structure of the sentence, based on the relations between the words (or phrases) and the roles they play in these relations (S6).

To achieve combinatorial binding of LMAs (and thus of words and phrases), each row is split into a 'head' and 'dependent' row. Each head row is connected to each dependent row in a 'connection matrix'. Specific 'binding cells' in the connection matrix allow the binding of any head row with any dependent row (Figure S6). A single connection matrix is sufficient to bind words and phrases in all structural roles required, including syntactical aspects such as agreement and case (S3.2) and long-term dependencies (S3.3). This constitutes perhaps the most minimal form of binding possible (S2.2). The use of a single connection matrix for all structural relations does impose constraints on sentence processing and performance (S8).

A single connection matrix for binding requires that it is not linguistically selective. Instead, linguistic selections are made in the input to and output from the connection matrix (S4.2). After the selection of the LMA type, these selections are based on the type of relation involved in a binding between two LMAs and their roles in the binding. A survey of *Cambridge Grammar* indicates that there could be around 100 different types of relations between LMAs.

Table 2 presents an overview of them, with the columns representing LMAs as 'heads' and the rows representing LMAs as 'dependents' in a relation specific HD-NBA binding (Tables S1-S10). For example, noun and adjective LMAs can bind to each other with noun as head and adjective as dependent, or vice versa. So, adjective is head and noun dependent in *it was more worth the effort*, whereas noun is head and adjective dependent in *financial advisers* (Table S5).

| Dependent LMAs | Head LMAs | | | | | | | | |
|---|---|---|---|---|---|---|---|---|---|
| | Adj | Adv | C | Co | Det | N | P | Pr | S | V |
| Adj | | x | x | x | | x, x, x, x | x, x | | x | x |
| Adv | x | | x | x, x | x | x, x | x, x | x | x | x, x, x |
| C | x | x | x, x, x | x | | x, x | x | x | x, x, x, | x |
| Co | x | x | x | | | x | x | x | | x |
| Det | x | x | | | | x | | x | | |
| N | x, x | x | x, x | x | | x, x, x | x, x, x | x | x, x | x, x, x, x |
| P | x, x | x | x, x | x | | x, x, x | x | x | x, x | x, x, x |
| Pr | | | x, x | x | | x, x | x | | x, x | x, x, x, x |
| S | | | | | | | | | | |
| V | | | x, x | x | | | x | | x, x | x |

**Table 2**. Overview of relations (x) between LMAs, based on *Cambridge Grammar*. Head and dependent refer to binding in the HD-NBA.

Hence, a connection path of a sentence structure needs to distinguish between the role of an LMA as head or dependent in a specific binding relation. In each blackboard row, this distinction is made with selection gates for head and dependent for each type of LMA (Figure S7). In the relation *likes pizza*, *likes* becomes the head by activating the selection gate for verb-as-head and *pizza* becomes a dependent by activating the selection gate for noun-as-dependent (Figure 2).

However, even when the roles of head and dependent are the same for a pair of LMAs, different types of relations are possible. For example, Table 2 gives four different relations with noun as head and adjective as dependent. In these relations, the adjective is either a complement, as in *financial advisers* (Table S5), a modifier, as in *unnecessary changes* (Table S6), a post-positive, as in *a leader younger than you* (Table S8), or an external modifier, as in *it's such a pity* (Table S9).

In Figure 2, different types of relations are distinguished with selective neural circuits referred to as 'structure groups'. All types of relations in Table 2 are distributed over 10 structure groups. This distribution is a main choice made in the implementation of the model (S2.2). The structure groups are presented in Tables S1-S10, with examples based on *Cambridge Grammar* for each type of relation.

The selection of the type of relation in the sentence blackboard occurs by binding the LMA to a specific structure group in the head and/or dependent rows (S4.2). So, in the relation *likes pizza*, V3-*likes* is bound to the structure group SG3-Object as head, and N4-*pizza* is bound to the structure group SG3-Object as dependent. The binding of an LMA to a structure group results in the activation of the neural circuit selective for that structure group. One role of these circuits is to sustain activation related to the structure group that was selected (S7.2). In this way, words at different sequential positions in a sentence can bind to each other, with a limited possibility of backward processing (S8).

In Figure 2, the structure group SG3-Object for V3-*likes* as head is concurrently active with the structure group SG3-Object for N4-*pizza* as dependent, even though the words are not concurrently active. Hence, a binding between V3-*likes* and N4-*pizza* can be made in the connection matrix. It is initiated by the concurrent activation of selection signals that open the selection gates for SG3-Object in head and dependent rows. This results in the binding of rows H3 and D4 in the connection matrix. Earlier steps provided the binding of N2-*Sue* as the subject and V3-*likes* as the verb of the sentence (which shows that the same LMA can be the head in one binding and the dependent in another). The entire process resulted in the construction of the connection path for *Sue likes pizza*. Figure 2 illustrates the open/active bindings in this connection path. A sentence structure is stored in the blackboard by the activated binding circuits in its connection path. These are sufficient to reactivate the sentence structure, e.g., for answering questions (S7.6).

In the HD-NBA, a connection path is an 'intrinsic' structure of a sentence (S5). Connection paths for all types of relations in Tables S1-S10 can be constructed, including sentences that consist of combinations of these relations (S4.4). Other linguistic aspects can also be dealt with in this manner, such as agreement and case (S3.2), long-distance dependencies (S3.3), scope (S5.1) and inflection (S5.2). Sentence structures can also be transferred from the neural sentence blackboard to long-term memory (van der Velde & de Kamps, 2006).

## 4. **Connection paths versus control-initiated activity**

Because the HD-NBA does not depend on address codes, information on which rows are involved in a sentence structure is not available. Consequently, control, as given by selection and binding signals, operates over the whole sentence blackboard. These control signals select the LMAs to which words (phrases) are bound, select whether LMAs functions as head or dependent, select the structure groups to which LMAs are bound, and control the flow of activation into or out of the connection matrix. The structure of a sentence corresponds with the sequence of control signals needed to construct its connection path.

Hence, two sentences will be structurally different if the construction of their connection paths requires different sequences of control signals. Conversely, two sentences are structurally similar when the same sequence of control signals constructs their connection paths. Nevertheless, their connection paths are different because they involve non-overlapping rows in the sentence blackboard. Consequently, multiple sentences can be stored concurrently (S7.6).

The difference between connection paths and control-initiated activity plays an important role in the simulations of the HD-NBA, as presented below.

## 5. Simulations of neural activity related to sentence processing and structure

Details of the simulations are presented in S7. All simulations were performed with four different capacities of the sentence blackboard, consisting of 15, 20, 25 and 30 rows (head, dependent), referred to as W15, W20, W25 and W30 respectively (S7.3).

The first simulation concerns the intracranial observations of brain activity during sentence and phrase processing, presented in Figure 1F of Nelson et al. (2017). The measurements were conducted in the brains of patients treated for epilepsy. They were aimed to observe how constituents of phrases or sentences are formed, with the assumption that in constituents such as *Ten students* or *Ten sad students*, the words would construct 'nodes' that would 'merge' when the constituent was complete. Hence, brain activity would increase when more nodes needed to be merged, and it would decline once the nodes were merged into a constituent.

In particular, Figure 1F of Nelson et al. (2017) presents the intracranial activity during the processing of *Ten students*, *Ten sad students* and *Ten sad students of Bill Gates*. Activity indeed increased during the processing of a constituent and declined after its completion. Moreover, in *Ten sad students of Bill Gates* this process not only occurred with the entire constituent but also with the part *Ten sad students.* This would indicate that humans start to form constituents as soon as possible (e.g., Tanenhaus et al., 1995).

Figure 3A shows the total activity in the neural sentence blackboard, averaged over W15 to W30, for the same phrases. Their HD-NBA structures are presented in Figure S19A. The total activity in the sentence blackboard also increases during the processing of a constituent and declines after its formation, including for *Ten sad students* in *Ten sad students of Bill Gates*, in line with the observations by Nelson et al. (2017, Figure 1F). See also Figure S19B.

Figure 3B illustrates the average total activity (Total) for *Ten sad students of Bill Gates*, and its division over the gating circuits (Gating), the working memory populations (WM, connection path), and the other populations in the blackboard (Other). Table S12 presents the number of populations in these categories for W15-W30. Most populations belong to the gating circuits, which play a crucial role in the processing of a sentence. Figure 3B illustrates that their activity dominates the total activity, including its activation pattern.

The working memory populations are crucial for storing sentence structures in the architectures. In particular, their activity constitutes the connection path of a sentence in the blackboard, and is sufficient to retrieve sentences and answer questions (S7.6). Figure 3B illustrates that their contribution to the total activity is small, and it does not have the same pattern.

Figure 3C illustrates Total activity for W15-W30. Their activation patterns are similar, but increase from W15 to W30. Figure 3C also illustrates the average WM (connection path) activity and that for W30. The latter is equal to the average WM activity (but see S7.5). This also illustrates the relation between WM activity and the sentence structure (connection path), because this structure does not change from W15 to W30.

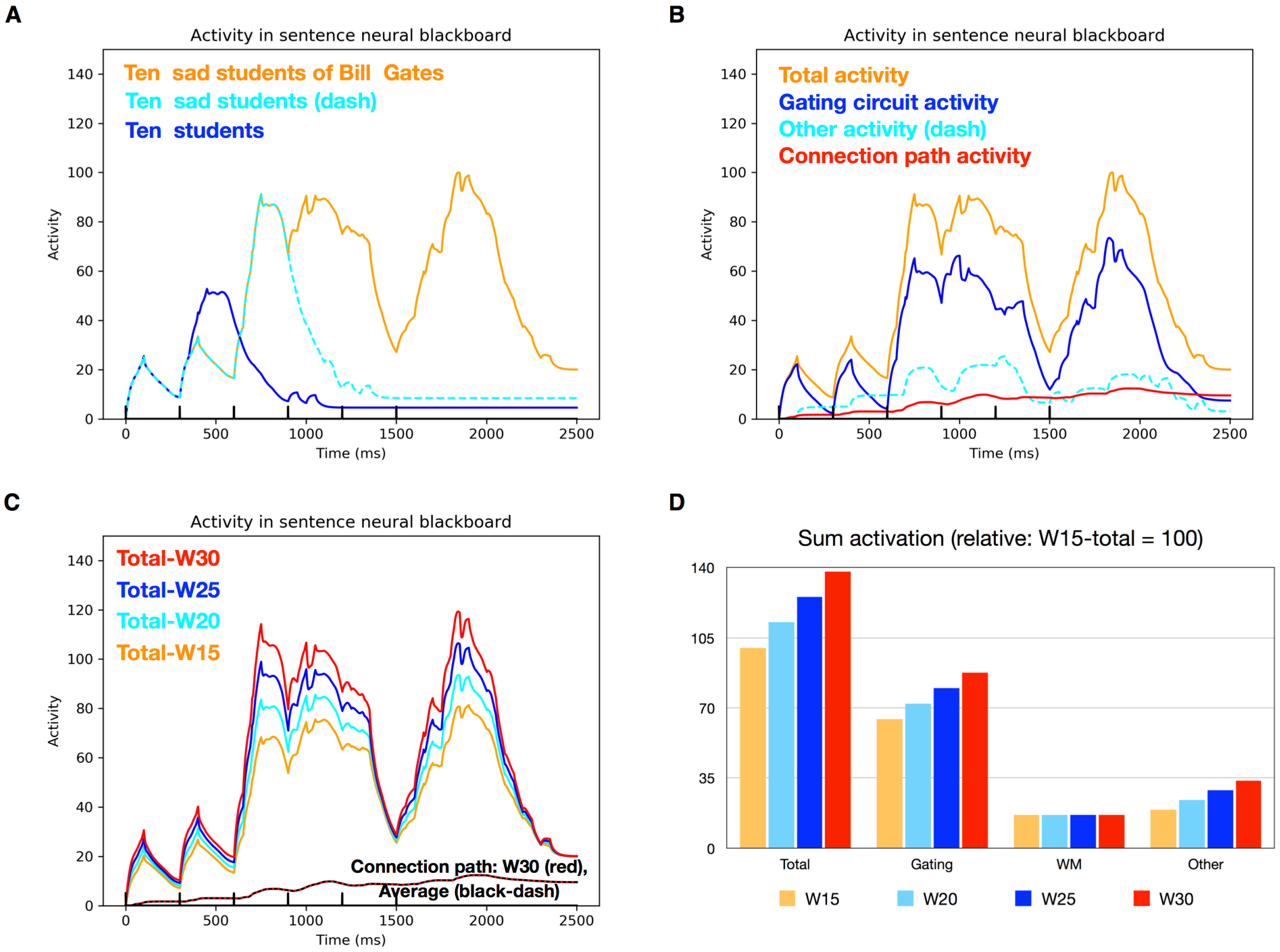

**Figure 3. Simulation of intracranial brain activity**. (**A**) Average total HD-NBA activity related to Figure 1F of Nelson et al. (2017). (**B**) Average HD-NBA activity of all populations (Total), Gating circuits, Working memory (WM) in the connection path, and Other populations (Table S12) for *Ten sad students of Bill Gates*. (**C**). Total activity in W15, W20, W25 and W30 (S7.3), and WM activity in W30 (red) and average (black-dash) of the phrase in (B). (**D**). Sum activity in W15 to W30, relative to Total in W15, for the categories Total, Gating, WM, and Other. Activity in spikes/ms. In (A) and (B), activity is normalized with the peak at 100. In (C), activity is normalized with Total average in (A).

Figure 3D illustrates the sum activity in the categories Total, Gating, WM and Other from W15 to W30, relative to the sum activity of Total in W15. It is clear that Gating dominates Total, and that activity increases with increasing architecture size, except for WM activity. The reason for this is the effect of control over the architecture during processing. Because this control operates over the entire blackboard, the activation it generates increases with the increasing architecture size (S7.5).

Figure 4 illustrates the influence of global control on activation with the simulation of the full sentence *I wonder how many miles I have fallen by this time*. Figure 4A shows again that the average total activity to produce the structure (given in Figure 4B) is primary based on gating circuit activity, which dominates its variation over time. See also Figure S19C.

Figure 4C illustrates the sum activity in the categories Total, Gating, WM and Other from W15 to W30, relative to the sum activity of Total in W15. It is again clear that Gating dominates Total, and that activity increases with increasing architecture size, except for WM activity.

This implies that observed neural activity (e.g., in sentence processing) would be primarily related to control, not to the sentence structure (connection path) itself.

Figure 4D illustrates this with the profile of 'syntactic effort' produced by a Minimalist Grammar, given by the node count (including empty nodes) in the tree structure for this sentence (Brennan et al., 2016, Figure 1). This profile, in particular the peaks at *how* and *by*, provided a better fit to fMRI activity observed during sentence processing compared to the more 'flat' profiles produce by a number of other grammars (Brennan et al., 2016).

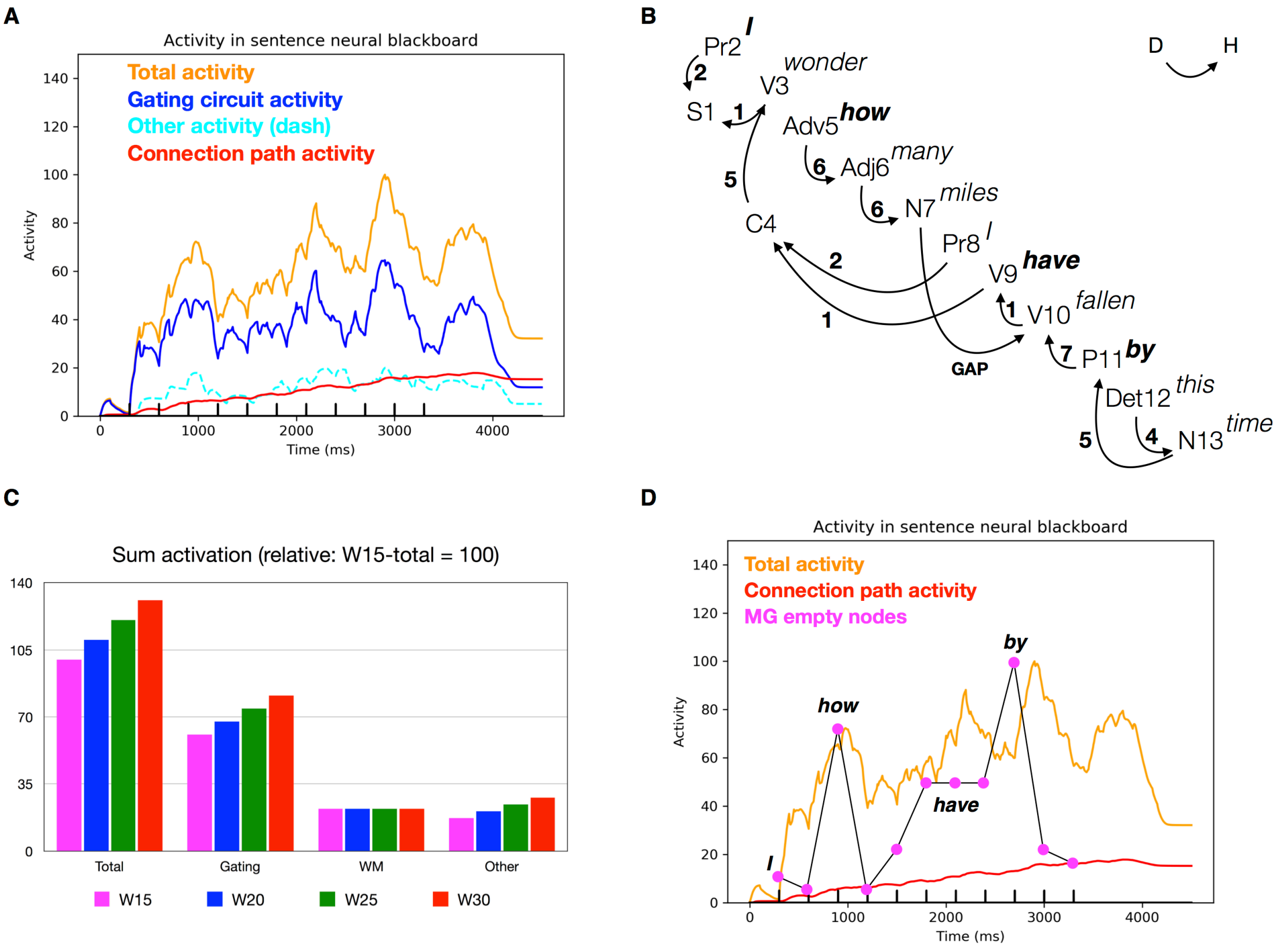

**Figure 4. Simulation of a sentence structure**. (**A**) Average HD-NBA activity of Total, Gating, Other, and WM (connection path) for *I wonder how many miles I have fallen by this time*. (**B**) HD-NBA structure of this sentence. Numbers refer to structure groups (SGs). LMAs in Table 1, SGs in Table S11. (**C**) Sum activity in W15-W30, relative to W15-Total, for the categories Total, Gating, WM, and Other. (**D**) Total and WM activity from (A) compared with a 'syntactic effort' profile produced by a Minimalist Grammar (Brennan et al., 2016). Activation in spikes/ms. In (A) and (C), activity is normalized with the peak at 100.

In Figure 4D, this profile is compared with the average Total and connection path (WM) activity in the HD-NBA obtained with this sentence. For comparison, the Minimalist profile is normalized with Total activity at *how*. Figure 4D shows that Total activity in the sentence blackboard also has (relative) peaks around *how* and *by*. But, as illustrated in Figures 4A and 4C, they are primarily derived from control-initiated activity over the entire blackboard, not from activity related tot the sentence structure itself.

This dominance of control-initiated activity suggests that even a 'flat' sentence-structure profile, as given by the connection path (WM) activity in Figure 4D, could produce a more varied pattern of brain activity, depending on the way it is processed. This would imply that deducing an activity pattern from a syntactic structure directly would not suffice to relate sentence processing to brain activity, as further illustrated below.

### 5.1. Simulations of sentence complexity and ambiguity

Figure 5A illustrates the average total and connection path activities in the HD-NBA related to the following sentences, with increasing sentence complexity (AC < SR < OR), from Just et al. (1996):

*The reporter attacked the senator and admitted the error.* (AC = Active conjoined)
*The reporter that attacked the senator admitted the error.* (SR = Subject relative)
*The reporter that the senator attacked admitted the error.* (OR = Object relative)

Their HD-NBA structures are presented in Figure S20. An increase of fMRI activity was observed with these sentences in the order AC < SR < OR (Just et al., 1996). Because the sentences have the same number of words, this increase could be contributed to their increasing complexity.

Figure 5B illustrates a similar increase in the relative sum of total activity in HD-NBA, which would be related to fMRI activity. But, as illustrated in Figure 5A, this increase does not result from activity related to the structures (connection paths) of these sentences. Their activity pattern is very similar. So, increased activity in the HD-NBA related to sentence complexity is also related to the global control over the entire architecture, not to activity related to sentence structures.

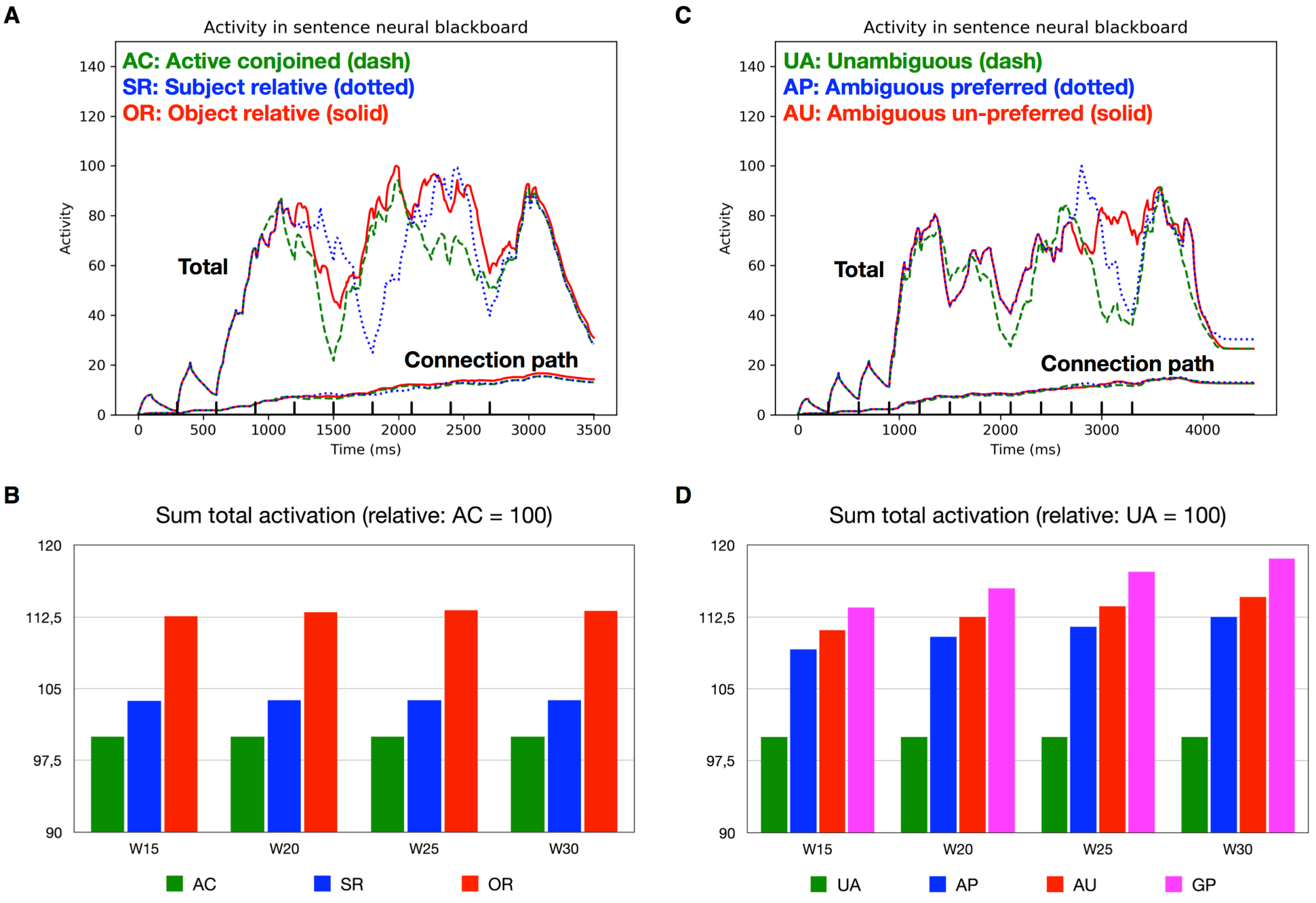

**Figure 5. Simulation of sentence complexity and ambiguity**. (**A**) Average total and connection path HD-NBA activity of sentences with increasing complexity. (**B**) Sum of total activity in (A) with W15 to W30, relative to active conjoined (AC). (**C**) Average total and connection path HD-NBA activity of sentences with increasing ambiguity. (**B**) Sum of total activity in (A) with W15 to W30, relative to unambiguous (UA). Activation in spikes/ms. In (A) and (C), activity is normalized with the peak at 100.

Figure 5C illustrates the average total and connection path activities in the HD-NBA related to the following sentences, with increasing sentence ambiguity (UA < AP < AU), from Mason et al. (2003):

*The experienced soldiers spoke about the dangers before the midnight raid.*
(UA = Unambiguous)
*The experienced soldiers warned about the dangers before the midnight raid.*
(AP = Ambiguous preferred)
*The experienced soldiers warned about the dangers conducted the midnight raid.*
(AU = Ambiguous un-preferred)

Their HD-NBA sentence structures are presented in and discussed with Figure S21, which also presents a garden path (GP) interpretation for the last sentence (AU). An increase of fMRI activity was observed with these sentences in the order UA < AP < AU (Mason et al., 2003). Because the sentences have the same number of words, this increase could be contributed to the increasing ambiguity of the sentences.

Figure 5D illustrates a similar increase in the relative sum of total activity in HD-NBA, including for the GP interpretation. But again, this increase does not result from activity related to the structures of these sentences (Figure 5C). So, increased activity in the HD-NBA related to sentence ambiguity is also based on global control over the entire architecture, not activity related to sentence structures.

This observation is further corroborated by the fact that in Figure 5D activity differences become larger with increasing architecture size (which also occurs slightly in Figure 5B). Given the individual differences in capacity for language comprehension (Just & Carpenter, 1992), these results predict that activity differences related to complexity and ambiguity would be higher for individuals with higher comprehension capacity.

The simulations presented here suggest that brain activity related to sentence processing primarily concerns control-initiated activity, instead of activity related to the sentence structure (connection path) itself. The simulations presented in Figure 5 also indicate that the HD-NBA can deal with sentence performance issues related to complexity and ambiguity (S8).

## 6. Conclusion

The productivity of natural language seems to be at odds with the 'fixed' connection structure of the brain, also because the latter could suggest that neural structures are associative only (Jackendoff, 2002). However, even though connections between individual neurons may be associative, this does not imply that higher-level connection structures are also associative.

Instead, the HD-NBA illustrates that they could have all the requirements needed for the productivity of natural language. At higher levels of organisation, neural structures can represent and process relations. Moreover, using abilities related to small world like network structures, it is also possible to combine in-situ word structures in novel and arbitrary sentence structures. However, this applies only to familiar languages, precisely because the connection structures needed have to emerge during development (S2).

The ability of the HD-NBA to deal with sentence structures at the level of *Cambridge Grammar* could be a first step in the development of a neural architecture of natural language processing in the brain. Many more steps would be needed before this is achieved. However, the HD-NBA aims to show that we could rely on the brain as a connection structure to achieve this aim.

## 7. References

References can be found in section S10 of the Supplementary Information.

## Acknowledgements

The author would like to thank dr. Marc de Kamps for his valuable comments on an earlier draft of the article.

Supplementary Information for

# Sentences as connection paths: A neural language architecture of sentence structure in the brain

*Frank van der Velde*

University of Twente, Cognition, Data and Education, BMS
Drienerlolaan 5, 7522NB Enschede, The Netherlands
f.vandervelde@utwente.nl; veldefvander@outlook.com

## Table of Contents

"For the entire nervous system *is* nothing but a system of paths between a sensory *terminus a quo* and a muscular, glandular, or other *terminus ad quem."*

William James (*Principles of Psychology*, 1890, p. 108, italics by James)

## S1. Overview

The supplementary information provides background and additional material related tot the main article.

- Section 2 discusses the Head-Dependent Neural Blackboard Architecture (HD-NBA) and its core assumptions in more detail. It also discusses the main choices that were made in the development of the architecture. One of the main choices in the development of the HD-NBA is to achieve unlimited productivity with 'minimal' means. This aim has resulted in significant modifications and extensions compared to the first version of the architecture (the NBA, van der Velde and de Kamps, 2006).

- Section 3 presents and motivates the structure groups in the HD-NBA, which are based on a survey of the comprehensive review and analysis of English sentence structures by Huddleston and Pullum (2002), referred to here as *Cambridge Grammar* (or CG).

- Section 4 presents more details on the structure and circuits of the HD-NBA, as introduced in Figure 2[1].

- Section 5 discusses the intrinsic structure of a sentence provided by a connection path in the HD-NBA, and its relations with issues like scope and inflection.

- Section 6 discusses parsing in the HD-NBA, related to selective control of binding in the architecture.

- Section 7 presents details related to the simulations presented in the main article, as well as some additional simulations.

- Section 8 discusses how the HD-NBA could account for performance issues, as related to sentence ambiguity and complexity.

- Section 9 presents a brief discussion of the HD-NBA.

- Section 10 provides the references for the main article and the supplementary information.

- Section 11 lists the abbreviations used in the main article and the supplementary information.

[1] References without 'S' refer to figures and tables in the main article.

## S2. Outline of the Head-Dependent Neural Blackboard Architecture

Figure S1 illustrates the main components that play a role in the Head-Dependent Neural Blackboard Architecture (HD-NBA). In correspondence with Figure 1, the bottom half of the figure is a representation of the hub and spoke model of semantic representation presented by Lambon Ralph et al. (2017), which forms the basis for the in-situ concept structures (in-situ concepts, for short) in the HD-NBA. The spokes in this model are neural representations of modality specific conceptual features. These neural representations are (potentially) distributed in two ways. First, they are found in different (cortical) areas in the brain (e.g., see Huth et al., 2016). Secondly, within each modality (e.g., vision) concepts will often share parts of their neural representation in that domain (e.g., McClelland and Rogers, 2003). However, a concept is not just a collection of domain specific and distributed spoke representations. Instead, these domain specific representations are also interconnected with and via one or more transmodal hub-like representations located in the anterior temporal lobes.

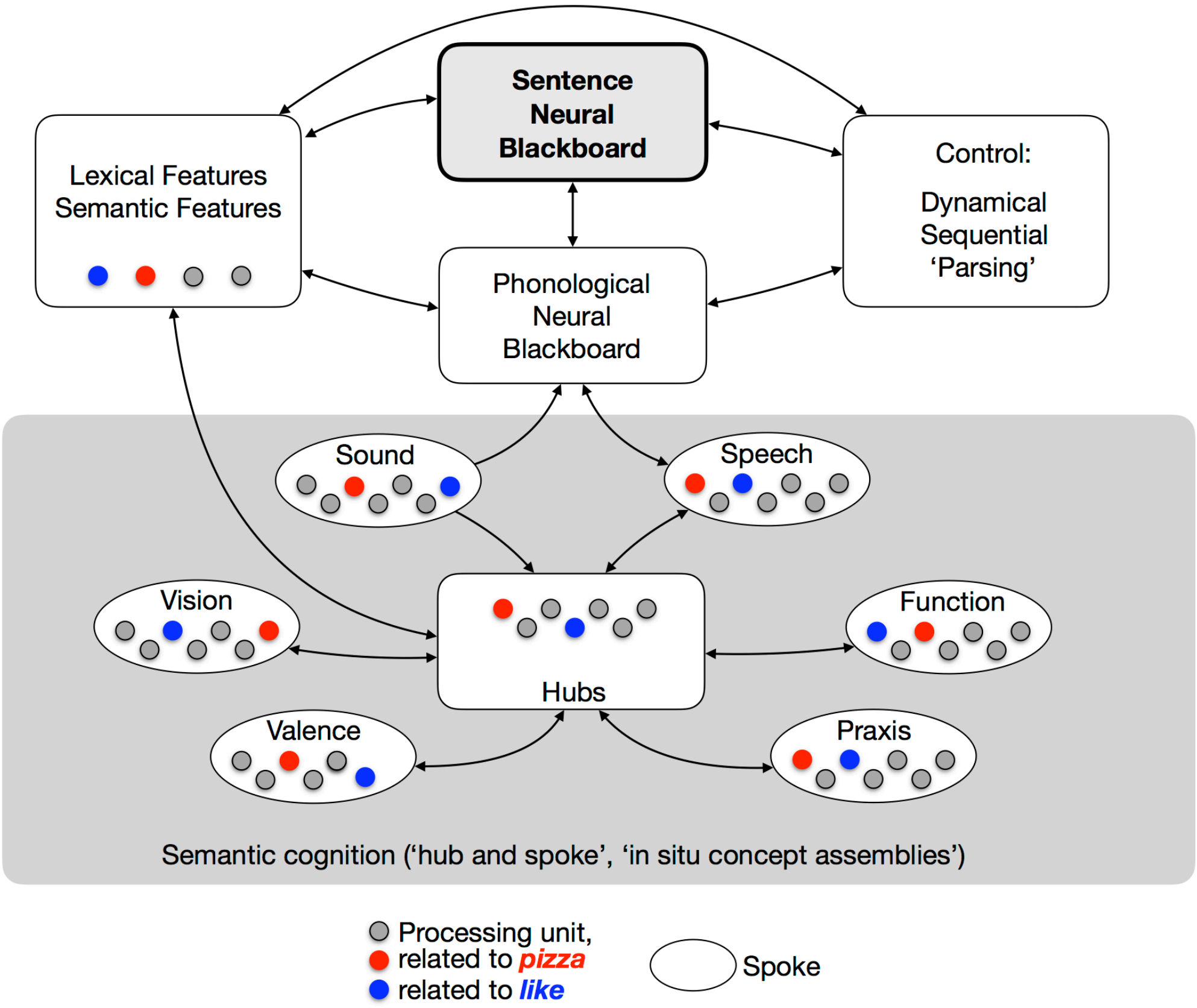

**Figure S1**. **Outline of the HD-NBA.** The main components of the HD-NBA with their interconnections and connections with in-situ concepts in a hub and spoke model of semantic representation, illustrated with *pizza* (red) and *like* (blue). The illustration of the hub and spoke model is based on Lambon Ralph et al. (2017).

Figure S1 also illustrates three other main components of the HD-NBA. These are the 'sentence neural blackboard', the 'phonological neural blackboard' and forms of control (dynamical, sequential, 'parsing') that regulate activation and processing in the architecture. The current development of the HD-NBA mainly concerns the structure of the sentence neural blackboard and how it operates in representing and processing sentence structures. But all other components are needed for this as well, so their role will discussed also, albeit in a more schematic manner (leaving their fuller development to future research).

The aim of the HD-NBA is to show how any arbitrary and potentially novel phrase or sentence structure can be created in terms of a connection path that reflects the structure and integrates the in-situ concepts (words) in it. The connection path allows the generation of behavior on the basis of the entire sentence structure, such as answering queries like *Sue likes?* in case of the sentence *Sue likes pizza*.

Figure S1 illustrates sentence processing in verbal interaction, which is the canonical way in which language is used and has evolved (but a similar process could be described for other forms of linguistic interaction). The words in the sentence *Sue likes pizza* are first processed based on their sound pattern. The HD-NBA assumes that this involves a phonological neural blackboard, in which a temporal phonological structure of each word in the sentence is formed, in interaction with the long-term phonological information about these words stored as lexical features.

The phonological neural blackboard also interacts with the sentence neural blackboard, in which a temporal sentence structure of the sentence is formed, in the form of a temporal connection path between the in-situ concepts (words) in the sentence. To reproduce that sentence, or answering queries about it, the sentence structure in the sentence neural blackboard interacts with the phonological neural blackboard, which results in the production of speech on the basis of the in-situ neural structures underlying the words in the sentence (which could interact in speech production). The entire process is controlled by neural circuits that regulate the parsing process, the sequential activation of the sentence (or answer), and the dynamics of activation in the architecture.

Furthermore, Figure S1 illustrates two types of information that also influence processing in the neural blackboards. These are the 'lexical features' and the 'semantic features' of words. The lexical features represent syntactic (structural) information about words that influence sentence processing in the sentence blackboard. Examples are the fact that *pizza* is a noun and *like* is a verb, or the fact that a word is single or plural. The semantic features represent semantic information about words that could also influence sentence processing. An example is given by the differences between preposition phrases in the following sentences (derived from Culicover and Jackendoff, 2005):

(1a) *Cook the potatoes that you bought at the market with Sue.*
(1b) *Cook the potatoes that you bought at the market for 15 minutes*.

These sentences have two preposition phrases that are modifiers or adjuncts of the verbs in the sentence. In sentence 1a they both seem to modify the last verb in the sentence (*bought*). One might assume that this follows from their sequential position in the sentence, because they are placed after the second verb. However, sentence 1b shows that this assumption is too simple. Here, the second preposition phrase in the sentence (*for 15 minutes*) clearly modifies the first verb *cook* and not the second verb *bought*, even though its position is the last in the sentence. In the HD-NBA, this difference in sentence structure would result from the semantic relation between a verb like *cook* and a preposition phrase like *for 15 minutes*, not from the linear order of these components in the sentence. However, the underlying processes are not further discussed here.

### S2.1. Core assumptions

The core assumptions of the HD-NBA are briefly summarized here, but they are discussed and illustrated throughout the article (further discussions can be found in van der Velde, 2015a, 2015b; van der Velde and De Kamps, 2006, 2011, 2015). Below, they are also compared with the specific choices that were made in the development of the HD-NBA.

First, as noted, sentence structures are connection paths that interconnect in-situ concepts, which are always content-addressable (directly accessible) and remain grounded in perception and action, also when they are part of a sentence structure. A connection path between concepts provides an 'intrinsic' structure of a sentence (section S5). In this way, the concepts *Sue* and *like* in the question *What does Sue like?* will activate the part of the connection path of *Sue likes pizza* they belong to, because the in-situ concepts *Sue* and *like* are the same in the question and in the sentence. In turn, further sequential activation of this connection path allows the activation of the answer *pizza* (section S7.6).

Second, the HD-NBA is based on the assumption that human cognition derives from a unique integration between (simply stated) forms of classification and forms of combination. Forms of classification (as could be achieved by associations or forms of deep learning) underlie the development of in-situ concepts and dynamical and parsing control of sentence processing. They derive from associations or learning mechanisms that operate in a (relatively) slow manner.

Forms of combination, on the other hand, concern the ability to process and produce novel combinations of familiar entities on the fly, such as novel sentences based on familiar words or novel visual scenes based on familiar shapes and colors. They operate fast and depend on combinatorial architectures. These architectures will have developed over time, but in a 'here and now' process their operation does not depend on learning.

The complementary role of classification and combination can be illustrated with parsing (section S6). In the HD-NBA, a dependency parser will give 'binding' instructions based on learned generalizations about the structure of the sentence. For example, the parser has learned that a sentence like *Sue likes pizza* is of the type Noun Verb Noun, in which the first noun has to bind as the subject of the sentence and the second noun as the object of the verb. This information controls the process of creating a sentence structure in the form of a connection path. However, this path interconnects the neural representations of the actual words used in the sentence, not just their type characteristics. This allows the production of behavior (e.g., answering questions) related to the actual (and potentially novel) sentence content. So, the parser and the neural sentence blackboard are both needed to form sentence structures as connection paths, and they play a complementary role in that process.

Third, the HD-NBA achieves the unlimited productivity of natural language using a 'fixed' network structure. This agrees with the observation that sentences can be processed and produced in a rate of 3 to 4 words per second (e.g., Rayner and Clifton, 2009). This rate precludes the development of new connections in the architecture to combine (or 'bind') words that have not been combined before (as is the case in any novel sentence). Of course, synaptic adaptations in the connection structure could occur in the overall architecture during this time, but they are not the basis of its combinatorial productivity. Instead, the productivity of the architecture depends on its 'small world' like network structure (section S4.3).

Fourth, the unlimited productivity in the HD-NBA depends on 'neural blackboards' that allow arbitrary and novel sentence (and word) structures to be formed. A crucial element in this process is a form of 'binding'. Binding ensures that words can be combined arbitrarily and in different roles in sentence structures. So, for example, every noun can bind to any transitive verb in the role of object. The neural blackboards provide the 'logistics' needed for this combinatorial productivity, achieved in a 'connection matrix'.

In terms of its component structure, a matrix is a (rank-2) tensor. Hence, binding with a connection matrix can be seen a form of 'tensor' binding (Smolensky, 1990). In the HD-NBA, however, words always remain content-addressable in sentence structures. This entails that they are not bound in, e.g., tensors directly on the basis of their neural structures (e.g., as 'vectors' that are then bound in tensors of higher ranks or in reduced vectors). Instead, words are bound by means of their type information (e.g., noun, verb).

Fifth, in sentence processing the HD-NBA aims to combine a form of incremental processing (e.g., Marslen-Wilson, 1973; Tanenhaus et al., 1995) with limited forms of backward processing (e.g., section S8).

### S2.2. Choices

Next to these core assumptions, there are a number of choices on which the current development of the HD-NBA is based. Two main choices are outlined here. Others (e.g., the circuits in the model) are discussed in the related sections.

The first main choice is the aim to achieve unlimited productivity with 'minimal' means. In the HD-NBA a single connection matrix is used to bind words in all structural roles available, including syntactical aspects such as agreement and case. As a rank-2 tensor, a matrix is already the most minimal form of tensor binding. Thus, an architecture that uses only one connection matrix for binding all forms of structural information would indeed depend on the most minimal form of neural binding possible. However, this does not rule out the use of other connection matrices, e.g., in semantic forms of processing. This is not further elaborated here.

The choice for one connection matrix for sentence structures is motivated by Table 2, which presents an overview of around 100 specific relations between LMAs, based on a survey of *Cambridge Grammar*. In the first version of the model (NBA, van der Velde and de Kamps, 2006) it was assumed that there would be a connection matrix for each specific LMA relation. However, having to use around 100 specific connection matrices is unwieldy. Hence, the amount of binding relations between LMAs illustrated in Table 2 prompted a number of modifications of the architecture presented by van der Velde and de Kamps (2006). These have resulted in the use of only one connection matrix for binding LMAs in all structural relations (in line with the aim of 'minimal' means).

The modifications of the original architecture (van der Velde and de Kamps, 2006) are based on the observation that connection paths in the neural sentence blackboard have to be distinguishable when they implement different sentence structures and relations (see Figure 1).

Connection paths in the neural sentence blackboard can be distinguished in a number of ways. First, by the pair of LMAs involved (Table 1). So, a connection path of an N-Adj binding is different from a connection path of a V-N binding. This difference is reflected in the different types of selection or binding gates that need to be activated to achieve the binding or to reactivate the sentence structure, e.g., to answer queries. The types of these gates depend on the types of selection or binding signals needed to activate them (section S4).

A second way to distinguish connection paths is with the role of head and dependent. The notions of head and dependent now used in the HD-NBA are related to but not equivalent with the notions head and dependent used in linguistics, as in *Cambridge Grammar*. In *likes pizza*, *likes* is indeed also the linguistic head and *pizz*a the linguistic dependent. However, in the sentence *Sue likes pizza*, the verb (*likes*) is the linguistic head of the whole sentence, but in the binding between S1 and V3-*likes* (Figure 1), S1 is the head and V1-*likes* is the dependent.

In the HD-NBA, head and dependent are distinguished by different selection gates, which create different connection paths in the architecture. So, the binding N-Adj in *unnecessary changes* (Table S6), in which N-*changes* is the head of Adj-*unnecessary*, is different from the binding Adj-N in *inches long* (Table S6), in which Adj-*long* is the head of N-*inche*s, because of different selection gates that select N or Adj as head or dependent.

But Table 2 shows that there are, for example, four relations in which N is head and Adj is dependent. So, these relations cannot be distinguished on the basis of LMAs and head versus dependent only. In van der Velde and de Kamps (2006), they were distinguished by the different connection matrices used for binding. These provided the different types of selection and binding gates needed to make the distinctions between the connection paths involved.

With just one connection matrix for binding, different connection paths expressing different structural relations cannot be distinguished by different selection and binding gates inside the matrix. But they can be distinguished by the different gates that regulate the flow of activation going in or out of the connection matrix. This is the approach taken here.

However, using around 100 different types of selection and binding, as suggested in Table 2, is again unwieldy. Therefore, to achieve this selective control and binding, the relations expressed in Table 2 will be grouped in a number of 'Structure Groups' or SGs. Each of these will function as a 'structural' type of selection and binding gates in the process of binding LMAs in the sentence structure.

In Table 2, different SGs are required for all LMA relations with more than one entry (x) in the table. Because the maximum amount of multiple entries in the table is four (e.g., for N-Adj or V-N relations), at least four different SGs are required (each one used in one of these relations). All other bindings could be distributed over these four SGs.

Yet, the distributed nature of these four SGs makes it hard to analyse their role in a sentence structure (connection path). Moreover, it is conceivable that in the course of the development of the architecture more SGs would develop than would strictly be needed for Table 2, also because of the other roles they could play in sentence structures. In particular, they could also be used for syntactical features like agreement (e.g., single vs plural) or case (e.g. nominative vs accusative), as outlined in section S3.2.

Therefore, the second main choice made here is to define a larger set of SGs, in a way that is more function related. However, their role in the HD-NBA is always given by the way they are used in connection paths, not by the label given to them. The SGs are introduced in section S3, focusing on their linguistic features.

## S3. Structure groups

### S3.1. Structure groups based on LMA relations

The survey of the relations between LMAs illustrated in Table 2 motivated a re-arrangement of these relations in a set of 10 structure groups. Each of these structure groups (or SGs) implements different forms of relational binding in the HD-NBA, as will be clear from the examples discussed.

Each table below presents one SG. It presents the LMAs from Table 1 that can bind with that SG as head or dependent. The name of an SG is just an indication of the role it plays in binding LMAs. Its actual role is found in the way it is used. A specific SG cannot be used to bind the same pair of LMAs in the same manner (i.e., as heads and dependents) in different structural roles. Hence, different structural roles between the same pair of (head-dependent) LMAs require binding with different SGs.

In the SG tables, the name and number of the SG is presented on top. The head and dependent LMAs in a binding are presented in the first two columns. The third column illustrates the specific head-dependent LMA binding with an example from *Cambridge Grammar* (or derived from one in a few cases), with page numbers and sentence labels in brackets (when available).

In a connection path of a sentence structure, each SG will (usually) be referred to by the number given in its table (e.g., see *Sue likes pizza* in Figure 1).

| SG1-Verb | | |
|---|---|---|
| Head | Dependent | Example sentence |
| P | V<br><br>V (g:v) | CGp443 [8i]: a larger galaxy *than* initial measurements **suggested**.<br>CGp81 [20i]: He was expelled *for* **killing** the birds. |
| S/C | V<br>V (aux) | CGp94 [8iia] *She* **speaks** *French.*<br>CGp94 [8ia]: *She* **can** *speak French.* |
| V<br>V (aux) | V (g:v)<br>V | CGp82 [22i]: Kim *hates* **writing** thank-you letters.<br>CGp94 [8ia]: She *can* **speak** French. |
| Head (main word or phrase) in italics. Dependent (main word) in bold.<br>aux = auxiliary verb; g:v = gerund-participle (CGp82).<br>CG = Cambridge Grammar. | | |

**Table S1:** Structure Group (SG) related to the binding of verbs (SG1-Verb).

Table S1 presents SG1-Verb. The dependent in this SG is always a verb, either in a plain form, as in *She speaks French*, or as a gerund (G) or auxiliary, as in *She can speak French*. Both *can* and *speak* bind to a Verb LMA, e.g., V1-*can* and V2-*speak*, and these bind to each other with this SG (V1-V2, with V1 as head and V2 as dependent).

| SG2-Subject | | |
|---|---|---|
| Head | Dependent | Example |
| S/C | Adj | CGp536: Rather more **humble** *is how I'd like him to be.* |
| | C | CGp23 [7b]: **That he was guilty** *was obvious.* |
| | N<br>N (g:n) | CGp234 [11ib]: The **rain** *destroyed the flowers.*<br>The **killing** of the birds (CGp81 [20ii]) *was a mistake.* |
| | P | **On** the stage *is where I'd like him to be.*<br>Derived from CGp536 (see above). |
| | Pr | CGp234 [12ia]: **We** *enjoyed the show*. |
| | V (g:v) | CGp81 [19i]: **Destroying** the files *was a serious mistake.* |
| Head (phrase) in italics. Dependent (main word or phrase) in bold.<br>g:n = gerundial noun (CGp81); g:v = gerund-participle.<br>CG = Cambridge Grammar. | | |

**Table S2:** Structure Group (SG) related to bindings in the role of subject (SG2-Subject).

Table S2 concerns LMA binding that corresponds to the role of subject, with the subject LMA as dependent. Typically, that is a noun (noun phrase) or pronoun as the subject of a clause or sentence. However, an adjective (phrase) or clause could also function as subject. The same could apply to a preposition (phrase) as in *On the stage is where I'd like him to be* (derived from the use of the adjective phrase as subject).

A specific case is the use of a gerund (G) as subject. Traditionally, *destroying* is seen as a noun in *Destroying the files was a serious mistake*, because of its role as the subject in the sentence. But *destroying* seems to be a verb (*Cambridge Grammar*, p. 81). For example, it takes a direct object (compare *destroying the files* with *likes the pizza*). In contrast, a noun takes an indirect object in the form of a preposition phrase, as in *The destruction of the files was a serious mistake*. Moreover, an adverb (e.g., *slowly*) is needed to modify *destroying*, as in *Slowly destroying the files was a serious mistake*. Again, in contrast, a noun needs an adjective (e.g., *slow*) for that, as in *The slow destruction of the files was a serious mistake.* Furthermore, a determiner (e.g., *the*) cannot be added to *destroying the files*, as in **The destroying the files was a serious mistake* (here, the * denotes an incorrect or infelicitous sentence structure).

In the HD-NBA there is no problem with this ambiguous use of a verb like *destroying*. It is first bound to a Verb LMA, say V1, which is then selected as head and dependent. As head, V1 can take an object and a modifier, using the respective SGs presented below. As dependent, V1 can bind as the subject of a sentence or clause using the subject SG, as illustrated in Table S2.

However, the gerund *killing* is clearly a noun in *The killing of the birds was a serious mistake*, because it has a determiner and an indirect object (and would take an adjective as modifier). In this case, the gerund would bind to a Noun LMA directly.

| SG3-Object | | |
|---|---|---|
| Head | Dependent | Example |
| V | N<br>N (g:n) | CGp53 [1b]: He *washed* the **car**.<br>CGp81 [20ii]: She had *witnessed* the **killing** of the birds. |
| | P | CGp246 [note 22]: He *considered* **under** the mat an unsafe place for the key. |
| | Pr | CGp50 [1]: He *takes* **her** to school. |
| Head (main word) in italics. Dependent (main word) in bold.<br>g:n = gerundial noun.<br>CG = Cambridge Grammar. | | |

**Table S3**. Structure Group (SG) related to the binding of objects with verbs (SG3-Object).

Table S3 presents SG3-Object, for binding an object to a verb. The standard version is that of a noun or pronoun. But a preposition phase can also bind as an object to a verb, as illustrated in the table. A special case is the use of a gerund. Here, it is clear that *killing* is a noun, because it takes a determiner and an object in the form of a preposition phrase.

| SG4-Det (Determiner) | | |
|---|---|---|
| Head | Dependent | Example |
| N | Det<br>Det (adj) | CGp328 [1i]: **The** *manager* has just arrived.<br>CGp352 [64iib]: **two** *doctors.* |
| | N (gen) | CGp467 [41i]: **Kim's** *father* has arrived. |
| | Pr | CGp1083 [14i]: **Whose** *car* did you borrow? |
| Head (main word) in italics. Dependent (main word) in bold.<br>adj = adjective; gen = genitive.<br>CG = Cambridge Grammar. | | |

**Table S4**. Structure Group (SG) related to the binding of determiners (SG4-Det).

Table S4 illustrates SG4-Det, for binding determiners to a noun as head (here, in HD-NBA and linguistic terms). A word like *two* could be seen as an adjective, as in *the two cars*, but it could also function directly as determiner, as illustrated here. Similarly for the pronoun in *whose car*. The HD-NBA allows both forms of binding (the difference will result from the way a sentence is processed).

| SG5-Compl (Complement) | | |
|---|---|---|
| Head | Dependent | Example |
| Adj | C | CGp600 [6iiia]: *glad* **that he saw her**. |
| | N | CGp607: It was more *worth* the **effort** than I'd expected it to be. |
| | P | CGp608 [15ii]: Sincere thanks are *due* **to** all those who gave so generously. |
| Adv | C | CGp572 [7i]: He came to see me *directly* **he got the letter**. |
| Det | P | CGp56 [4ib]: *Several* **of** the boys were ill.<br>Note: fused head construction (CGp56). |
| N | Adj | CGp452 [3]: **financial** *advisers.* |
| | C | CGp439 [3ii]: the *rumour* **that the city had been captured**. |
| | N | CGp439 [2ia]: a **flower** *seller.* |
| | P | CGp439 [3i]: the *journey* **to** Rome. |
| | V (p-p) | CGp65 [1iv]: the *proposal* **recommended** by the manager. |
| P | Adj | CGp599 [3iv]: They took me *for* **dead**. |
| | Adv | CGp599 [3ii]: I didn't know about it *until* **recently**. |
| | C | CGp641 [19i]: This happened *after* **Stacy left**. |
| | N | CGp635: She stayed *in* the **house**. |
| | P | CGp599 [3i]: The magician emerged *from* **behind** the curtain. |
| | Pr | CGp239 [8ii]: The others were taken *by* **her**. |
| Pr | C | CGp637 [5iiia]: It was *I* **who told them**. |
| V | Adv | CGp224 [22ii]: They *treat* us quite **abominably**. |
| | C | CGp225 [25i]: I hadn't *noticed* **that she was looking so worried**. |
| | N | CGp218 [9]: I *gave* **Jo** a key.<br>Note: indirect object. |
| | P | CGp54 [4ia]: He *referred* **to** her article. |
| | Pr | CGp217 [4ii]: She *told* **him** the truth.<br>Note: indirect object. |
| Head (main word) in italics. Dependent (main word or phrase) in bold.<br>p-p = past-participial (CGp65).<br>CG = Cambridge Grammar. | | |

**Table S5**. Structure Group (SG) related to bindings in the role of complement (SG5-Compl).

SG5-Compl in Table S5 combines many different LMA bindings in the form of a 'complement.' *Cambridge Grammar* gives reasons for making distinctions between the roles of complements versus modifiers, adjuncts and external modifiers (presented below). Here, there are a number of different heads with each a set of dependents that function as complements. Most of these are described as complements in *Cambridge Grammar*.

Two different forms are included here as well. The first one is the 'fused head' construction in *Several of the boys were ill.* In *Cambridge Grammar*, *several* is seen as both head and dependent (in linguistic terms) because it acts as a noun phrase on its own. Here, is is just a determiner that takes a complement directly. The second one is the use of an indirect object as 'complement'. That is, this SG is used for binding indirect objects with verbs like *give.* Again, the name of an SG is just an indication. This SG can also be used for indirect object binding, because that does not stand in the way of any other binding with verbs with this SG.

| SG6-Mod (Modifier) | | |
|---|---|---|
| Head | Dependent | Example |
| Adj | Adv | CGp526 [2iii]: Their response was **unnecessarily** *long*. |
| | Det | CGp547 [33ii]: It surely isn't **that** *important*. |
| | N | CGp547 [33iii]: The nail was two **inches** *long*. |
| | P | CGp547 [33iv]: The view was *beautiful* **beyond** description**.** |
| Adv | Adv | CGp526 [2iv]: They had treated him **unnecessarily** *harshly*. |
| | Det | CGp573 [11ii]: I hadn't expected to be able to do it **this** *easily.* |
| | N | CGp574 [12i]: We arrived three **hours** *late*. |
| | P | CGp574 [13i]: They had behaved *badly* **in** the extreme. |
| Co | Adv | CGp1278 [9]: the guests *and* **indeed** his family **too**. |
| Det | Adv | CGp355 [3i]: **almost** *every* tie. |
| N | Adj<br>Adj (det) | CGp526 [2i]: They made a lot of **unnecessary** *changes*.<br>CGp330 [7i]: the **two** *mistakes* I made. |
| | N | CGp537 [27]: a **governmen**t *inquiry.* |
| | V | CGp78 [9ii]: a hurriedly **written** first *draft.* |
| P | Adv | CGp562 [2ie]: They are **almost** *without* equal. |
| | N | CGp599 [2i]: She died two **years** *after* their divorce. |
| Pr | Adv | CGp1277 [8ia]: They allowed the others but **not** *me* a second chance. |
| V | Adv | CGp526 [2ii]: They had *worried* **unnecessarily**. |
| Head (main word) in italics. Dependent (main word) in bold.<br>det = determinative.<br>CG = Cambridge Grammar. | | |

**Table S6**. Structure Group (SG) related to bindings in the role of modifier (SG6-Mod).

SG6-Mod in Table S6 combines many different LMA bindings that function as modifiers (i.e., the dependent modifies the head). The typical examples are an adjective that modifies a noun and an adverb that modifies a verb.

Here, *two* is seen as an adjective, instead of as part of a determinative phrase, as in *Cambridge Grammar* (but that could be implemented in HD-NBA as well). Also, *not* could be a determiner instead of an adjective.

| SG7-Adjunct | | |
|---|---|---|
| Head | Dependent | Example |
| N | C | CGp446 [15i]: Where's the *book* **I lent you**? |
| | P | CGp452 [3]: financial *advisers* **in** the city. |
| | Pr | CGp438 [7]: The *manager* **herself** had approved the proposal. |
| Pr | Det | CGp428 [7ia]: *We* **all** enjoyed it. |
| | N | CGp353: *we* **veterans**.<br>Note: CG (p353) sees *we* as determiner here. |
| S/C | C | CGp669 [7x]: **His assignment completed**, *Ed went down to the pub.* |
| | P | CGp604 [3iib]: **Ahead** of the ship, *there was a small island.* |
| V | Adv | CGp570: *Wait* here a moment, **please**. |
| | N | CGp669 [7vi]: They *arrived* last **week**. |
| | P | CGp669: She *ran* to the station **in** twenty minutes. |
| | Pr | CGp238 [7]: Sue typed the letter and *posted* it **herself**. |
| Head (main word or phrase) in italics. Dependent (main word or phrase) in bold.<br>CG = Cambridge Grammar. | | |

**Table S7**. Structure Group (SG) related to bindings in the role of adjunct (SG7-Adjunct).

SG7-Adjunct in Table S7 combines different LMA bindings that function as adjunct. In general, adjuncts are more distant in a phase or sentence compared to modifiers (*Cambridge Grammar*).

| SG8-PC (Predicate Complement<br>/Predicate Adjunct /Postpositive) | | |
|---|---|---|
| Head | Dependent | Example |
| N | Adj | CGp552 [46ii]: They want a *leader* **younger** than you. |
| | Adv | CGp452 [3]: financial *advisers* in the city **too**. |
| | P | CGp636 [4i]: I regard their *behavior* **as** outrageous. |
| Pr | P | CGp637 [8ia]: I regard *her* **as** a friend. |
| V | Adj | CGp217 [5i]: Ed *seemed* quite **competent**. |
| | N | CGp538: This sheet *is* pure **cotton**. |
| | Pr | CGp254: It *was* **he** who wrote it. |
| Head (main word) in italics. Dependent (main word) in bold.<br>CG = Cambridge Grammar. | | |

**Table S8**. Structure Group (SG) related to bindings in the role of predicate complement (PC), predicate adjunct or postpositive (SG8-PC).

SG8-PC in Table S8 combines different LMA bindings that function as predicate complement or adjunct, or as a postpositive modifier, as in *leader younger than you* (here the adjective *younger* itself has a preposition phrase as complement, using SG5).

| SG9-ExMod (External Modifier / Prenucleus) | | |
|---|---|---|
| Head | Dependent | Example |
| N | Adj | CGp551 [41ii]: It's **such** a *pity* you can't come. |
| | Adv | CGp452 [5i]: **even** all the *shareholders.* |
| P | Adj | CGp633 [21iib, 22]: **Incredible** *though* it seems, sales of these cars are falling. |
| S/C | Adv | CGp666 [1xxiv]: **Fortunately**, *we got there on time*. |
| | C | CGp1090 [35i]: **Whether it's ethical**, *I'm not so certain*. |
| | N | CGp1085 [20]: The **others** *I know are genuine*. |
| | P | CGp1092 [41ii]: **To** which safe *is this the key*? |
| | Pr | CGp1088 [31i]: I told her **what** *you insisted that we need.* |
| Head (main word or phrase) in italics. Dependent (main word or phrase) in bold.<br>CG = Cambridge Grammar. | | |

**Table S9**. Structure Group (SG) related to bindings in the role of external modifier or prenucleus (SG9-ExMod).

SG9-ExMod in Table S9 combines different LMA bindings that function as external modifier or prenucleus. These are more distant in a phrase or sentence structure, compared to adjuncts or modifiers.

| SG10-Coord (Coordination). | | |
|---|---|---|
| Head | Dependent | Example: |
| Adj | Co | CGp1286 [33ib]: He was extremely *tired* **and** irritable. |
| Adv | | CGp1286 [33iib]: She spoke very *quickly* **and** fluently. |
| N | | CGp1275 [2i]: *Kim* **and** Pat speak excellent French. |
| P | | CGp1276 [6iib]: He can see you *on* Tuesday **or** on Tuesday. |
| Pr | | They allowed *me* **but** not the others a second chance.<br>Derived from CGp1277 [8ia] (see below). |
| S/C | | CGp1280 [14iib]: *They arrived on Tuesday* **or** they arrived on Wednesday. |
| V | | CGp1286 [33vib]: He *hugged* **and** kissed his wife. |
| Co | Adj | CGp1286 [33ib]: He was extremely tired *and* **irritable**. |
| | Adv | CGp1286[33iib]: She spoke very quickly *and* **fluently**. |
| | C | CGp1280 [14iib]: They arrived on Tuesday *or* **they arrived on Wednesday**. |
| | N | CGp1275 [2i]: Kim *and* **Pat** speak excellent French. |
| | P | CGp1275 [2ii]: He can see you this afternoon *or* **on** Tuesday. |
| | Pr | CGp1277 [8ia]: They allowed the others *but* not **me** a second chance. |
| | V | CGp1286 [33vib]: He hugged *and* **kissed** his wife. |
| Head (main word or phrase) in italics. Dependent (main word or phrase) in bold.<br>CG = Cambridge Grammar. | | |

**Table S10**. Structure Group (SG) related to bindings in the coordinations (SG10-Coord).

Table S10 presents SG10-Coord, for bindings in coordinations. Here, the first LMA (e.g., *Kim* in *Kim and Pat*) is seen as the head (in HD-NBA terms) in the first binding (*Kim and*), whereas the Co LMA (*and*) is seen as the head in the second binding (*and Pat*). This is (more) in line with the use of coordination in *Cambridge Grammar*, which presents arguments as to why the coordinator forms a constituent with the second coordinate that follows. There are as yet no reasons for a different order of these bindings in the HD-NBA.

Coordinations can occur between different syntactic categories, as in *He can see you this afternoon or on Tuesday* (CGp1275 [2ii]), which are coordinations between a noun phrase and a preposition phrase. Coordinations can be ambiguous. For example, with *long poems and essays* (CG p1285 [32]) it is not clear whether the structure is (*long poems*) *and essays* or *long* (*poems and essays*).

### S3.2. Agreement and case

The SGs listed in Tables S1-S10 are based on a categorization of the relations between LMAs listed in Table 2. However, the categorization given is certainly not the only one possible. For example, SGs might be categorized in different ways based on learning mechanisms developed in further research.

But different SGs are needed to provide for the different connections paths in the HD-NBA that implement different sentence structures, as illustrated in Figure 2. Binding and selection gates related to SGs allow the selective and sequential creation and reactivation of a sentence structure. In particular, the SGs are needed to express the relations between LMAs (e.g., words) in a binding. So, in the phrase *likes pizza* (Figure 2) N4-*pizza* is the dependent in the relation SG3-Object and the V3-*likes* is the head.

The ability of SGs to express relations can be extended to other aspects of sentences structure such as agreement and case (if present). Agreement occurs, for example, in the difference between *this pizza* and *these pizzas*. Here, one part of the phrase (e.g. the determiner) has to match the other part (the noun) in a particular characteristic (single versus plural).

Because the adjective and the noun are already bound with the SG4-Det in these phrases, this can be used to implement the role of agreement as well. That is, the SG4-Det could be distinguished into two SGs, one for single (e.g., SG4-Det-s) and one for plural (SG4-Det-p). The in-situ concept of, e.g., *pizza* would select a binding of a determiner as dependent with SG4-Det-s, which also would be used to select SG4-Det-s for the binding of the head (*pizza*), and conversely for the plural case.

In a similar way, a distinction can be made for the binding with SG1-Verb and SG2-Subject as single (SG1-Verb-s and SG2-Subject-s) and the binding with SG1-Verb and SG2-Subject as plural (SG1-Verb-p and SG2-Subject-p). So, in *Sue likes*, N-*Sue* would be bound to the Sentence LMA S with SG1-Verb-s and V-*likes* would be bound to S with SG1-Verb-s. In, e.g., *boys like*, the plural versions of these binding would be applied.

So, the role of agreement in binding would require a straightforward extension of the HD-NBA structure as illustrated in Figure 2, in the form of additional SGs. The same applies for case. In fact, the difference between nominative and accusative would already be implemented with the use of SG2-Subject for nominative and SG3-Object for accusative. Other forms of case could be dealt with in the same way.

However, long-distance dependencies require a more detailed discussion, given in the next section.

## S3.3. Long-distance dependencies, questions and gaps

The issue of long-distance dependencies, and with it the role of 'gaps' in sentence structures (*Cambridge Grammar*) can be illustrated with questions. Figures S2A and S2B illustrate the HD-NBA dependency structures of *Sue signed [the] letter* and *Who signed [the] letter?* (CG, p. 1084 [17iii]). Here and below, the determiner (*the, a*) is ignored (indicated with []), because it is bound to the Noun LMA in the manner as illustrated in Figure 2. According to *Cambridge Grammar* (p. 1084), both sentences have the same structure. Hence, the question word Pr1-*who* is in the canonical subject position.

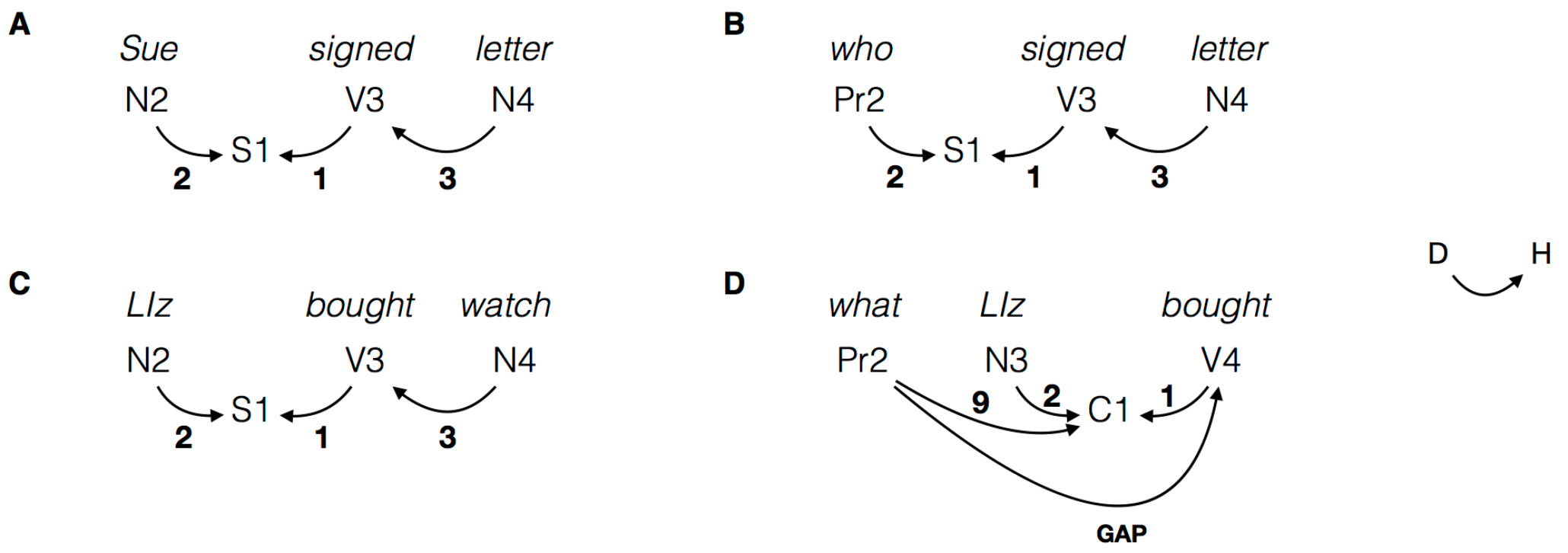

**Figure S2. Binding with GAP**. (**A**) The HD-NBA dependency structure of *Sue signed [the] letter* (ignoring *the*). (**B**) HD-NBA dependency structure of *Who signed [the] letter?* (ignoring *the*). (**C**) HD-NBA dependency structure of *Liz bought [a] watch* (ignoring *a*). (**D**) HD-NBA dependency structure of *what Liz bought*. Words are bound to LMAs (see Tables 1, S11). Numbers refer to SGs in Tables S1-S10 and GAP refers to SG11-GAP (see also Table S11). Dependents (D) point to heads (H).

Figures S2C and S2D illustrate the HD-NBA structures of *Liz bought [a] watch* (CG p. 48 [4a]) and *what Liz bought* from *I wonder what Liz bought* (CG p. 48 [4b]). According to *Cambridge Grammar* the question word *what* in Figure S2D is in the prenuclear position of the clause, not in the object position of the verb. Instead (in CG), there is a 'gap' at the object position of *bought.*

Here, the question word P1-*what* is bound with SG9-ExMod to C1. To express the relation of *what* with the object position of *bought* (*what* queries the content of that position) there is a binding between Pr1-*what* and V1-*bought* via a specific 'Gap' SG (SG11-GAP), illustrated in Figure S2D. Binding with SG11-GAP is also a form of H-D binding, here with Pr1-*what* as dependent and V1-*bought* as head. So, SG11-GAP also belongs to each head row and each dependent row, with the same connection structure as for the other SGs, as illustrated in Figure 2.

However, this raises an issue for the dependent rows. An illustrated in Figure 2, only one SG in the dependent row will bind in the connection matrix. But in Figure S2D, Pr1-*what* needs to bind as dependent to C1 (with SG9-ExMod) and to V1 with SG11-GAP. To solve this issue, one could assume that binding with SG11-GAP occurs in another connection matrix. However, given the aim of a 'minimal' structure outlined in section S2.2, this issue can be solved by assuming that SG11-GAP is an exception to the rule of single SG binding in dependent rows. This reflects another choice made in the development of the model. So, a dependent LMA can bind with only one SG from Tables S1-S10, but also with SG11-GAP. However, the use of SG11-GAP in sentence structures is limited and very selective. For example, an SG11-GAP binding with V as head will signify a binding as object (gap), because there is no other form of a gap that binds in this way.

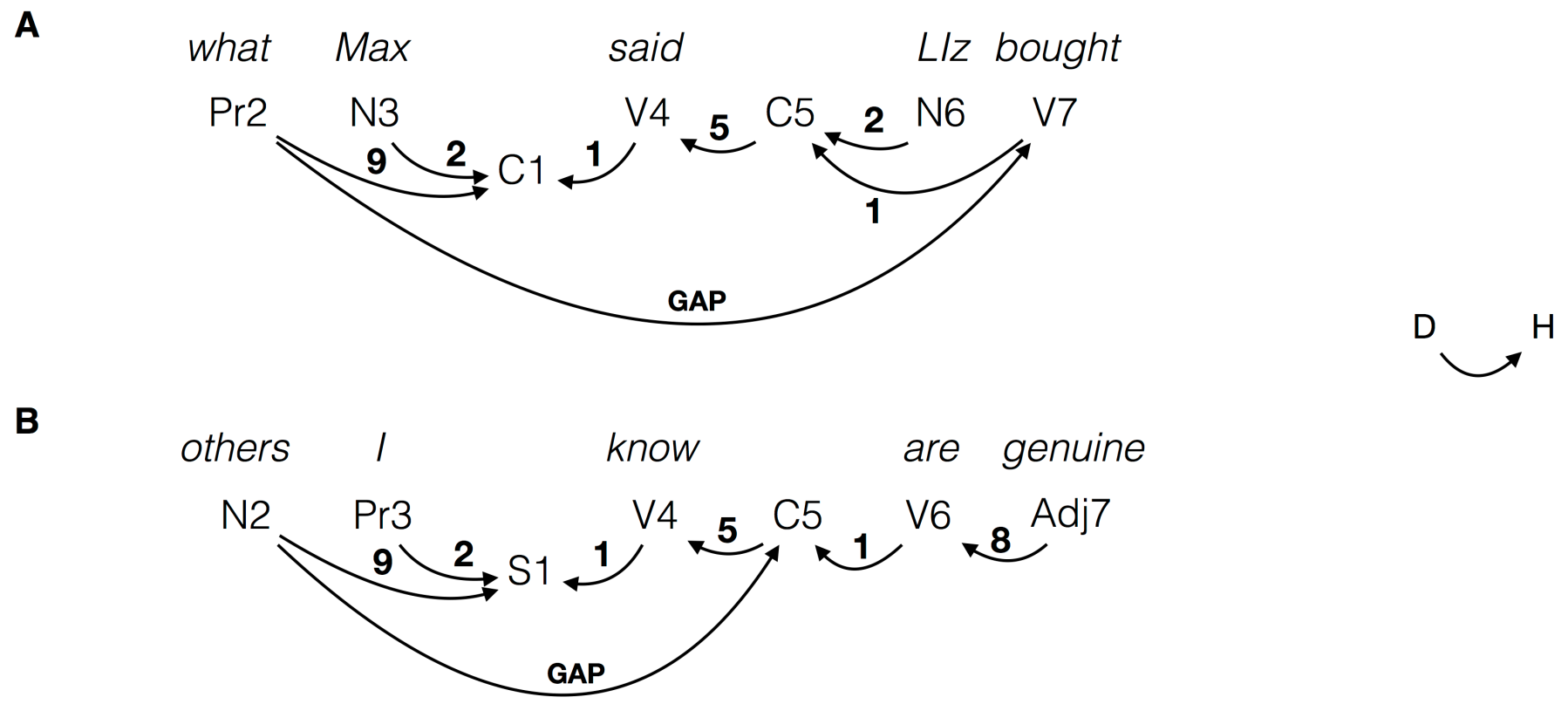

**Figure S3. Long-distance dependencies.** (**A**) HD-NBA dependency structure of *what Max said Liz bought*. (**B**) HD-NBA dependency structure of *others I know are genuine.* Words are bound to LMAs (see Tables 1, S11). Numbers refer to SGs in Tables S1-S10 and GAP refers to SG11-GAP (see also Table S11). Dependents (D) point to heads (H).

Figure S3A illustrates the role of SG11-GAP in establishing long-distance dependencies in the HD-NBA structure of the clause *what Max said Liz bought* (CG p. 49 [7]), from *I can't remember what Max said Liz bought* (CG p. 49 [6]). With SG11-GAP, Pr1-*what* is bound to V2-*bought*, even though it is in the prenuclear position of another clause (C1). Generally, the idea of first binding Pr1-*what* with C1 (using SG9-ExMod) is to prevent dangling LMAs during the progression of (sequential) sentence processing. It also marks a special role of Pr1-*what* in the sentence structure, which is clear from its fronted position in the sentence (hence not needed for Pr1-*who* in Figure S2B).

Figure S3B shows a gap in a subject position in an embedded clause in the sentence *[The] others I know are genuine* (CG p. 1085). It is clear that *others* is the subject of the embedded clause, as in *I know the others are genuine*. But in Figure S3B it is in the prenucleus position of the matrix sentence. In the embedded clause, it is bound to the subject position of C1 via SG11-GAP.

So, SG11-GAP can be used for different types of binding (i.e., expressing different roles). The type at hand is given by the combination of LMAs as the head and dependent in SG11-GAP binding. If a noun binds with V as head it is in the object position of the verb. Hence, SG11-GAP entails 'object'. If it binds with C as head, it is in the subject position of the clause, and SG11-GAP entails 'subject'. These differences result in different connection paths, and thus different intrinsic structures (see section S5), even though the same SG is used. This situation is characteristic of all SGs. They can play different roles in different connection paths, or conversely, they allow the formation of different connection paths, depending on the head and dependent LMAs they bind.

## S4. Basic structure of the architecture

Table S11 presents an overview of the LMAs and SGs in the HD-NBA. The LMAs are the same as in Table 1. The SGs are those presented in Tables S1-S10 and SG11-GAP. Table S1 also presents the labels used for the LMAs and SGs in sentence structures.

| LMAs | LMA labels | SGs (SG circuits) | SG labels |
|---|---|---|---|
| Adjective | Adj | SG1-Verb | **1** |
| Adverb | Adv | SG2-Subject | **2** |
| Clause | C | SG3-Object | **3** |
| Coordinator | Co | SG4-Det | **4** |
| Determiner | Det | SG5-Compl | **5** |
| Noun | N | SG6-Mod | **6** |
| Preposition | P | SG7-Adjunct | **7** |
| Pronoun | Pr | SG8-PC | **8** |
| Verb | V | SG9-ExMod | **9** |
| Sentence | S | SG10-Coord | **10** |
| | | SG11-GAP | **GAP** |

**Table S11. LMAs and SGs with labels in HD-NBA.** LMA = Lexical main assembly. SG = Structure group.

In this section, the basic structure of the HD-NBA, introduced in Figure 2, is presented in more detail. A further elaboration of the structure as used in the simulations is presented in section S7. The structure of the HD-NBA is based on 'populations', which are seen as local interconnected group of neurons that function as a unit (see section S7). This entails that a population is the smallest entity used in the architecture. All circuits are based on populations (or other circuits).

### S4.1. Circuits and working memory

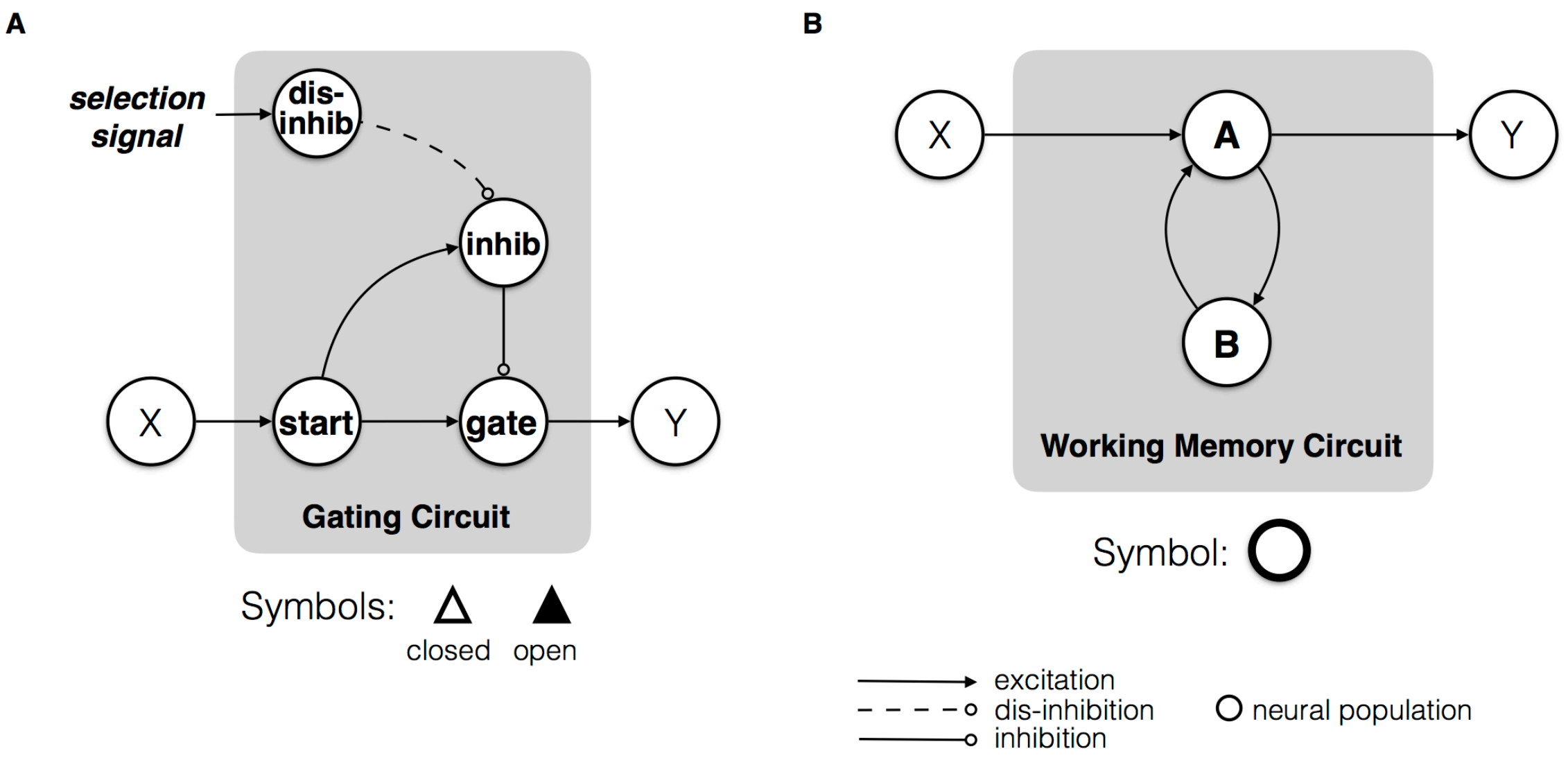

**Figure S4. Selection gate and working memory.** (**A**) Gating circuit for selective control, with symbol used to represent it in other circuits. (**B**) Working memory circuit to sustain activation in a 'working memory population' (population A), with symbol used to represent a working memory population in other circuits.

Figure S4A illustrates the gating circuit used for selective control (selection gate). It consists of a 'dis-inhibition' circuit. Circuits of this type are found in many places in the cortex and elsewhere in the brain (e.g., Letzkus et al., 2015). Here, it controls the flow of activation from population X to population Y. When X is active, it activates the excitatory population **start** in the circuit. This population activates the excitatory population **gate**. But it also activates the inhibitory population **inhib**, which inhibits **gate**. Because inhibition operates faster, **gate** is not activated, which blocks the flow of activation from X to Y.

The gating circuit can be unblocked or opened ('activated') by activating an external selection signal, which activates the inhibitory population **dis-inhib**. In turn, this inhibits population **inhib** (hence the name dis-inhibition). As a result, population **gate** is no longer inhibited, so activation can flow between the populations X and Y it connects. Hence, due to the gating circuit, the connection between X and Y is a 'conditional connection'. X will activate Y only when an external selection signal is active. Because this signal is selective, the flow of activation in the architecture is selectively gated (e.g., see Figure 2). The selection gate can be open (active) or closed (inactive), as illustrated with the respective symbols.

Figure S4B illustrates a working memory circuit that functions as a 'working memory population'. This entails that it operates as a population that remains active for a while after an initial (brief) external activation. It consists of two interconnected populations, **A** and **B**. Of these two, only population **A** is connected to the rest of the architecture. When population X activates **A**, it activates population Y and population **B**. In turn, **B** activates **A**. In this way **A** and **B** reactivate each other. Hence, population **A** remains active even when it is no longer activated by population X.

Once activated, a working memory population remains active (for a while) due to its internal (reverberating) connectivity. Working memory related activity in the frontal cortex is predominately found in superficial layers and is controlled by activation in deep layers (Bastos, et al., 2018). Over time, this activity will decline (Amit, 1989).

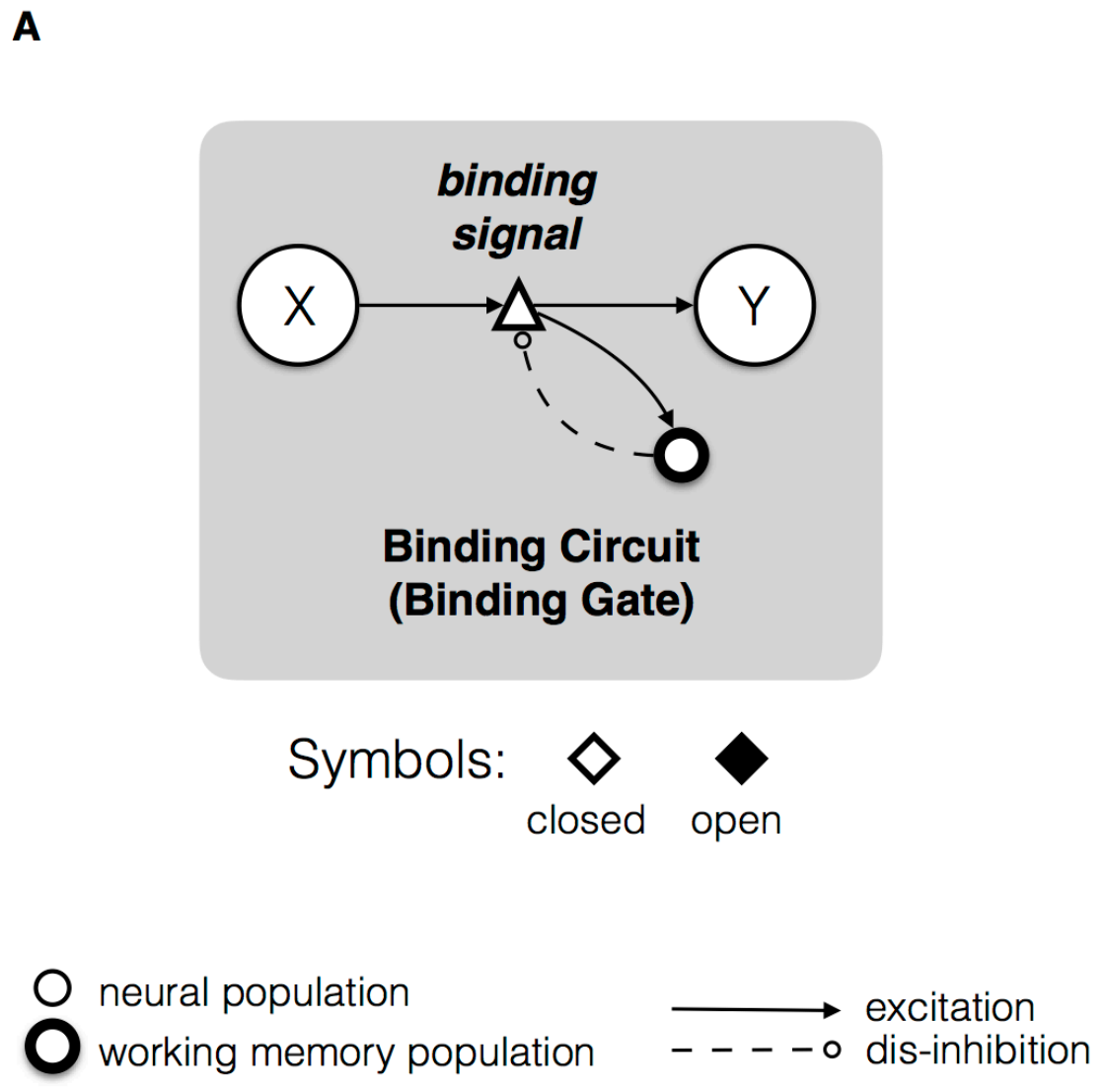

**Figure S5. Binding and linked binding gates.** (**A**) Gating circuit controlled by a working memory population. (**B**) Linked binding circuit, based on two connected binding gates.

Figure S5A illustrates the binding circuit (binding gate). It consists of a gating circuit in which the **gate** population activates a working memory population that in turn activates the **dis-inhib** population. This constitutes a form of 'binding'. As long as the working memory population is active, activation will flow from population X to population Y, which effectively makes them a single population. The binding gate can be open (active) or closed (inactive), as illustrated with the respective symbols.

Figure S5B illustrates the linked binding gate. It consists of two binding gates, in which the working memory population of one binding gate also activates the **dis-inhib** population of the other binding gate. Hence, the two binding gates in the linked binding gate will always be in the same state (open or closed). The linked binding gate can be open (active) or closed (inactive), as illustrated with the respective symbols.

Figure S6A illustrates the connection matrix in the HD-NBA. It consists of a matrix of 'cells'. Each cell is connected to a specific combination of a head (red) and dependent row (blue), which it can bind. Input to and output from a cell runs via specific populations (CM-in and CM-out, respectively) in the head row and in the dependent row connected to the cell.

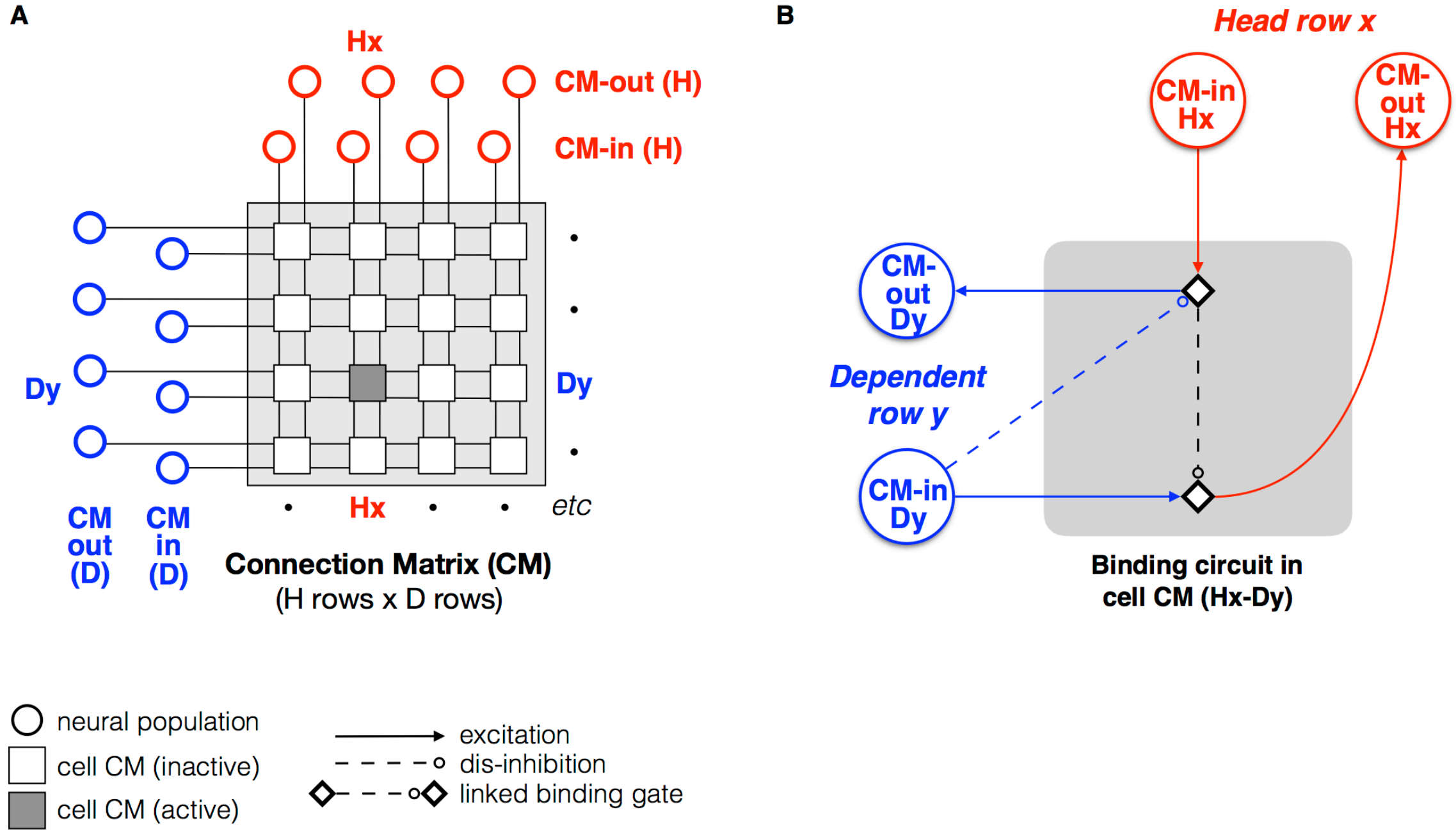

**Figure S6. Connection matrix.** (**A**) Overview of the connection matrix in the HD-NBA. It consists of cells, each connected to a specific combination of a head and dependent row. (**B**) A cell with a linked binding gate.

Figure S6B shows that a cell in the connection matrix consists of a linked binding gate. Of this, the top binding-gate receives input from the head row and produces output to the dependent row. Its activation is initiated by dis-inhibition input from the dependent row, that functions as the binding signal. The bottom binding-gate in the linked binding gate receives input from the dependent row and produces output to the head row. Once initiated, the linked binding gate in the cell remains active as long as the working memory population in its top binding-gate is active. As a result, activation can flow from the head row to the dependent row and vice versa.

The connection matrix is used for ongoing ('online') sentence processing. Its capacity is given by the number of head and dependent rows (assumed to by equal). This capacity can vary between language users (Just and Carpenter, 1992). A representation of a sentence structure (connection path) in the connection matrix and other circuits in the neural blackboard will last as long as the working memory activity in the matrix and circuits lasts. Over time, this activity will decay (or can be inhibited), so that the connection matrix can be used for the processing of newly presented sentences. Van der Velde and de Kamps (2006) describe and illustrate how sentence structures can be transferred from the neural blackboard to long-term memory.

### S4.2. Head and dependent rows

Figure 2 presents an overview of the architecture. It shows that the HD-NBA is organized in terms of rows. Each row starts with a 'word' population (W) that interconnects the phonological and sentence neural blackboards. Only one word population is active at any time. Because each row has only one word population, rows can be numbered with the number given to the word population. At some point, each row splits up in a head row and a dependent row, which each connect to the connection matrix. The number of rows available determines the capacity of the architecture (see section 7.3).

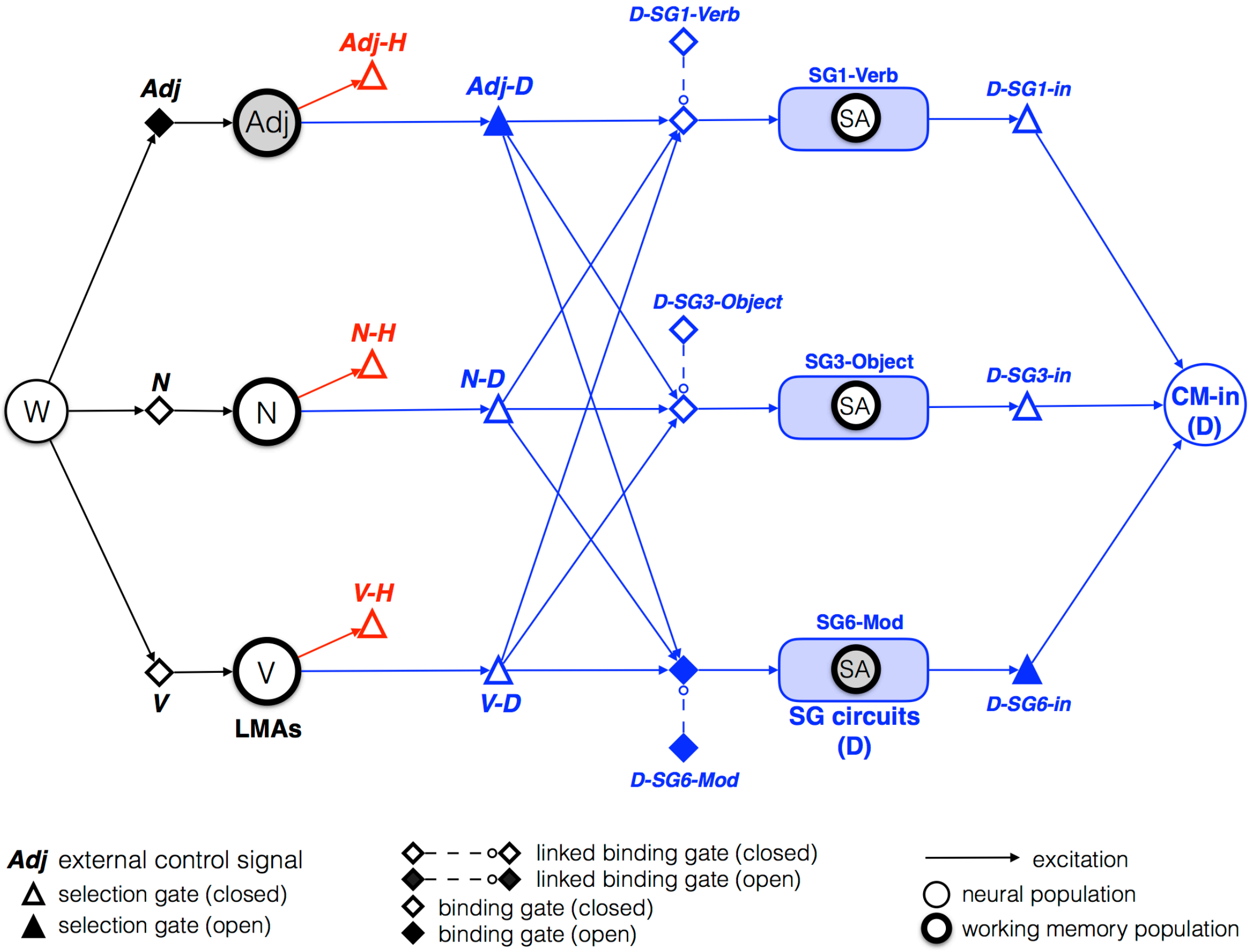

**Figure S7. Dependent and head row (to CM).** Illustration of the connection structure of a dependent row from the word population (W) to the connection matrix (CM). LMAs and SG circuits (SGs) in Table S11. D = Dependent. H = Head. SA = Subassembly.

Figure S7 illustrates the flow and control of activation from word population W to the connection matrix via a dependent row. The structure of all dependent rows is the same.

The structure of all head rows is similar, with the label D replaced by H.

The word population W is connected to binding gates, one for each specific LMA (Table S11). In turn, these are connected to working memory populations, one for each LMA. For simplicity, Figure S7 illustrates only the binding gates and LMAs for Adjective, Noun and Verb, but all are implied.

It is assumed that only one LMA is bound to a word population (which could be enforced explicitly with circuits if needed). Binding is initiated by the corresponding binding signal. This binding signal works over all rows, but is effective only for the row with the active word population. In Figure S7, the binding signal for Adjective has activated the binding gate for Adjective, which results in the activation of the Adjective LMA (Adj). In sentence structures, the activated LMA will have the same number as the word population (except in section S8). As long as the binding gate remains active, W is bound to Adj.

Once activated, LMAs retain their activity, because they are working memory populations. In the HD-NBA this activity will last until it is inhibited when an LMA of the same type has to be bound to another word population. As with all other forms of control, this (brief) selective inhibition also runs over all rows in the architecture. The sustained activation of LMAs helps the binding process in the architecture. In particular, it affords the possibility to select the SG used for binding in the sentence structure based on further sentence information. See section S6 for more details on the binding process in the architecture.

Next, a selection can be made between the head row and the dependent row for each LMA, based on LMA-selective head or dependent selection gates. In Figure S7, the selection is made for the dependent row, with the selection signal Adjective-Dependent (ADj-D). This selection gate is now open (for as long as this signal is active). Again, this selection signal works over all dependent rows. But is it effective only in the row with the active Adjective LMA. Activation flows through this open gate as long as Adj is active.

Then, each LMA-selective dependent selection gate is connected to SG-selective linked binding gates for all SGs listed in Table S11. In Figure S7, this is illustrated with the connections from the selection gates Adj-D, N-D and V-D to the linked binding gates D-SG1-Verb, D-SG3-Object and D-SG6-Mod. But all connections of this kind are implied.

This choice is in line with the aim to achieve unlimited productivity with 'minimal' means, as outlined in section S2.2. Another choice would be to have the full set of SG-selective linked binding gates for each of the LMA types separately, which would substantially increase the circuitry in the architecture. Alternatively, the connections between LMA-selective and SG-selective gates could be more restricted, depending on specific relations between LMAs and SGs. This is not further explored here.

Assume that Adj needs to bind in the connection matrix as dependent with SG6-Mod. To achieve this, the linked binding gate for SG6-Mod in the dependent row is activated with the binding signal D-SG6-Mod (again working over all dependent rows in the architecture). This produces the binding between Adj-D (and thus Adj) and D-SG6-Mod. This results in the activation of a working memory population referred to as 'subassembly' (SA) in a circuit selective for SG6-Mod in the dependent row. Details of this circuit are presented in section 7.2. Here, only the activation of the subassembly is relevant.

In the first version of the architecture (NBA, van der Velde and de Kamps, 2006), subassemblies played an important role in the binding between LMAs. In particular, they selected the specific connection matrix used for binding. In the HD-NBA, only one connection matrix is used. But subassemblies still play an important role in the binding process. Like LMAs, subassemblies retain their activation when activated. This also helps the binding process, in particular by postponing the binding in the connection matrix. Binding in the connection matrix occurs when the inputs from a dependent row and a head row are active simultaneously (Figure S6). The sustained activity of subassemblies helps in the timing of this event. In particular, it retains the selection of an SG even when the activation of the LMA has been inhibited by the selection of another LMA of the same type in another row.

In turn, the activated subassembly for SG6-Mod is connected to a selection gate that is activated (opened) with the SG-selective dependent selection signal D-SG6-in. The output of the selection gate D-SG6-in activates the population CM-in, which provides the entry from the dependent row to the connection matrix. This would result in binding in a cell of the connection matrix, if a head row population CM-in would be concurrently active as well.

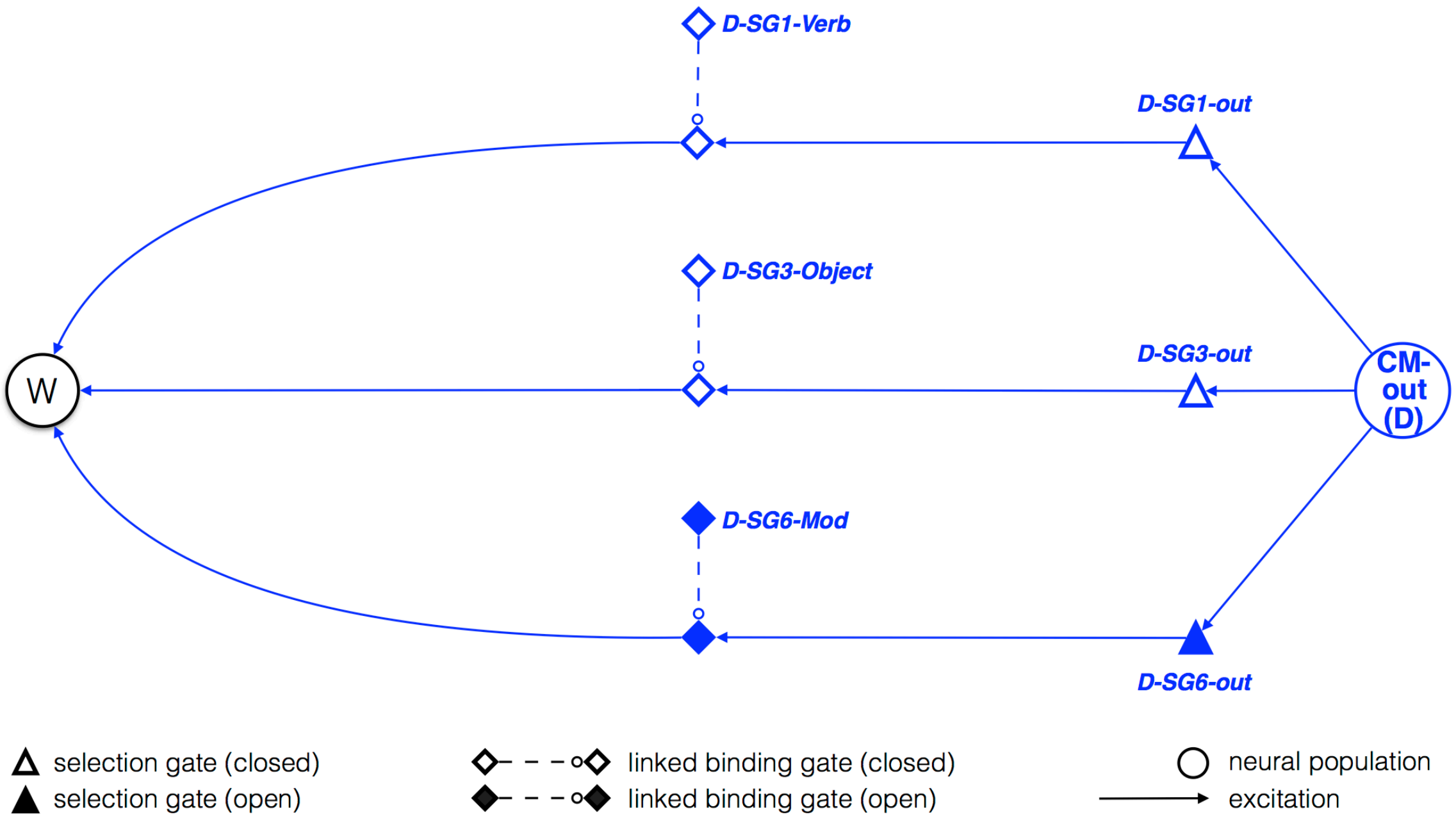

**Figure S8. Dependent and head row (from CM).** Illustration of the connection structure of a dependent row from the connection matrix (CM) to a word population (W). D = Dependent.

Figure S8 illustrates the connection structure from the connection matrix to the word population W in a dependent row. It is similar in head rows, with the label D replaced by H. The population CM-out in the dependent row can be activated by output from the connection matrix. In turn, this population activates SG-selective selection gates in the dependent row. These ensure that the flow of activation out of the connection matrix is controlled and that selective questions can be asked. For example, with *Sue likes pizza*, the query *Sue likes?* will activate V3 in Figure 2. To obtain the answer, the selection gates for SG3-Object in the head and dependent rows need to be opened (section S7.6).

In Figure S8, the selection gate D-SG6-out is opened with the selective signal D-SG6-out.

Again, this selection signal works over all dependent rows, but is it effective only in the dependent row with the active population CM-out.

However, activating SG-selective selection gates is not sufficient for activation to flow from the connection matrix to W. The actual SG binding in the dependent (or head) row also needs to influence this process. This is achieved with linked binding gates. In Figure S7, the selection gate Adj-D (and thus Adj) is bound to the linked binding gate D-SG6-Mod, via one of its binding gates. As a result, the selection gate D-SG6-out in Figure S8 is bound to D-SG6-Mod as well, via its other binding gate. This effectively binds population CM-out in the dependent row to W.

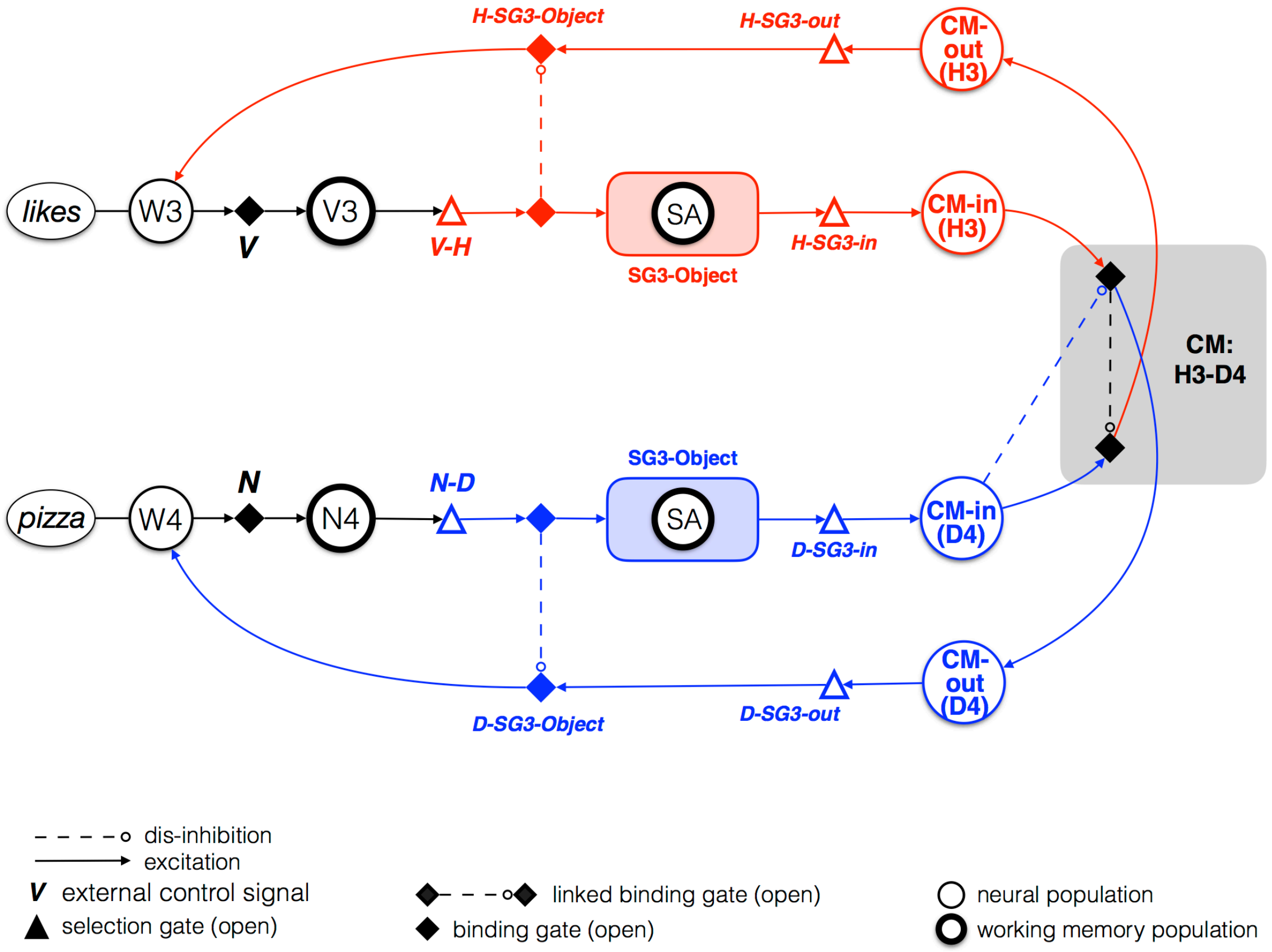

**Figure S9. Binding of *eats pizza*.** Combination of the connection structures in Figures S7 and S8. LMAs and SG circuits (SGs) in Table S11. CM = connection matrix. D = Dependent. H = Head. SA = Subassembly. W = word population.

Figure S9 combines the connection structures in Figures S7 and S8 to illustrate the binding of *likes pizza* in Figure 2. So, *likes* is bound to the population W3 in the third row (row 3). In turn, W3 is bound to the LMA V3, which is bound to SG3-Object in the head row (H3). In row 4, *pizza* is bound to W4, which is bound to the LMA N4. In turn, N4 is bound to SG3-Object in the dependent row (D4). The binding of *likes pizza* is completed in the connection matrix, where the head row H3 is bound to the dependent row D4 in cell H3-D4.

## S4.3. Small world network structure

A number of issues have been raised in the literature on whether architectures such as the NBA of van der Velde and de Kamps (2006) could serve as a full linguistic architecture. By extension, these issues could perhaps also relate to the HD-NBA as presented here. The issues seem to relate to the notion that these architectures depend on a 'fixed' network structure, as described in section 2.1.

One issue concerns the magnitude of the network structure, as given by the set of binding circuits needed for the original NBA (Hadley, 2006; Eliasmith, 2015). In particular, for the binding of words to their LMAs and for the binding of LMAs to each other, in line with their relations. The latter of these binding issues is solved in the HD-NBA with the use of a single connection matrix in a small-world like network structure. As described below, the first one is solved in the HD-NBA (and NBA) with the use of a phonological neural blackboard.

Another issue concerns the flexibility of the network structure. Feldman (2013) argued that it would not be able to handle sentences with a new word, because that word would not be stored as a concept structure yet. As a result, the new word could not bind to an LMA in the sentence neural blackboard. So, assuming that *Sonnet* is a novel name you have just heard, *Sonnet* could not bind to a Noun LMA, which would entail that the HD-NBA could not represent and process sentences like *Sonnet likes pizza*.

This issue is also solved by the small-world like network structure of the architecture, with the inclusion of the phonological neural blackboard. Language has (at least) a two-tier structure (Jackendoff, 2002). First, phonemes and morphemes combine into words and then words combine into sentences (simply stated). Indeed, it would be inconceivable that we could have a 'lexicon' of around 60.000 entries (e.g., Bloom, 2000) if each word consisted of a unique sound pattern, not decomposable into identifiable components. In contrast, with phonemes and morphemes it is conceivable to have a large lexicon, because words themselves are combinatorial structures. In the HD-NBA, these combinatorial structures are represented and processed in the phonological neural blackboard. The sentence neural blackboard gets its input from the phonological neural blackboard, not from the in-situ concept structures directly.

Hence, the phonological and sentence neural blackboards form a small-world like network structure that significantly reduces the magnitude of the network structure of the architecture. The phonological neural blackboard also explains why we can understand novel words in sentence structures. A word may be novel in that it does not have an in-situ concept structure yet, but it will be a regular phonological structure. Research has show that, at birth, babies are equally sensitive to sounds from any human language, but within the first year a specialization occurs in which they become more sensitive to the sound patterns of the language they hear, compared to other languages (e.g., Doupe and Kuhl, 1999). So, our ability to understand new words is flexible only within the language we are familiar with. We do not have that ability for languages we are completely unfamiliar with, and we cannot understand what role such a new word would have in a sentence structure.

The baby research explains why: from early on in our lives we learn our own language. In terms of the HD-NBA that means that we gradually develop the network structures needed for the phonological neural blackboard, the sentence neural blackboard, and their interactions and controls. So, the productivity of language (although unlimited) is confined to that language only. It does not exist for totally unfamiliar languages, precisely because we do not have the neural ('fixed') machinery for that.

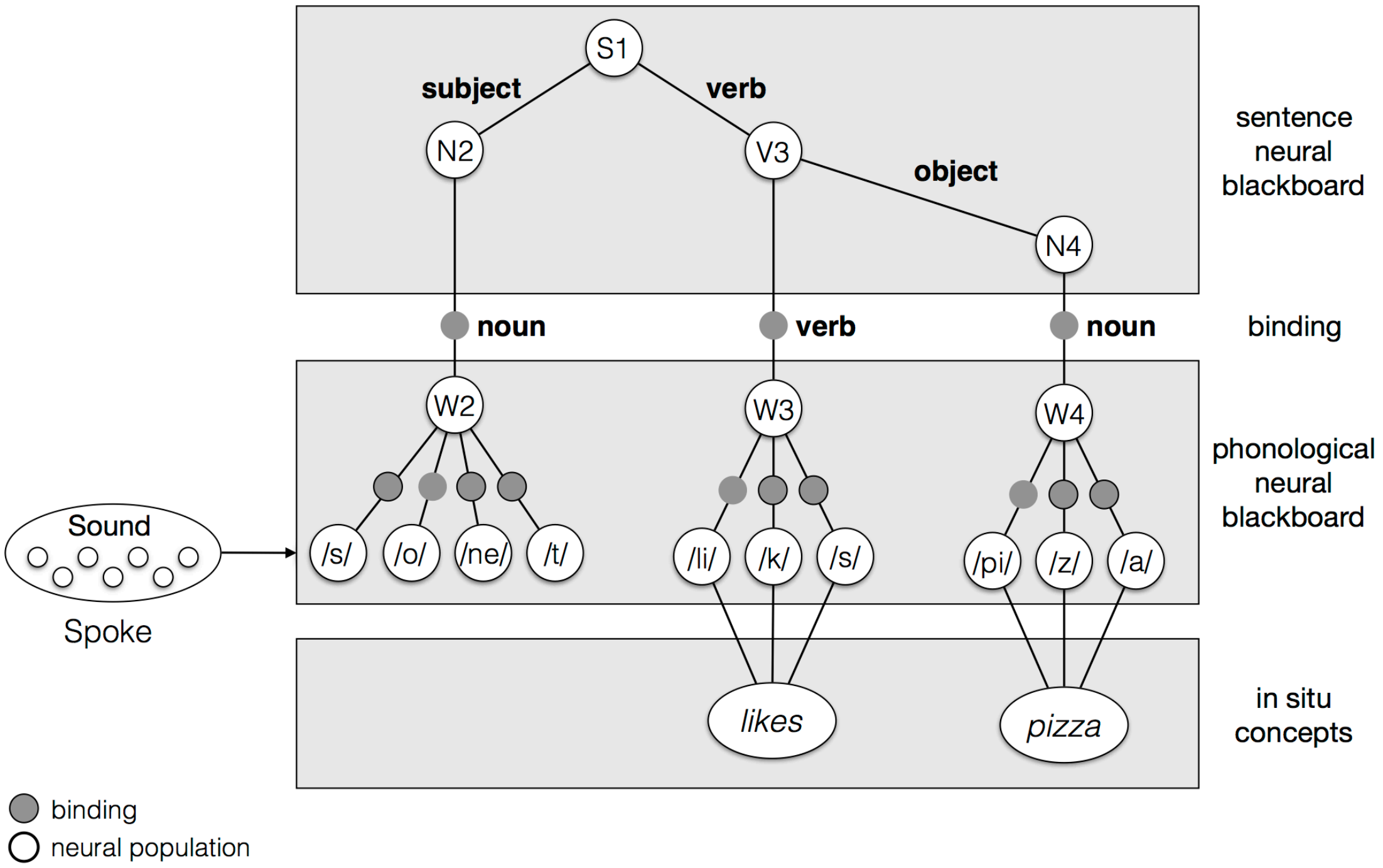

**Figure S10. Small world-like network structure.** Schematic outline of the interaction between the phonological neural blackboard and the sentence neural blackboard in HD-NBA. LMAs in Table S11.

In a language we are familiar with, we can understand novel words in sentences due to the interaction between the phonological neural blackboard and the sentence neural blackboard. Figure S10 shows this in a schematic way for the sentence *Sonnet likes pizza*. The phonological neural blackboard will not be further developed in this article, but it is assumed to be a binding architecture in which the phonemes and morphemes of a word temporarily bind to a word population (W) in line with binding mechanisms outlined here. Within phonology one can also distinguish semi-independent tiers that combine into a word representation (Jackendoff, 2002), but here it is assumed that they bind directly to a word population.

So, when a word is heard, say *pizza*, it activates its in situ concept structure, its lexical features (e.g., noun), and the related phonological structures (phonemes, morphemes) in the phonological neural blackboard. These bind to a word population (W4 in Figure S10). In turn, this population interacts with the sentence neural blackboard, as illustrated in Figure 2. Based on its lexical features, *pizza* is a noun, so W4 will bind to the Noun LMA N4. A similar process binds the verb *likes* to W3 and that to V3.

The word *Sonnet* is different in that it does not have an in-situ concept structure or lexical features yet. But it is a regular phonological structure in a familiar language, which can be decomposed in its phonemes and morphemes, and these can also bind to a word population (W2 in the figure). Here, the sentence context can be used to decide that this word functions as a noun, so that it can bind to the sentence structure as illustrated in the figure (which is an example of typed variable binding in the HD-NBA).

So, in short, the HD-NBA can handle novel words in novel sentence structures, even though it depends on a fixed connection structure. It uses the combinatorial nature of its small-world like network structure, as given by the phonological and sentence neural blackboards, to do that. This mall-world like network structure also solves the size problem for binding in the architecture. Concepts (words in sentences) do not bind directly to the sentence neural blackboard. Instead, their phonological structures do (a similar argument can be made for visual word representations). The phonemes and morphemes in a language form a limited and fixed set, so their binding to word populations does not result from a large set of binding circuits, as assumed by Hadley (2006) and Eliasmith (2015). Arbitrary relations between words and phrases, in line with the structure of the language at hand, can then be encoded in the sentence neural blackboard, as illustrated in the next section.

### S4.4. Arbitrary sentence structures

A third issue for the HD-NBA concerns its ability to deal with a fully developed natural language at the sentence level. Dehaene et al. (2015) argued that an architecture like the HD-NBA (by extension) could not deal with arbitrary sentence structures as found in given language. However, they did not present specific examples to illustrate this. Instead, they just referred to a textbook on linguistics, perhaps with the (implicit) assumption that such a book would contain examples of linguistic structures that the HD-NBA (by extension) would not be able to handle.

Again, the fact that the HD-NBA depends on a 'fixed' connection structure might play a role here, because that would seem to prevent unlimited (structural) productivity. There are two ways to counter this argument. First, by the argument that any fully developed neural architecture for language would consist of a fixed connection structure (section S2). So, either the productivity of language does not exist or it can occur in a fixed connection structure. The second way is to be as comprehensive and precise as possible in showing that the HD-NBA, and hence its fixed connection structure, can represent any grammatically regular sentence structure. Indeed, this is a main challenge for any neural model of language representation and processing.

The aim to show that the HD-NBA can indeed represent any grammatically regular sentence structure motivated the presentation of the SG Tables S1-S10 in section 3.1. The explicit representations of LMA relations in these tables show that the HD-NBA can handle the comprehensive set of sentence structures presented and discussed in *Cambridge Grammar*.

As an illustration, many of the SGs presented in the Tables S1-S10 are combined in the phrase (CG, p. 452 [3]):

*All those grossly over-rewarded financial advisers in the city too*

This phrase combines a number of (linguistic) dependents of the linguistic head *advisers*. Their order is quite rigid, given in *Cambridge Grammar* by: pre-head eternal modifier (*all*) - determiner (*those*) - pre-head internal modifiers (*grossly over-rewarded*) - pre-head complement (*financial*) - head (*advisers*) - post-head internal dependent (*in the city*) - post-head external modifier (*too*).

Figure S11A presents the HD-NBA structure of this phrase, as a dependency structure between the LMAs. This shows the connections between the LMAs bound to the words in the sentence. The LMAs that are dependents in the HD-NBA binding point to the head LMAs they bind with, and the numbers next to the connections are the SGs used the binding.

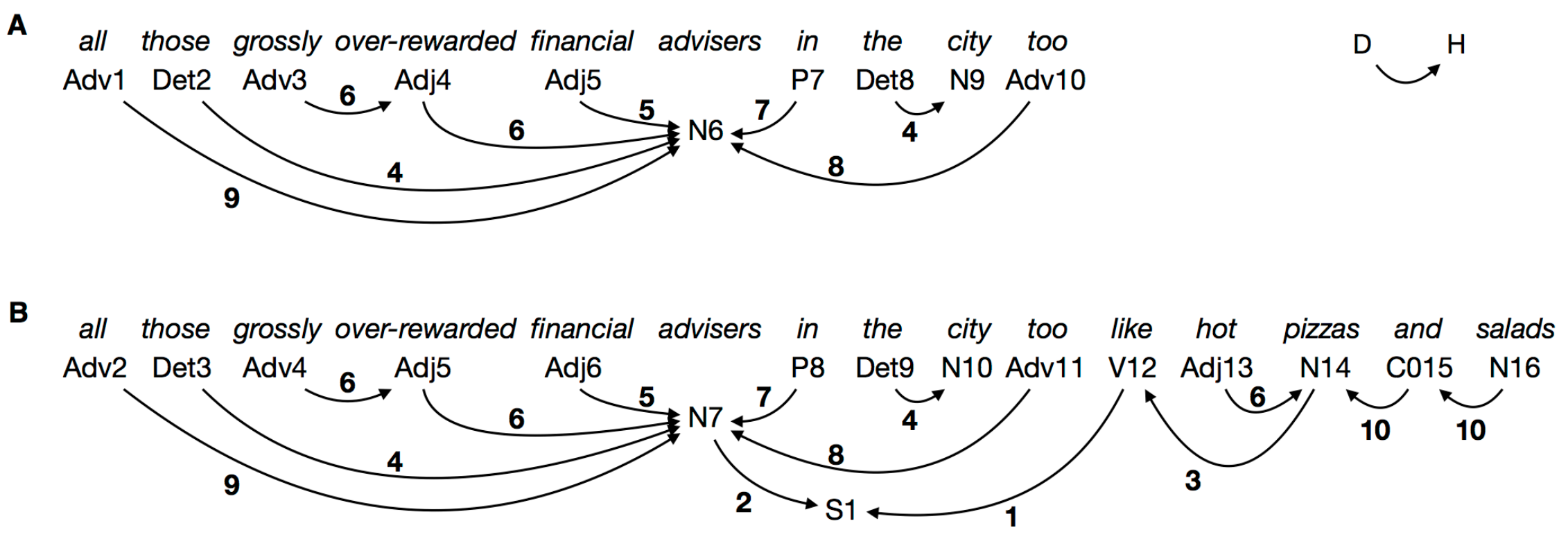

**Figure S11. Arbitrary sentence structures.** (**A**) HD-NBA structure of *All those grossly over-rewarded financial advisers in the city too* (CG p. 452 [3]). (**B**) The HD-NBA structure of the sentence *All those grossly over-rewarded financial advisers in the city too like hot pizzas and salads*. All SGs from Tables S1-S10 are included in this sentence. LMAs in Table S11. Dependents (D) point to heads (H).

In the binding of the phrase, six of the ten SGs in Tables S1-S10 are used. Three of the other SGs are SG1-Verb, SG2-Subject and SG3-Object. They were used in the structure of *Sue likes pizza* illustrated in Figure 2. It is easy to combine these two structures by replacing *Sue* with the phrase above. Also replacing *pizza* with *hot pizzas and salads* combines all SGs in Tables S1-S10 in the sentence *All those grossly over-rewarded financial advisers in the city too like hot pizzas and salads.* The HD-NBA dependency structure of this sentence is given in Figure S11B.

## S5. Connection path as intrinsic sentence structure

The HD-NBA sentence structures as presented in Figure S11 are 'flat' in the sense that dependents are bound directly with heads, regardless of how many dependents there are. This is in line with the 'flat syntax' proposed by Culicover and Jackendoff (2005), but differs from the branching structures found in mainstream grammar. An example of a (right-branching) structure for *even all the preposterous salary from Lloyds that Bill gets* is presented in Figure S12A, adapted from *Cambridge Grammar* (p. 332 [11]). The triangles here indicate phrases that are not further analyzed. Figure S12B illustrates the HD-NBA dependency structure of this phrase. It is remarkably more flat than the branching structure given by *Cambridge Grammar*. The difference can be seen in the red (thick) links in both figures, which correspond to the relation between *even* and *salary* in the sentence structure.

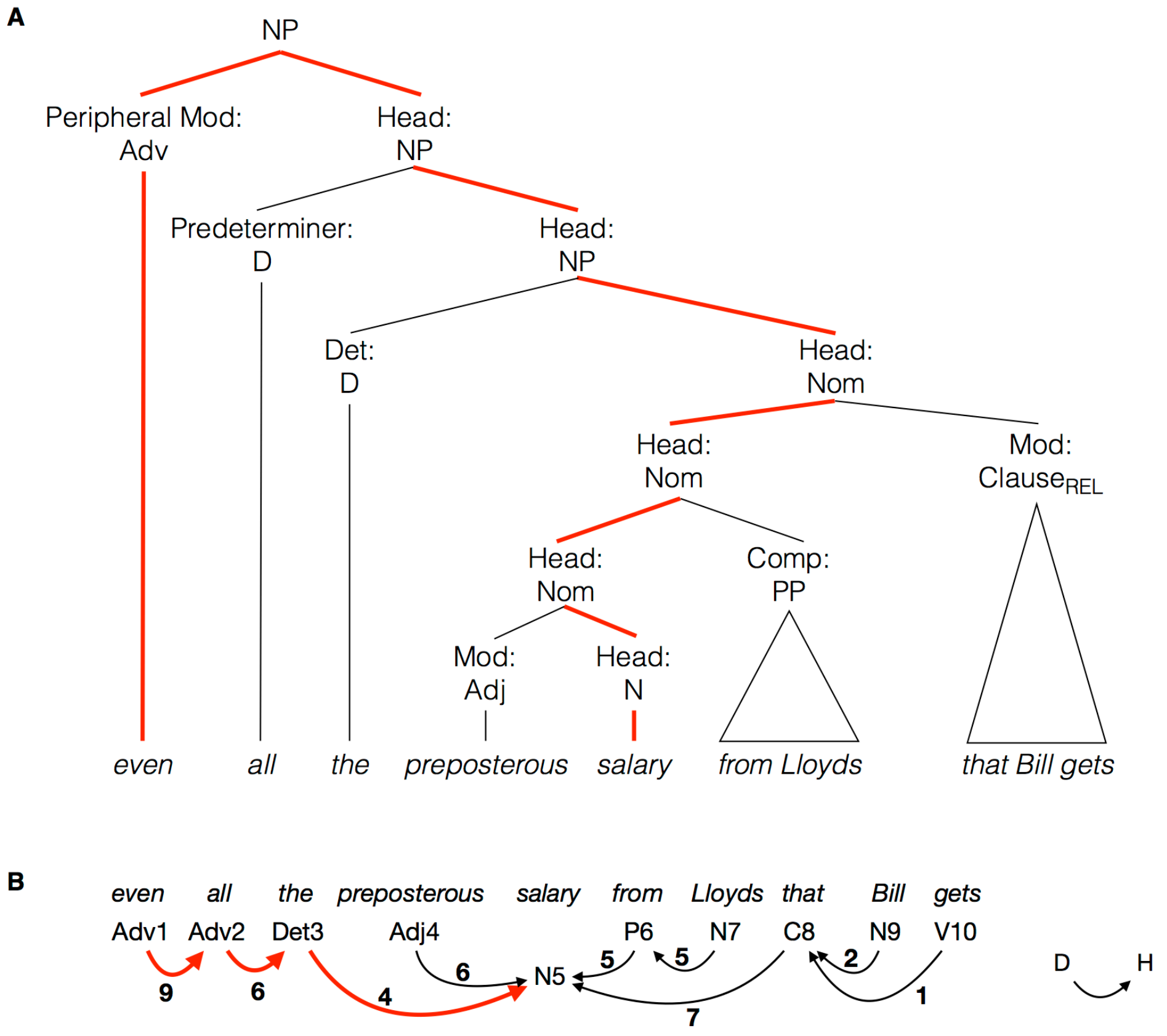

**Figure S12. Intrinsic structure**. (**A**) The right-branching structure of *even all the preposterous salary from Lloyds that Bill gets* (CG p. 332 [11]). (**B**) The HD-NBA dependency structure of the same phrase. LMAs and SGs in Table S11. Dependents (D) point to heads (H). NP = Noun Phrase.

Culicover and Jackendoff (2005) give a number of linguistic arguments against a branching structure as illustrated in Figure S12A. Here, I will give an argument based on the fact that a sentence structure in the HD-NBA is a connection path in the brain. It concerns the notion that this connection path corresponds to the 'intrinsic structure' of a sentence. It has to be noted that this does not refer to the 'deep structure' of a sentence, underlying its 'surface' structure, as found in Transformational Grammars (e.g., see Trask, 1993). There is only one structure of a sentence in the HD-NBA, given by its connection path. The notion 'intrinsic' structure refers to the perspective from which the structure should be developed and analyzed.

The notion of intrinsic structure derives from the work of Gauss on geometry, and curvature in particular. Gauss noted that we could study the curvature of a 2-dimensional (2D) space (such as the surface of a sphere) from a 3D perspective, to arrive at the conclusion that the space is indeed curved. But this approach cannot be extended to the 3D space we live in. So, how could we study the curvature of our 3D space to see if this is curved as well? Gauss realized that we would have to determine that from the 3D perspective itself. In the same way, one should be able to determine the curvature of a 2D space from a 2D perspective, as would be the case for a "society of ants" that lives in that space and is confined to it. Gauss discovered that the curvature of a 2D space could indeed be determined from a 2D perspective, which he referred to as the 'intrinsic' curvature of that space, in contrast with the extrinsic curvature as determined from a higher (3D) perspective. The discovery of intrinsic curvature significantly influenced the development of General Relativity Theory (Misner et al., 1973).

Two aspects stand out from the example of intrinsic curvature. First, the intrinsic curvature of a space is not always the same as its extrinsic curvature. For example, the 2D surface of a cylinder has an extrinsic curvature, but not an intrinsic curvature (i.e., it is flat from the intrinsic perspective).

Second, intrinsic curvature is based on measurements that are entirely confined to the space itself, that is, are confined to local positions in that space (positions that can be occupied by the "ants"). Yet, curvature is a global aspect of a space. So, the procedure of determining intrinsic curvature necessarily depends on a sequence of steps, each based on a local measurement (for examples, see Misner et al., 1973, p. 335-337).

Figure S12 can be used to discuss how the notion of intrinsic structure, and the sequential procedure needed for it, can be applied to sentence structures. Figure S12A gives the (right) branching structure of the phrase *even all the preposterous salary from Lloyds that Bill gets* in the manner of mainstream grammar theory. The branching structure aims to show both the relations and the range or scope of these relations in the phrase. This is indeed clear to see. Basically, the higher it is, the more scope a constituent has. So, *even* is seen to range over the rest of the phrase, whereas *all* ranges over the part *the preposterous salary from Lloyds that Bill gets.* Conversely, if we start with the linguistic head *salary*, the adjective *preposterous* is closer to the head compared to the complement *from Lloyds*. One can argue about these choices. For example, one could argue about whether the phrase expresses (*preposterous salary*) *from Lloyds* (as depicted in Figure S12A) or *preposterous (salary from Lloyds)*. But this argument just concerns the arrangement within the structure. It does not affect the overall idea of representing relations and their scope in this way. Also, it is clear that it is much harder to identify the scope of relations in the HD-NBA structure of the phrase illustrated in Figure S12B.

However, the branching structure of the phrase in Figure S12A is an extrinsic representation. It works because we can look at the overall structure from an outside perspective. This allows us to see what the scope of a relation is in this phrase, by looking at the intermediary layers of representation. But if Figure S12A would be a representation of the phrase representation in the brain (or any other system), how would a "society of ants (neurons)" determine the relations and their scope in this phrase? They do not have the overall view we have of the structure. Instead, they would be able to determine only the intrinsic structure of the phrase. As noted above, this depends on a sequential procedure, based on 'local' observations or measurements.

So, for example, the question is how the relation between the modifier *even* and the rest of the phrase can be established from a perspective within the structure. That is, if you are at, say, the position of *salary*, how could you tell that *even* modifies the rest of the phrase? There is more than one way to do this. For example, you could first go from *salary* to *even* along the red connection path in Figure S12A, to change the within-structure (intrinsic) perspective to the position of *even*, and then follow the connection paths from *even* to the rest of the phrase. But you could also use *salary* as a basis and trace these connection paths from there.

For example, the fact that *even* is a more distant modifier (with wider scope) compared to *preposterous* is reflected in the different connection paths of these words with *salary*. In Figure S12A a word is more distant if the connection path runs through more and 'higher' syntactic nodes, which could be recognized and counted in traversing the path. In Figure S12B the difference between these connection paths is reflected by the different SGs, selection and binding gates they run through, and thus the different control signals that need to be activated to open the connections. The point is, however, that in both cases the relations between words and the difference between these relations reside in the sequence of steps that have to be taken to go from one word (say *salary*) to the other words. In that sense, the intrinsic structure of a sentence, given by the (more) flat structure in Figure S12B, captures the same relations in a more direct way compared to the branching structure of Figure S12A, which looks clear from an extrinsic perspective, but very convoluted and overladen from an intrinsic one.

Another illustration of intrinsic sentence structure is given by the predeterminer *all*. In Figure S12A it stands between *even* and *the* in the branching structure. In Figure S12B it is connected as a modifier to the determiner *the* (which it seems to modify). Nevertheless, the notion that *all* refers to *the preposterous salary from Lloyds that Bill gets*, whereas *even* refers to *all the preposterous salary from Lloyds that Bill gets*, is reflected in the different connection paths between these words (*even, all*) and *salary* in Figure S12B.

### S5.1. Intrinsic structure and scope

Intrinsic structure is also related to the issue of scope. For example, *great white whale* could refer to a whale that is both great and white. But it could also refer to a white whale that is great. In this case, *great* has scope over *white whale*. The difference between these two interpretations can be illustrated with phrase structures. Figure S13A presents a phrase structure of *great white whale* in which both *great* and *white* are direct modifiers of the noun. This structure is 'flat' because both *great* and *white* are directly connected to the NP (Noun Phrase) node of *whale*. Figure S13B shows the HD-NBA dependency structure of the phrase in (A). Figure S13C shows a hierarchical phrase structure to represent the scope of *great* over *white whale*. This scope is clear because *great* is connected to the highest NP node, which is connected tot the NP node for *white whale*.

However, Figure S13C is clearly an extrinsic structure, because we use the outside perspective to see that *great* is connected to the highest NP node. As noted above, the intrinsic structure of this phrase would require a sequential process based on 'local observations'. Here, we could start at *great*, move upwards to the highest node, and observe that it is connected to a lower NP node (e.g., in terms of head vs dependent relations). Then, we could move to *white* and *whale* to establish that *great* dominates over *white whale*.

Figure S13D presents two HD-NBA dependency structures that would provide a connection path (intrinsic structure) in which *great* has scope over *white whale*. In each, *great* is connected directly to *white*, which is connected to *whale*. So, a sequential procedure starting from *great* would have to go via *white*, which would establish the scope relation between them. This is more direct than in the phrase structure in (C) because in that you could go directly from *great* to *whale* via the two NP nodes.

The connection path that establishes the scope of *great* over *white whale* would have to be different from a connection path in which an adverb like *pale* modifies *white* and *white* modifies *whale*. The difference would derive from the different LMAs and SGs involved. In (D) there is a direct connection between Adj1 and Adj2, instead of between Adv1 and Adj1, as in (E). The connection Adj-Adj is not found in Tables S1-S10. But it could be established with SG-GAP with either Adj2 or Adj1 as head, as illustrated in (D). Other possibilities of connection path differences using other SGs, in relation with the LMAs involved, could be explored as well.

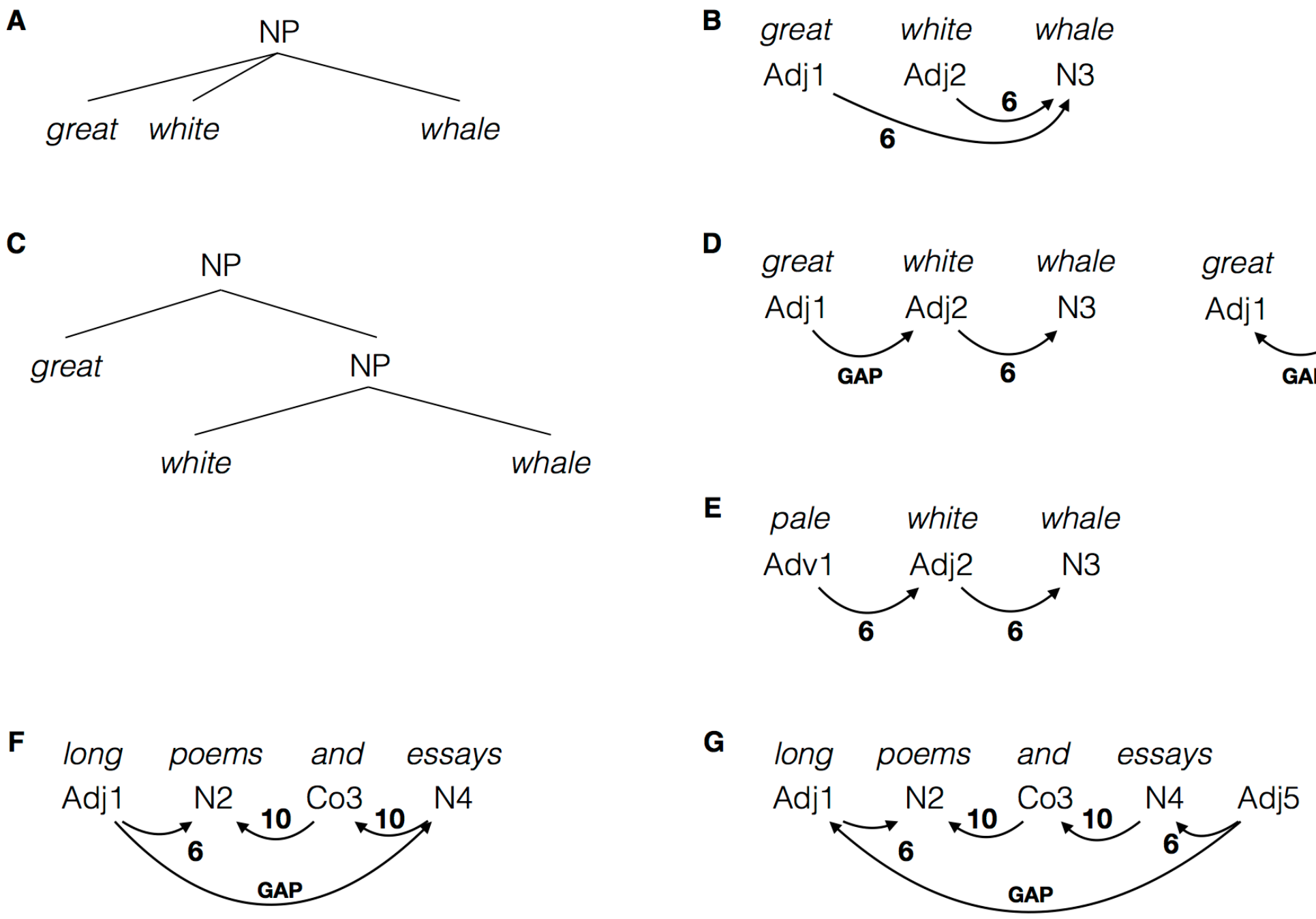

**Figure S13. Intrinsic structure and scope**. (**A**) Phrase structure of *great white whale.* (**B**) HD-NBA dependency structure of (A). (**C**) Phrase structure of *great (white whale).* (**D**) Two possible HD-NBA dependency structures of (C). (**E**) HD-NBA dependency structure of *pale white whale*. (**F**) One possible HD-NBA dependency structure of *long (poems and essays).* (**G**) Another possible HD-NBA dependency structure of *long (poems and essays).* LMAs and SGs in Table S11. Dependents (D) point to heads (H). NP = Noun Phrase.

Figures S13F and S13G show two HD-NBA dependency structures of *long poems and essays*. This phrase is ambiguous and could either mean *(long poems) and essays* or *long (poems and essays)*, in which essays are also long. The HD-NBA structure of *(long poems) and essays* is the same as that of *(hot pizzas) and salads* in Figure S11B. So, the focus here is on *long (poems and essays)*. In (F) *long* is connected to *essays* via SG-GAP. This combination could reflect that *long* is also a modifier of *essays*. In (G) Adj1-*long* is connected to Adj2 via SG-GAP as head, and Adj2 is a modifier of *essays.* Here, there is no word bound to Adj2.

The possible HD-NBA structures in (F) and (G), and in (D), illustrate that intrinsic structures given by connection paths can be related to issues of scope. As noted, other possibilities could be explored as well. Future research would have to decide between the illustrated and other options.

### S5.2. Intrinsic structure and inflection

Another issue that is related to intrinsic structure is inflection. As used here, inflection stands for differences in tense or agreement (which was discussed already). Tense is typically expressed with verbs, as in the difference between *Sue likes pizza* versus *Sue liked pizza.* In some grammatical theories (e.g. in 'Government and Binding', see van Valin, 2001) tense is expressed with a specific grammatical node in the phrase structure of a sentence. But that is an extrinsic structure.

In the HD-NBA, tense is given directly by the use of in-situ concept structures. As illustrated in Figure 2, a sentence structure in the HD-NBA is directly connected to the connection structures of the in-situ concept structures underlying the words in the sentence. The in-situ structure of *likes* will reflect that it is in the present, that of *liked* will reflect that it is in the past. Hence, there is no need to reflect this difference in the dependency structure at the level of the sentence neural blackboard. Instead, tense is given by the overall connection structure of the sentence. To establish tense in the intrinsic structure of the sentence, the sequential procedure will go from the sentence structure in the blackboard to the connection structure of the verb (concept structure) involved

## S6. Parsing: Control of binding in the connection structure

The HD-NBA aims to combine a form of incremental processing (e.g., Marslen-Wilson, 1973; Tanenhaus et al., 1995) with (limited) forms of backward processing without explicit backtracking. Incremental processing in terms of the HD-NBA entails that at any moment only one word (referred to as the 'current' word here) enters processing in the sentence neural blackboard, but the next word could be processed already in the phonological neural blackboard.

Hence, when the current word enters the sentence neural blackboard, its processing could be determined by the information available up to the identification of the next word. Later on, however, information beyond the next word could also influence the role of the current word in the sentence structure (without using forms of explicit backtracking). This allows the HD-NBA to process and resolve ambiguities in the sentence structure (up to a point) even though the words in the sentence are processed in a forward-like manner (section S8).

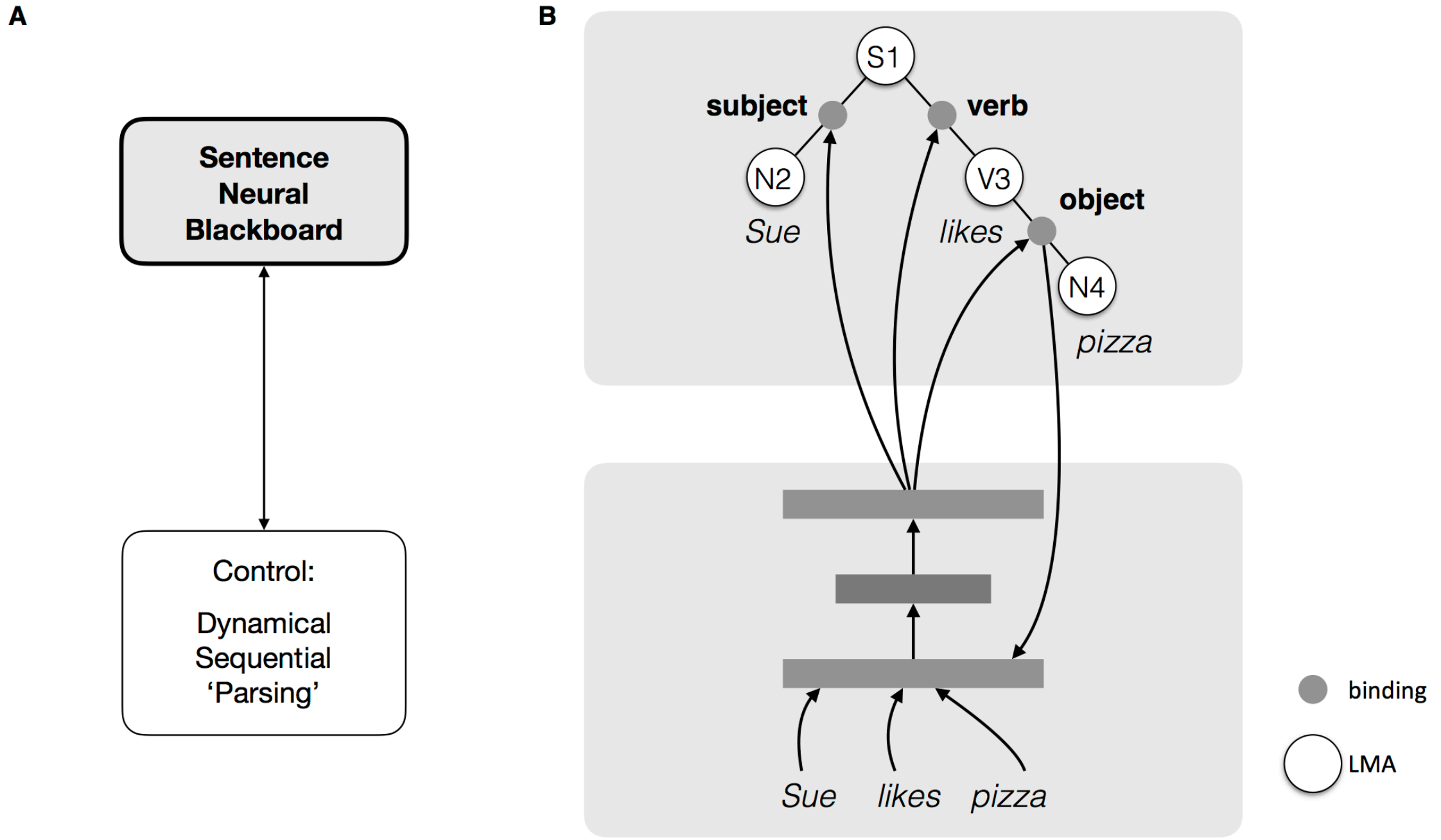

**Figure S14. Control of binding.** (**A**) Sources of processing control in the sentence neural blackboard. (**B**) Illustration of binding control (parsing) in the HD-NBA with a feed-forward network. Based on van der Velde and de Kamps (2010). LMAs in Table S11.

Figure S14A illustrates the sources of processing control over the sentence neural blackboard. Dynamical control relates to issues like the timing of activation or inhibition of LMAs in the blackboard during processing, and sequential control relates to the sequence of steps needed in processing sentence structures or retrieving information. These will be illustrated in particular in the process of answering questions (section S7.6). Parsing control in the HD-NBA is closely related to forms of dependency parsing (e.g., Nivre, 2008). It relates to the control of the binding process in the sentence neural blackboard.

Binding in the HD-NBA is controlled by the activation of selection gates and the initial activation of binding gates. These are based on specific control (selection and binding) signals. They operate over the entire blackboard, but are effective in those rows in which related LMAs or SGs are active. This interaction between control over the entire blackboard and row-based activation determines which LMAs will bind to the words in the sentence, and in which specific role (that is, as head or dependent and with which specific SG) they bind to each other in the connection matrix. In this way, a specific connection path of a sentence structure is formed in the architecture (Figure 2) even though control operates globally over the entire sentence blackboard.

This process is incremental because there is only one active word population in the sentence neural blackboard during processing, representing the current word. When the next word in the sentence appears, the word population bound to the current word needs to bind with an LMA, so that a new word population could become active in the sentence neural blackboard. However, the lexical type of the next word could already have been identified, so it could influence the (type of) binding of the current word in the sentence neural blackboard.

Figure S14B illustrates the binding (parsing) process investigated in van der Velde and de Kamps (2010). Here, a feed-forward network is used to control the binding process. With the sentence *Sue likes pizza*, the words are presented sequentially to the network, using their lexical type (noun, verb) as input. Also, input from the neural blackboard (e.g., expected bindings) is given as input. The output consists of binding signals that activate the binding gates related to the roles of the words in the sentence. So, with *Sue* as the first word in the sentence, the network activates the binding of N2-*Sue* to S1 as subject. It then activates the binding of V3-*like* to S1 as verb and anticipates the binding of an object, resulting in the binding of N4-*pizza* as object of the verb when *pizza* is presented to the network.

The network was trained with noun-verb-noun sentences and sentences with relative or complement clauses. The network achieved the correct bindings in all training sentences and also in sentences derived from combinations of the learned sentences.

Andor et al. (2016) also trained a feed-forward network ("Parsey McParseface") to operate as a dependency parser. Their network was trained on a much larger set of sentences, using global information related to the entire sentence structure. It achieved high success rates of dependency parsing on a wide rage of sentence input.

So, it seems that a feed-forward network could be used to control binding operations in the HD-NBA. This illustrates the integration of 'classification' and 'combination' in which classifier networks regulate the binding process in the HD-NBA, and the neural blackboards of the architecture allow the representations of arbitrary sentence structures. Future research is needed to further investigate this integration. In particular, whether anticipation based on neural blackboard activation could provide the more global information needed for sentence parsing.

Future research is also needed to investigate the influence of semantic information on the binding process, as illustrated with the sentences (1a) and (1b) in section S2.

## S7. Simulations

This section provides additional information on the simulations presented in the main article, and presents additional supporting simulations.

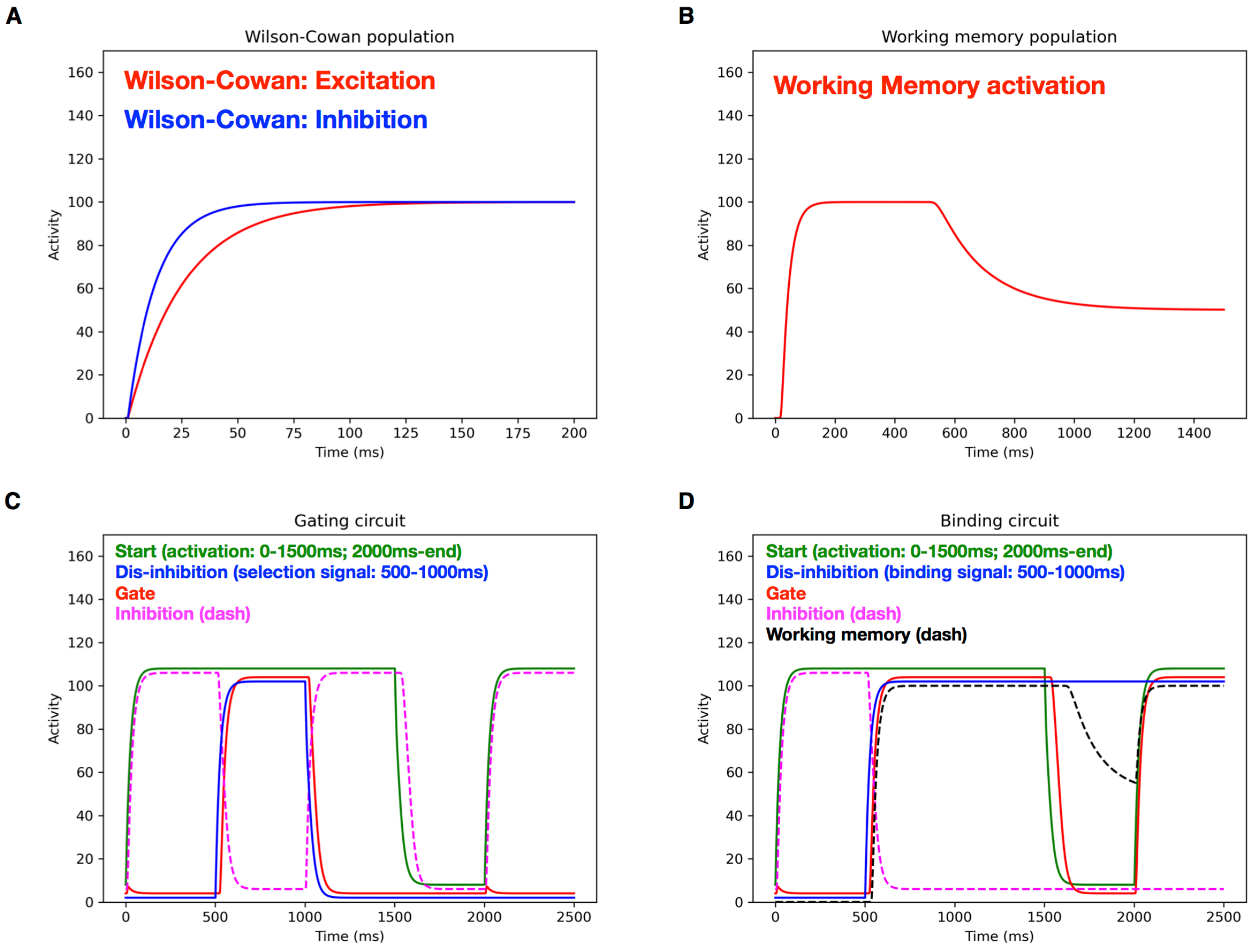

**Figure S15. Population dynamics**. (**A**) Simulation of a single Wilson-Cowan population. (**B**) Simulation of a working memory population. (**C**) Simulation of a gating circuit. (**D**) Simulation of a binding circuit. Activation in spikes/ms.

The basis element of simulation in the HD-NBA is the average behavior of a population, seen as a local group of interacting excitatory and inhibitory neurons that operates as a unity. All circuits in the architecture consist of such populations or other circuits based on them. The choice for a population as the basis element is motivated by two considerations. First, the behavior of the architecture will not depend on the fate of a single neuron (e.g., as given by the randomness of its behavior or the chance of decay). Secondly, populations can respond faster to input activation than individual neurons, which are influenced by, e.g., their refractory period. In a population, there will always be neurons that are ready to respond directly to input activation.

All simulations are based on Wilson and Cowan (1972, 2021) population dynamics (as used in van der Velde and de Kamps, 2006). Wilson-Cowan populations consist of two interacting sub-populations, one excitatory and one inhibitory. Figure S15A illustrates the dynamics of the Wilson-Cowan population used for all populations in the architecture. Activation is calculated in steps of one millisecond (ms). The maximum activation level (100 spikes/ms) is set as a parameter. In Figure S15A, input (100 spikes/ms) starts at time = 0 and remains constant. The figure shows that inhibition in the Wilson-Cowan population operates faster than excitation.

Input to a Wilson-Cowan population in the architecture is always presented to both of its sub-populations. The output of an excitatory population in the architecture is given by the excitatory Wilson-Cowan sub-population. The output of an inhibitory population in the architecture is given by the inhibitory Wilson-Cowan sub-population. This ensures that inhibition operates faster than excitation in the architecture.

### S7.1. Circuit dynamics

Figure S15B shows the activation of the A population in a working memory circuit (section S4.1). Input (100 spikes/ms) is given from 0 to 400 (ms). After that, the activation of the working memory population first declines but then remains stable as sustained activity.

Figure S15C shows the activation of the populations in the gating circuit (section S4.1). Input (100 spikes/ms) is given in the manner as presented in the figure. The initial input activates the Start (green) population, which in turn activates the Inhibition (magenta) population. The activation of the Dis-inhibition (blue) population (by the selection signal) then inhibits the Inhibition population. This allows Start to activate the Gate (red) population. At this moment the gate is open. It is closed again when the selection signal stops so that Dis-inhibition is no longer active. The re-activation of the Start population at 2000 ms does not activate Gate.

Figure S15D shows the activation of the populations in the binding circuit (section S4.1). Input (100 spikes/ms) is given in the manner as presented in the figure. The initial input activates the Start population, which in turn activates the Inhibition population. The activation of the Dis-inhibition population (by the binding signal) then inhibits the Inhibition population, which results in the activation of the Gate population. This activates the working memory population, which ensures that the Dis-inhibition population remains active, even when the binding signal is stopped. Now, the re-activation of the Start population does activate the Start population. Hence, the populations connected by the binding circuit are bound.

### S7.2. Structures for competitions in the architecture

Figure S16 illustrates a dependent row as presented in Figure S7 in more detail. In particular, the shaded area represents the circuit of a Structure Group (SG), here SG1-Verb. This circuit is similar for all SGs in all head and dependent rows.

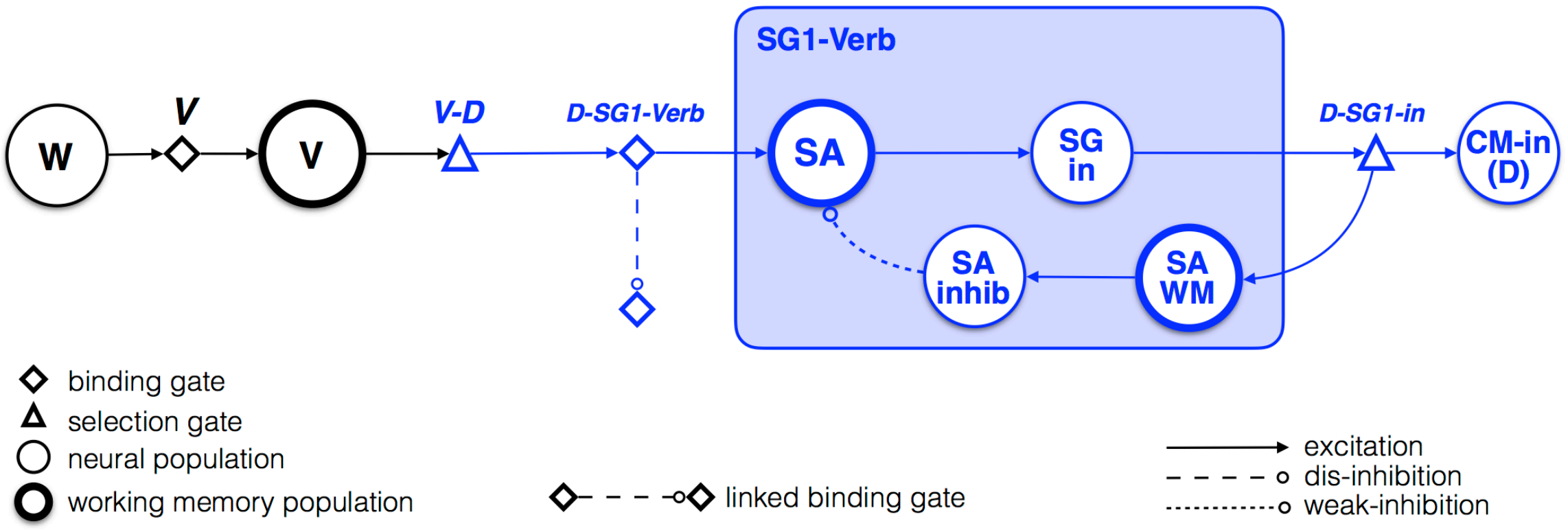

**Figure S16. Circuit of a Structure Group (SG)**. LMAs and SGs listed in Table S11. SA = Sub-assembly. W = Word population.

Here, a Verb LMA binds to the linked binding gate D-SG1-Verb, which then activates the sub-assembly SA in the circuit for SG1-Verb. In turn, this activates the population SG-in. This population plays an important role in the competition between dependent rows, as discussed below. The population SG-in provides the input for the selection gate D-SG1-in, illustrated in Figure S7.

When the selection gate D-SG1-in is opened, its output also activates the working memory population SA-WM. Its activity indicates that a binding was initiated in the connection matrix. In turn, it activates the inhibitory population SA5-inhib, which inhibits the sub-assembly SA with 'weak' inhibition. This ensures that SA is inhibited when it is active on its own, but not when it receives activation from the LMA it is bound to (V). So, once a binding is achieved, binding and competition activity (see below) stops when SG1-Verb receives no activation (to prevent interference with other bindings needed in the connection matrix). But activation can still flow from the active LMA to the connection matrix, as needed to answer questions (section 7.6).

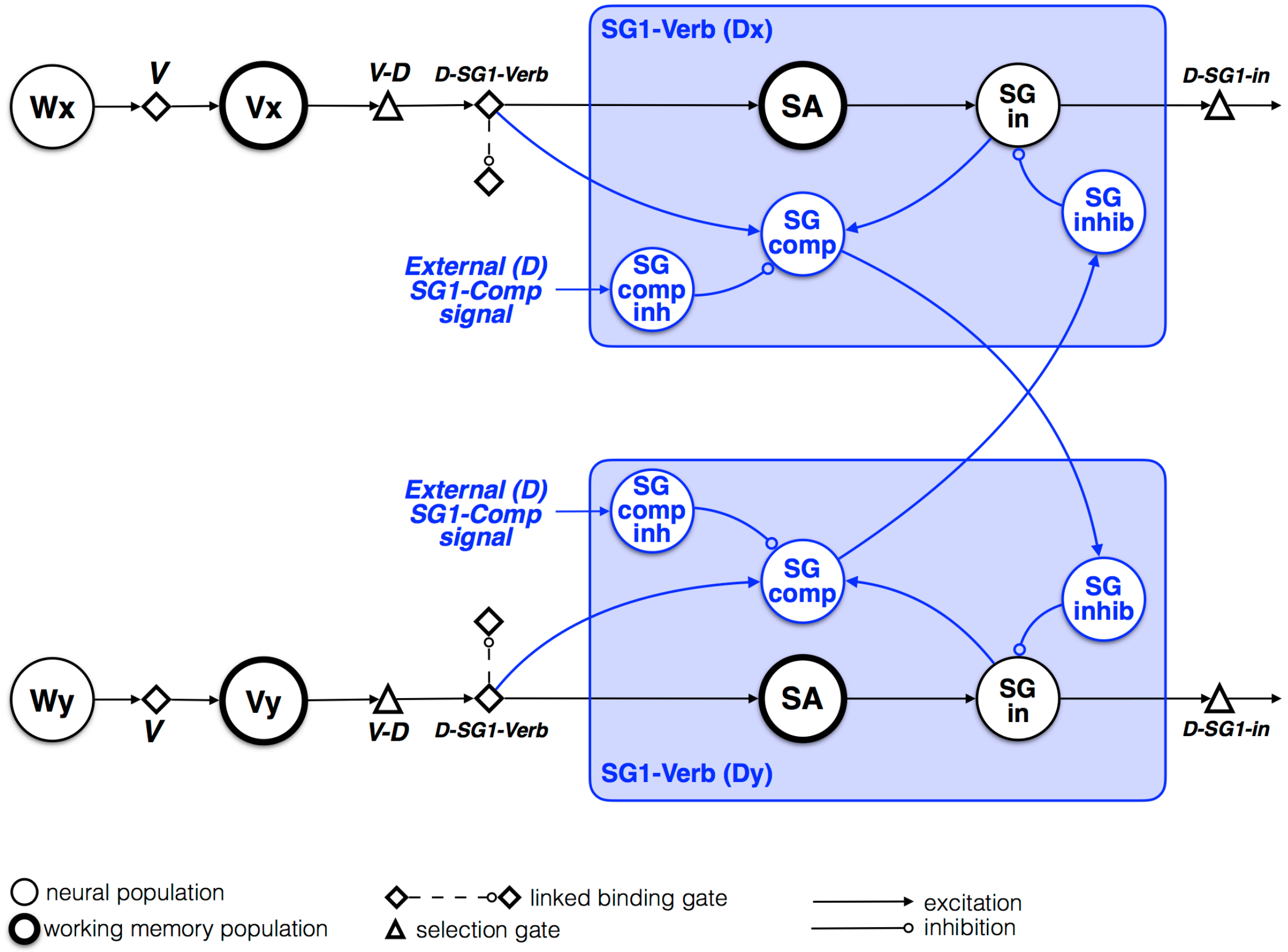

**Figure S17. Competition between same SG in different rows.** The structure derived from previous figures is black. New populations and connections are blue. LMAs and SGs listed in Table S11. SA = Subassembly. W = Word population.

Figure S17 illustrates the competition between the same SGs in different dependent rows, here illustrated with SG1-Verb in Dx and Dy. The structure is similar for all SG circuits in the dependent rows and in the head rows. This allows the HD-NBA to postpone the binding of LMAs in the connection matrix. For example, the binding of the first verb, say the verb in a relative clause, could be postponed until the binding of a second verb (the verb of the main sentence) has been achieved. This illustrates how the structure of the sentence neural blackboard can influence sentence processing.

The competition is initiated with population SG-comp, which is activated by the linked binding gate D-SG1-Verb and the population SG-in of the SG1-Verb circuit in the same dependent row. The SG-comp population of the SG1-Verb circuit in one dependent row inhibits the SG-in populations of the SG1-Verb circuits in all other dependent rows. This blocks their entry to the connection matrix. The inhibition is achieved with the local inhibitory population SG-inhib, which ensures that inhibition is always local in the HD-NBA.

The competition favours the last activated LMA of the same type, because all LMAs of the same type are first inhibited when a new LMA of the same type is activated. Hence, this is the only active LMA during the competition. In this way, the competition represents a stack-like operation in which the last activated LMA (here Vy) is the first to bind. After this binding, the subassembly SA in its SG circuit is inhibited (Figure S16). As a result, the previously activated LMA can bind. It will depend on the neural dynamics of sustained activity to what extent this stack-like operation can be reliably applied when more LMAs of the same type have been activated earlier. This will be investigated in future research.

In retrieving information from the sentence blackboard, as in answering questions, the competition between same SGs can be blocked by an external signal. This operates via the local inhibitory population SG-comp-inh in each SG1-Verb circuit. This also illustrates that the HD-NBA can be in different 'modes' depending on the task at hand, as here processing a sentence structure versus retrieving information.

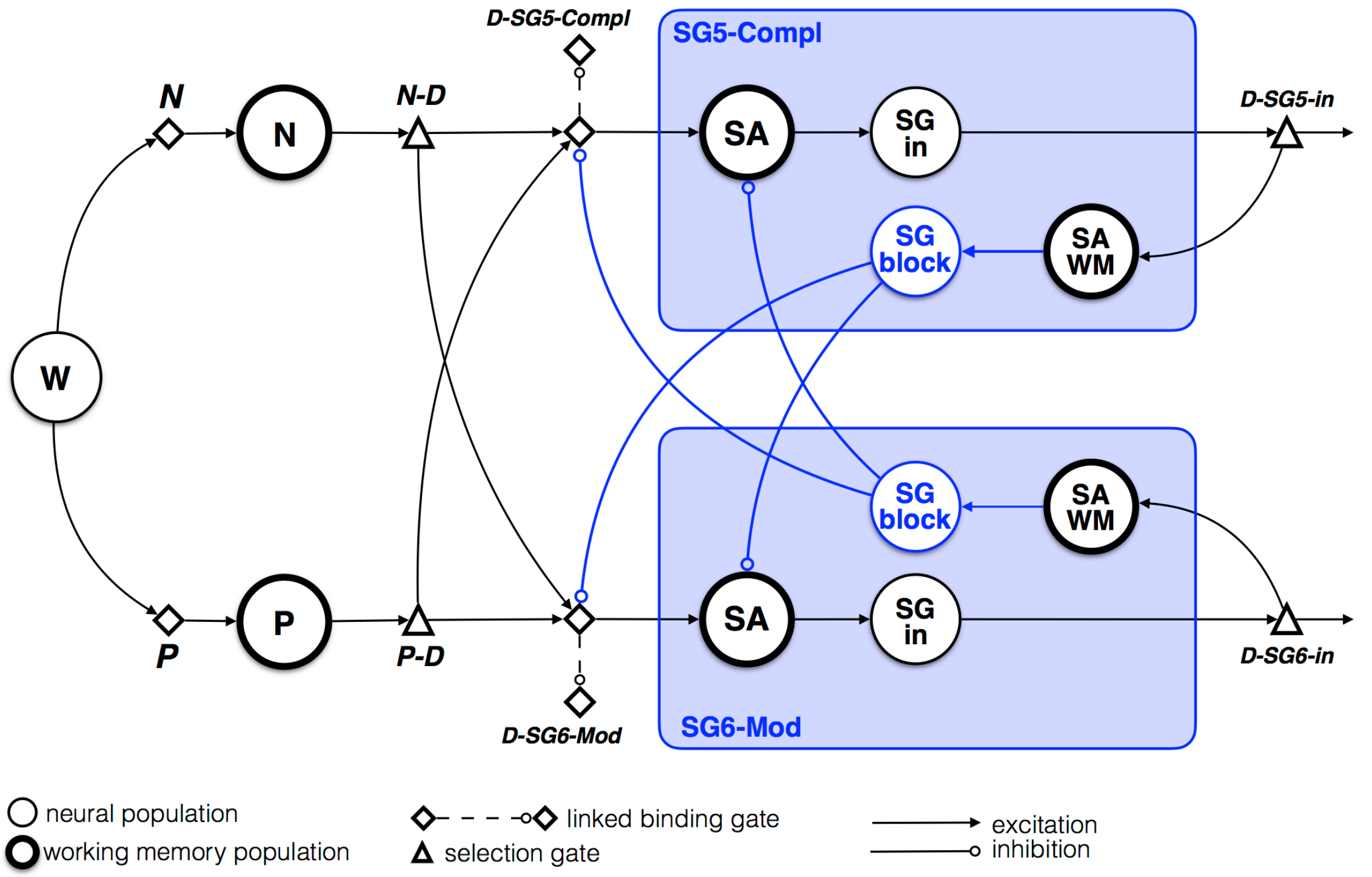

**Figure S18. Competition between different SGs in same dependent row.** The structure derived from previous figures is black. New populations and connections are blue. LMA and SG listed in Table S11. SA = Subassembly.

As illustrated in Figure 2, an LMA can bind with more than one SG in a head row, but with only one SG in a dependent row (with the exception of SG11-GAP). Yet, during the processing of a sentence it could be useful to bind with other SGs in the dependent row as well, in anticipation of the selection of the correct binding. Then, when a binding is achieved, the other LMA-SG bindings in the dependent row should be eliminated, to prevent them from interfering with the ongoing processing of the sentence or its reactivation in retrieving information.

The structure illustrated in Figure S18 allows this to happen. Multiple SGs can initially bind with the active LMA in the same dependent row. For example, LMA N can bind with both D-SG5-Compl and D-SG6-Mod. Assume D-SG5-Compl is then selected for binding in the connection matrix, by activating the selection gate D-SG5-in. This results in the activation of population SA-WM in the circuit D-SG5-Compl, which activates the inhibitory population SG-block. This inhibits each subassembly (SA) in each of the other SG circuits in the same dependent row (except in SG11-GAP), as illustrated with the SA in SG6-Mod. It also inhibits the working memory populations in the linked binding gates for these other SGs, as illustrated with D-SG6-Mod. Hence, this inhibition eliminates these other LMA-SG bindings in the same dependent row.

The structure illustrated in Figure S18 can be (and is) applied in the dependent rows but not in the head rows, because multiple LMA-SG bindings are possible in the head rows. This shows that the selectivity of binding in the connection matrix primarily depends on the dependent rows.

Figure S18 also illustrates that the HD-NBA is capable of (limited) forms of backward processing (without explicit backtracking). For example, LMA N could initially bind only with D-SG5-Compl. But when more information about the sentence structure becomes available, it could then bind with another SG in that dependent row, such as D-SG6-Mod, provided that the LMA N is still active. This also illustrates the role of working memory populations such as LMA and subassemblies in the HD-NBA. Examples of this form of backward processing are discussed in section S8.

### S7.3. Variable size of the architecture

The capacity of the HD-NBA to store sentences depends on the number of rows, as illustrated in Figure 2. Because each row contains only one word population (W), the capacity can be expressed in terms of the number of W populations in the architecture. In all simulations in the main article, four sizes of the architecture were used, with 15, 20, 25 and 30 rows. They are referred to as architectures W15, W20, W25 and W30 respectively.

| Source | W15 | W20 | W25 | W30 |
|---|---|---|---|---|
| Gating circuit | 8880 | 12640 | 16800 | 21360 |
| Working memory | 3030 | 4240 | 5550 | 6960 |
| Other | 2055 | 2740 | 3425 | 4110 |
| *Total neural blackboard* | *13965* | *19620* | *25775* | *32430* |
| External | 144 | 149 | 154 | 159 |
| Total in simulation | 14109 | 19769 | 25929 | 32589 |
| Wilson-Cowan | 28218 | 39538 | 51858 | 65178 |

**Table S12: Numbers of populations used in simulations.** W = word population.

Table S12 presents the number of populations in each of these architectures, divided over different categories. The top row gives the number of populations in all gating circuits (selection and binding) in the architectures. They play a crucial role in the processing of a sentence. The second row gives the number of populations in all working memory circuits (populations). They are crucial for storing a sentence structure in the architectures. The third row gives the number of all other populations in the architectures. They consist of populations such as word populations or the populations for input to and output from the connection matrix. The fourth row gives the number of total populations in the neural blackboards of the architectures, which is the sum of the first three rows.

The fifth row in Table S12 gives the number of external populations. These are the populations that provide the input to the architectures, such as the activation of the word populations and the activation of selection and binding signals. The sum of row four and five gives the total number of populations in the simulations for each of the architectures (row six of the table). However, each population in a simulation is a Wilson-Cowan population, which consists of two interacting (sub)-populations (as illustrated in Figure S15A). The number of Wilson-Cowan populations simulated for each of the architectures is given in the bottom row of the table.

### S7.4. Simulated sentence structures

In the simulation of each phrase or sentence in the main article, each word is presented for 300 milliseconds by activating its word population during that time (indicated by the small black vertical bars in figures of sentence related activation). The gating circuits (selection and binding) related to the role of a word (LMA) in the sentence structure are then activated (opened) and deactivated (closed), in relation with the sentence structures at hand. In this way, sentence structures are processed and stored in the HD-NBA.

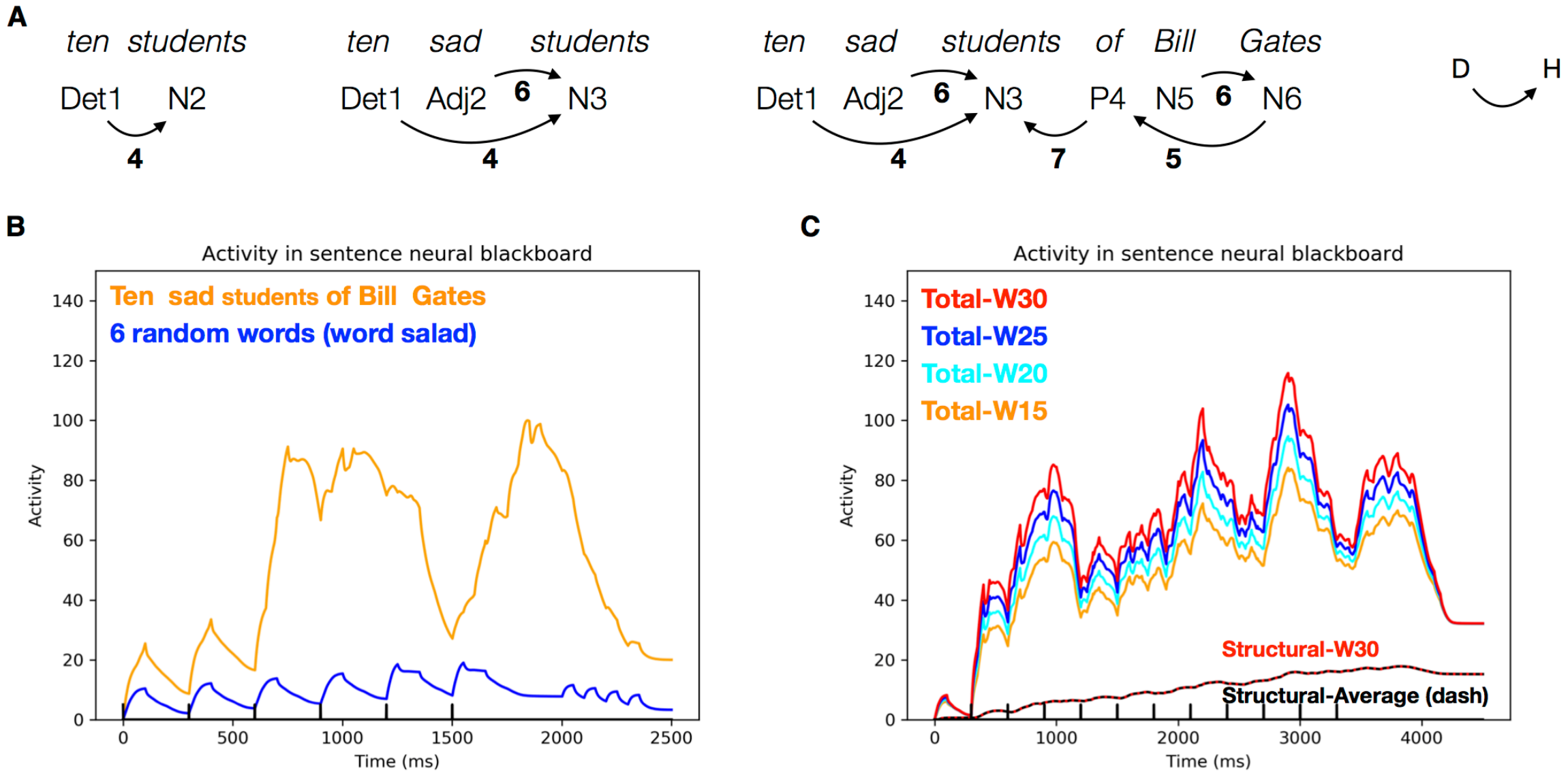

**Figure S19. Additional information for Figures 3 and 4** (**A**) HD-NBA structures for the simulated phrases from Nelson et al. (2017). LMAs and SGs in Table S11. Dependents (D) point to heads (H). (**B)** Comparison between the simulation of a phrase structure and a word salad. (**C**) Simulations (total activity) of the sentence *I wonder how many miles I have fallen by this time* with W15, W20, W25 and W30. Connection path (structural) activity for W30 and average over W15-W30.

Figure S19A gives the HD-NBA sentence structures of the phrase simulations based on Nelson et al. (2017), presented in Figure 3. Figure S19B presents an HD-NBA simulation of another observation in Nelson et al. (2017). They concluded that brain activity increases for sentence length, based on the structure of the sentence and not just on the number of words. That is, longer sentences (in general) have a more complex structure, which results in increased brain activity during sentence processing.

Figure S19B shows the total activity averaged over W15, W20, W25 and W30 of the six word phrase *Ten sad students of Bill Gates* from Nelson et al. (2017), and the average activity obtained with a six word 'salad', i.e., with six randomly chosen words (LMAs). The figure shows that activity in the HD-NBA to a large extent indeed depends on the structure of a phrase and not just on the number of its words. An unstructured sequence of six words produces far less activity than a structured phrase of six words.

Figure S19C presents four simulations of the sentence *I wonder how many miles I have fallen by this time* from Brennan et al. (2016). The HD-NBA structure of the sentence is presented in Figure 4B. Figure S19C shows the total activation in architectures W15, W20, W25 and W30. It also shows the average activity in these architectures and the activity in W30 of the working memory populations (labelled 'structural' activity in the figure).

Figure S19C shows that the total activity increases with increased size of the architectures, as also illustrated in Figure 4C. But the structural activity given by the working memory populations does not increase (with the exception analysed in the next section). This underlines the fact that most activity generated in the architect is related to the control of creating the sentence structure, as given by the selection and binding gates. The control signals of these gates operate over the entire architecture, hence the activity they generate increases with increased size of the architectures (as analysed in the next section). In contrast, the active working memory populations are related to the structure of the sentence, which does not change by increasing the size of the architecture.

Figure S19C also shows that the profile of the total activity is similar for all architectures. Hence, the relation with the profile of 'syntactic effort' produced by a Minimalist Grammar, illustrated in Figure 4D, remains the same for all sizes of the HD-NBA.

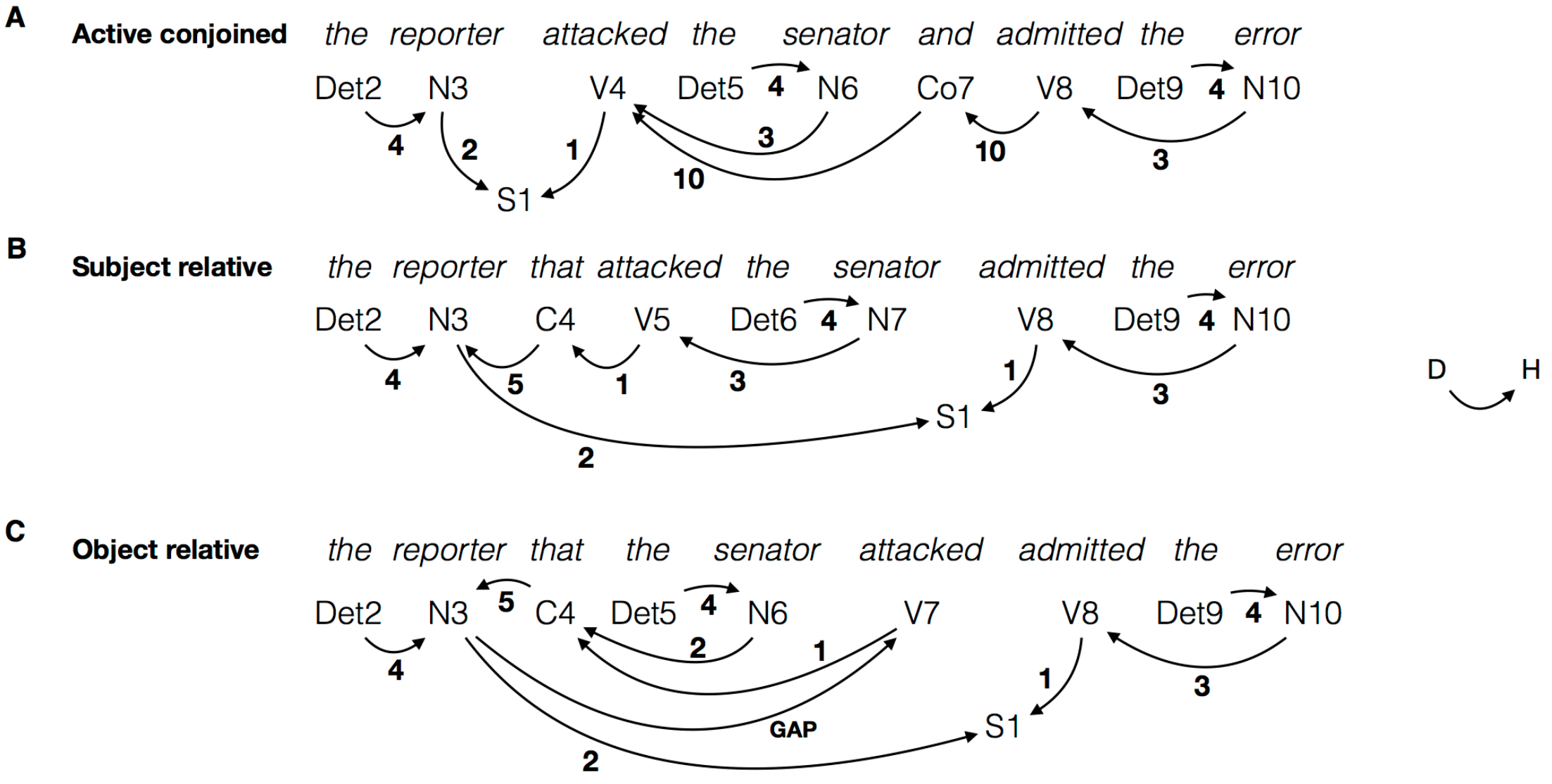

**Figure S20. Additional information for Figure 5A,B.** HD-NBA structure for the simulated sentences in Figure 5A. LMAs and SGs in Table S11. Dependents (D) point to heads (H).

Figure S20 gives the sentence structures of the sentence simulations based on Just et al. (1996), presented in Figures 5A and 5B.

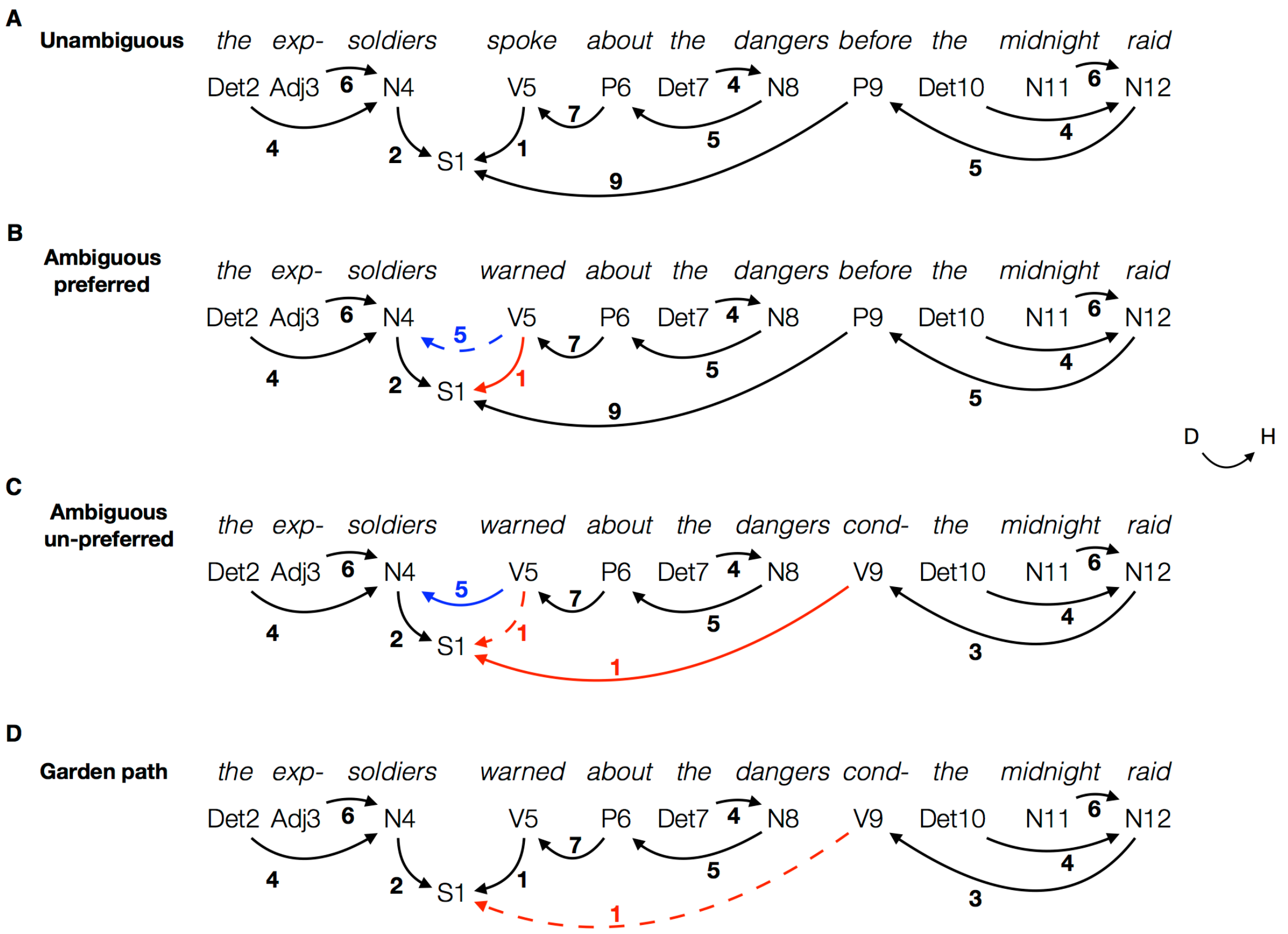

**Figure S21. Additional information for Figure 5C,D.** HD-NBA structure for the simulated sentences from Mason et al. (2003). *cond-* = *conducted. exp-* = *experienced.* The dashed connections represent conflicting bindings that are not realized. LMAs and SGs in Table S11. Dependents (D) point to heads (H).

Figure S21 gives the sentence structures of the sentence simulations based on Mason et al. (2003), presented in Figures 5C and 5D. Figure 5A present a (regular) binding of the unambiguous sentence. Figure 5B presents a conflict between the bindings of V5-*warned* and N4-*soldiers* with SG5-Compl and that of V5-*warned* and S1 with SG1-Verb. This conflict can be resolved when *before* appears, resulting in the binding of V5-*warned* and S1 (V5 is still the active verb LMA at this moment), as illustrated with the solid connection between V5 and S1. In Figure 5C, the conflict can be resolved when *conducted* appears, resulting in the binding of V5-*warned* and N4-*soldiers* with SG5-Compl, as illustrated with the solid connection between V5 and N4. This binding has to be achieved before the conflicting activation of V9 and binding of V9-*conducted* with S1.

The binding of V5 and S1 is the preferred solution of the ambiguity conflict in Figures 5B and 5C (Mason et al., 2003). Hence, it could proceed earlier of even directly. This is fine for the sentence in Figure 5B, but not for the sentence in Figure 5C. Here, this early binding of V5 and S1 will cause a processing breakdown known as a garden path. Figure S21D presents a garden path interpretation in which V5-*warned* binds directly with S1. As a result, the binding of V9-*conducted* remains unresolved, but the other bindings at the end of the sentence continue. The sum total activity of this garden path (GP) structure simulated with each HD-NBA architecture is presented in Figure 5D.

### S7.5. Effect size architecture on simulations

The simulations of the sentences presented in the main article illustrate the effect of the size of the architecture. Activity increases with increasing size of the architecture, except for the working memory populations (Figures 3C, 3D, 4C, S19C). Moreover, the increase of activation is stronger with more complex sentences (Figures 5B, 5D).

This increase results from the fact that, e.g., control signals for gates operate over the entire architecture, which generates increased activity with increased size of the architectures. This can be illustrated with the binding process in the sentence *I wonder how many miles I have fallen by this time* from Brennan et al. (2016), simulated with architecture W20.

Figure S22A shows the structure of this sentence, with an emphasis on the binding between P11-*by* and N13-*time*, with P11 as head and N13 as dependent. Figure S22B shows four cells in the connection matrix, based on the dependent rows D13 and D20 and the head rows H11 and H20.

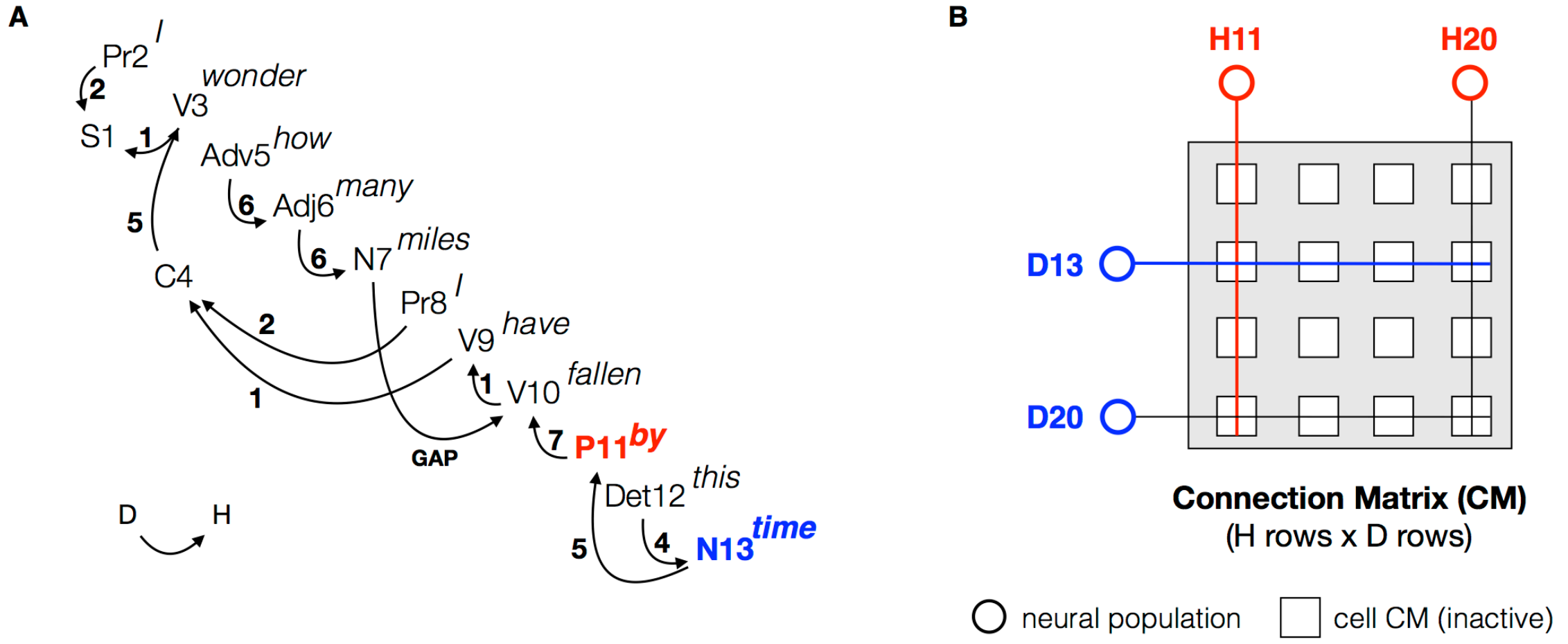

**Figure S22. Activation in different cells of the connection matrix**. (**A**) Structure of the simulated sentence from Brennan et al. (2016). (**B**) Four specific cells in the connection matrix. LMAs and SGs in Table S11. Dependents (D) point to heads (H). Red = head. Blue = dependent.

Row H11 is involved with P11-*by* and row D13 is involved with N13-*time*. So, the binding P11-N13 occurs in cell H11-D13. Rows H20 and D20, and with it cell H20-D20, are not involved in any binding. But row D20 is connected with row H11 in cell H11-D20, and row H20 is connected with row D13 in cell H20-D13. The difference between these four cells in the connection matrix is reflected in their activations, as given by Table S13.

Table S13 presents the activation of the populations Working memory, Start, Disinhibition, Inhibition and Gate (Figures S4A, S5A) involved in binding gates in each of the four cells in the connection matrix (Figure S6). The activations are averaged over the entire duration of the simulation (4500 ms).

The binding P11-N13 in cell H11-D13 is initiated by the concurrent activation of Start by input from H11 and Disinhibition by input from D13. This activates Gate and by it Working memory, which in turn activates Disinhibition. Gate is activated as long as Start is active. When the input from H11 stops, Start is no longer active. Hence, its average activation is lower than that of Working memory and Disinhibition. The activation of Gate also stops (after a decline period). The difference between Working memory and Disinhibition follows from Figures S15A and S15B. Working memory activation is reduced when there is no input, but Disinhibition continues to receive input from Working memory. The activation pattern in cell H11-D13 is characteristic of each binding in the connection matrix, but the average activation of Working memory (and Disinhibition) is higher for bindings that occur earlier in the sentence structure.

| Hrow-Drow: / Population in Binding Gate: | H11-D13 Role: Binding P11-N13 | H11-D20 Role: H active D inactive | H20-D13 Role: H inactive D active | H20-D20 Role: H inactive D inactive |
|---|---|---|---|---|
| Working memory | 12.6 | 0.005 | 0.005 | 0.005 |
| Start (Initiated in row H11) | 4.2 | 4.2 | 0.005 | 0.005 |
| Disinhibition (initiated in row D13) | 16.6 | 0.005 | 4.9 | 0.005 |
| Inhibition | 0.004 | 5.3 | 0.004 | 0.005 |
| Gate | 5.3 | 0.03 | 0.005 | 0.005 |

**Table S13. Activation in four cells of the connection matrix.**

The effect of global selection and binding control in the HD-NBA is illustrated here with binding related activity over the entire connection matrix. The cells H11-D20 and H20-D13 are not involved in the sentence structure. However, the input from H11 runs over the entire H11 row in the connection matrix. So, the binding process in cell H11-D13 also activates the Start population in H11-D20 (which activates Inhibition in this cell, because there is not binding). Similarly, the input from the D13 row also activates the Disinhibition population in cell H20-D13.

So, when the size of the connection matrix increases, the sum of binding related activity in it increases as well. The cell H20-D20 illustrates that this is even true for H and D rows that are not involved in any binding. The residue activation in this cell results from the fact that Wilson-Cowan populations do have a minimum activation, even without any input. So, even the sum activation of Working memory populations will grow slightly when the size of the architecture increases, even though the sentence structure remains the same.

### S7.6. Simulating question answering with content-addressable words

A way to retrieve information from the sentence neural blackboard is to ask questions. As argued in the main article, closed questions in particular provide information about the sentence structure queried. This can be used directly to activate that structure in the blackboard, due to the in-situ nature of the neural word structures involved. That is, words are always content-addressable in the HD-NBA, also when they are part of a sentence structure.

Figure S23 illustrates the role of content-addressable words in answering closed questions in the HD-NBA. For the question *Sue likes ?* three sentences are consecutively stored in the sentence blackboard (size W15). The question provides the subject and verb and asks for the object. The subject and verb of the first sentence are *Sue likes*, so it is the queried sentence, and the LMAs N2 and V3 will be activated by the question. However, *Sue* is also the subject of the second sentence, and *likes* is the verb of the third sentence. Hence, the LMAs N6 and V11, bound to these words, will also be activated by the question.

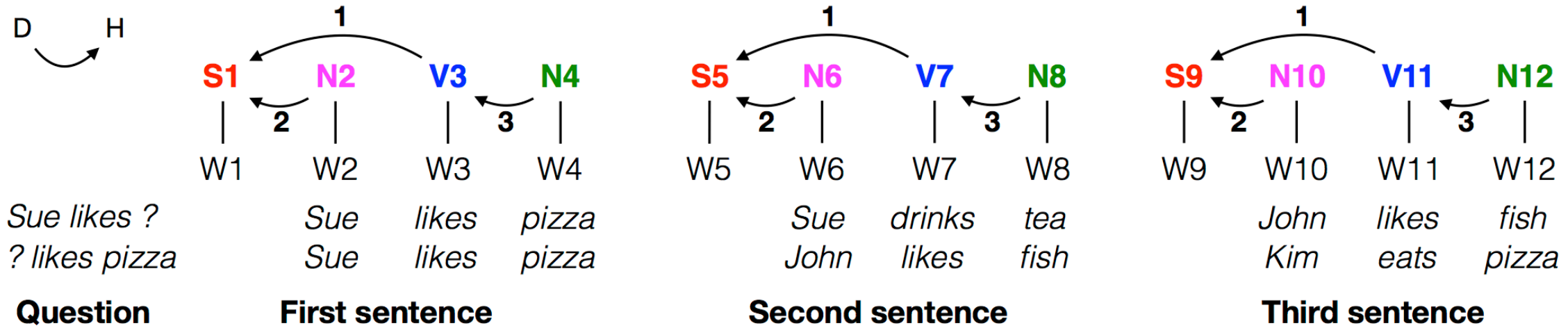

**Figure S23. Questions and related sentences**. LMAs and SGs in Table S11. Dependents (D) point to heads (H).

For the question *? likes pizza* three sentences are consecutively stored in the sentence blackboard as well (size W15). This question provides the verb and object and asks for the subject. The verb and object of the first sentence are *likes pizza*, so it is the queried sentence, and the LMAs V3 and N4 will be activated by the question. However, *likes* is also the verb of the second sentence, and *pizza* is the object of the third sentence. Hence, the LMAs V7 and N12, bound to these words, will also be activated by the question.

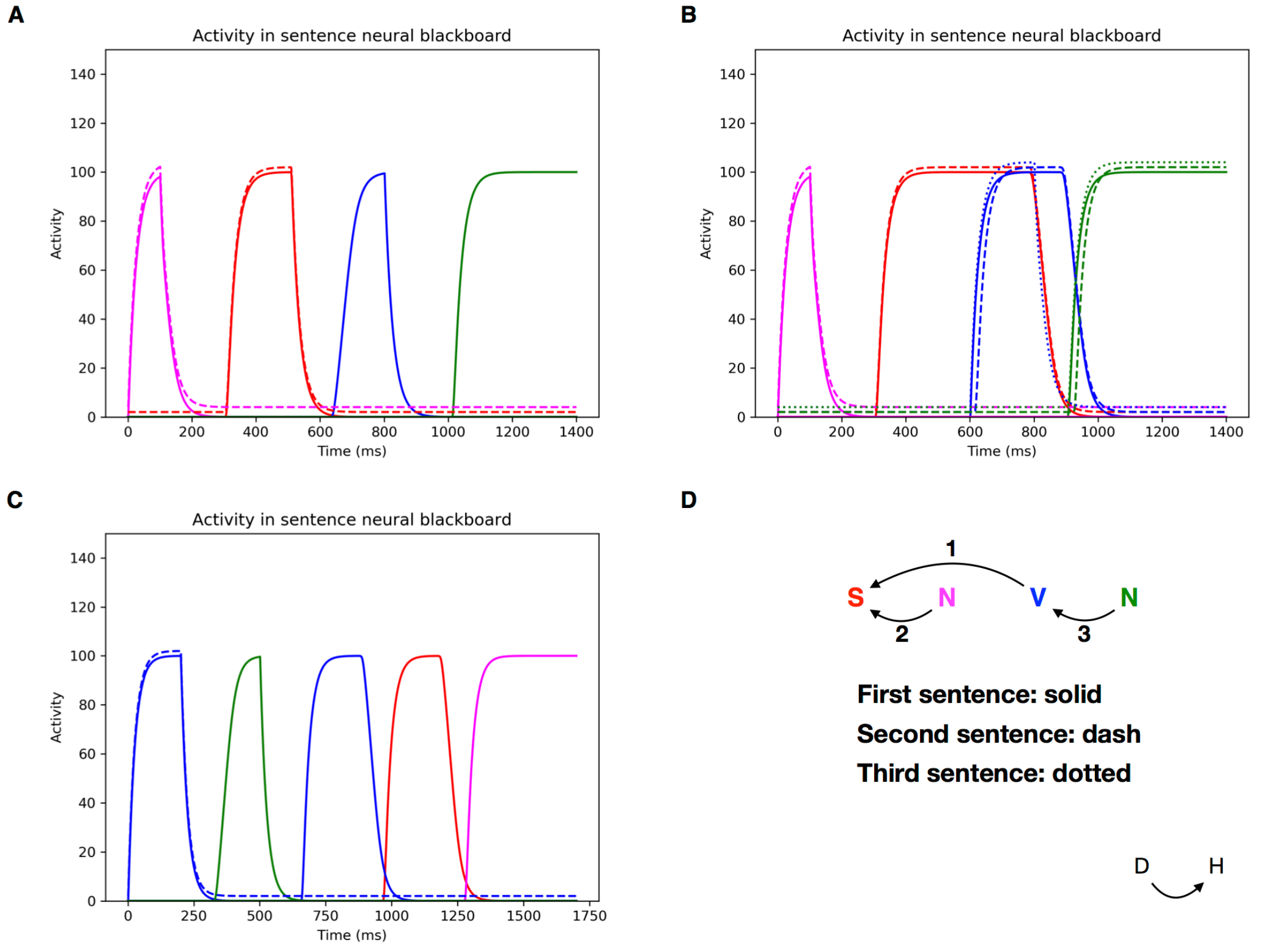

**Figure S24. Answering questions.** (**A**) Answer to the question *Sue likes?* based on LMA competition (**B**) Answer to the question *Sue likes?* without LMA competition. (**C**) Answer to the question *? likes pizza*. (**D**) Color and line type coding of the activations of the LMAs in the three sentences in Figure S23. The dashed and dotted activations are slightly increased artificially, to show the difference between the three sentences. LMAs are as in Figure S23.

Figure S24A schematically illustrates the process to answer the question *Sue likes ?*. First,

*Sue* activates N2 (via W2) and N6 (via W6). By opening the gate for subject, these activate S1 and S5. Then, *likes* would activate V3 and V11. However, by opening the gate for verb, S1 will also activate V3 and S5 will activate V7. So, all Verb LMAs would be activated in this way, but only V3 receives activation from both *likes* and the Sentence LMA. This difference can be used to solve the competition, by also inhibiting the Verb LMAs (as in sentence processing, LMAs of the same type can be inhibited simultaneously). The result is that V3 remains as the only active Verb LMA, because of its double activation. Then, by opening the gate for object, N4-*pizza* can be activated as the answer.

Figure S24B illustrates the answer to the question *Sue likes ?* when the competition between the Verb LMAs is not applied. Then, all Verb LMAs for the three sentences become activated, which results in three answers to the question.

Figure S24C illustrates the answer to the question *? likes pizza*. Here, *likes* activates V3 and V7 and *pizza* would activate N4 and N12. However, by opening the gate for object, V3 will also activate N4. The competition between LMAs of the same type described above will then result in the activation of N4 only, which in turn selects V3 as the only active Verb LMA. This will then activate S1, which in turn activates the answer N2-*Sue*.

## S8. Performance related to sentence complexity and ambiguity

This section provides a few illustrations of how performance issues related to sentence complexity and ambiguity could be dealt with in the HD-NBA.

In Figure 2, the index numbers of the LMAs run parallel to those of the word populations. This convention has been applied to all other figures up to now. However, this is rather cumbersome and often not very informative (binding with word populations is implied for all words in sentence structures). In this section, therefore, index numbers of LMAs start with 1 when an LMA of a specific type is first used, and are then numbered consecutively for LMAs of the same type.

### S8.1 Ambiguity resolution and backward processing

The combination of incremental and backward processing in the HD-NBA can be illustrated with the resolution of ambiguity in sentence processing.

An example is the resolution of an 'Un-Problematic Ambiguity' (or UPA) based on Kimball (1973) and Ferreira and Henderson (1990). UPAs appear in two related forms (a, b). They concern the transition between two related but different sentence structures, which are accepted without difficulty by language users. The UPA in question is listed as UPA1 in Lewis (1993), and has the following form:

UPA1a *I knew the man*
UPA1b *I knew the man hated me passionately*

In UPA1a, *man* is the object of *knew*. In UPA1b, *man* is the subject of the clause *the man hated me passionately*, which is the complement of *knew*.

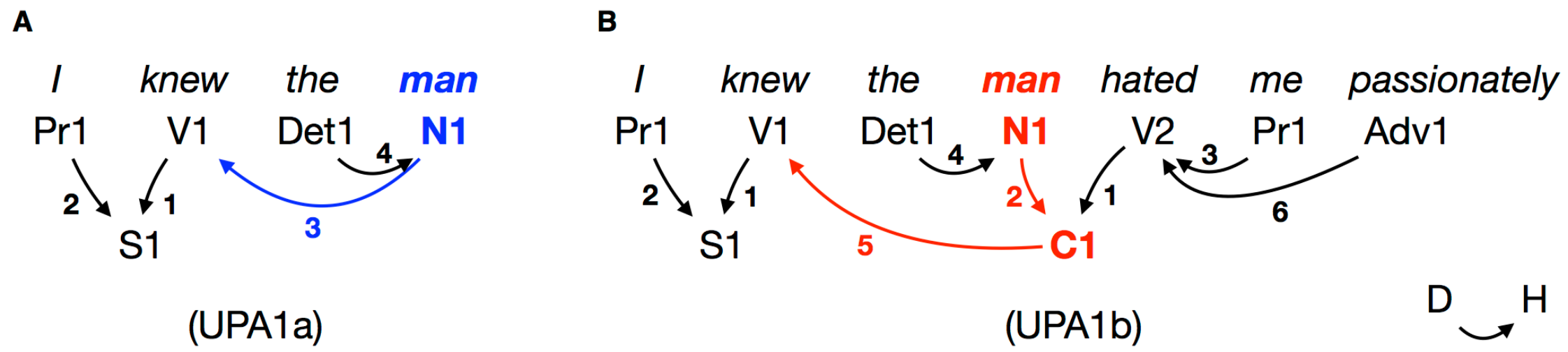

**Figure S25. Unproblematic ambiguity (UPA)**. (**A**) HD-NBA dependency structure UPA1a. (**B**) HD-NBA dependency structure UPA1b. LMAs and SGs in Table S11. Dependents (D) point to heads (H). Sentence labels based on Lewis (1993).

Figure S25 presents the HD-NBA dependency structures of both sentences. In the HD-NBA, the resolution of this ambiguity could be based on the possibility to activate multiple SGs in a head or dependent row and the competition between different SGs in the same dependent row (section 7.2). The verb *knew* could have an object or a complement clause, so SG3-Object and SG5-Compl could be activated in the head row of V1-*knew*. Similarly, in the dependent row N1-*man*, SG3-Object and SG2-Subject could be activated. However, no bindings are made yet in the connection matrix for any of these possibilities.

If the sentence stops after *man*, binding with SG3-Object will occur, as indicated with the blue connection in Figure S25A. If the sentence continues, N1-*man* will bind to C1 with SG2-Subject and C1 will bind to V1-*knew* with SG5-Compl, as indicated with the red connections in Figure S25B. The SG competition illustrated in Figure S18 will inhibit the other binding alternatives in the dependent rows. So, the role of the word *man* and the bindings of the verb *knew* in the sentence are determined by the phrase following *man* in the sentence. This is a form of backward processing without explicit backtracking (the words in the sentence are still processed in a forward manner).

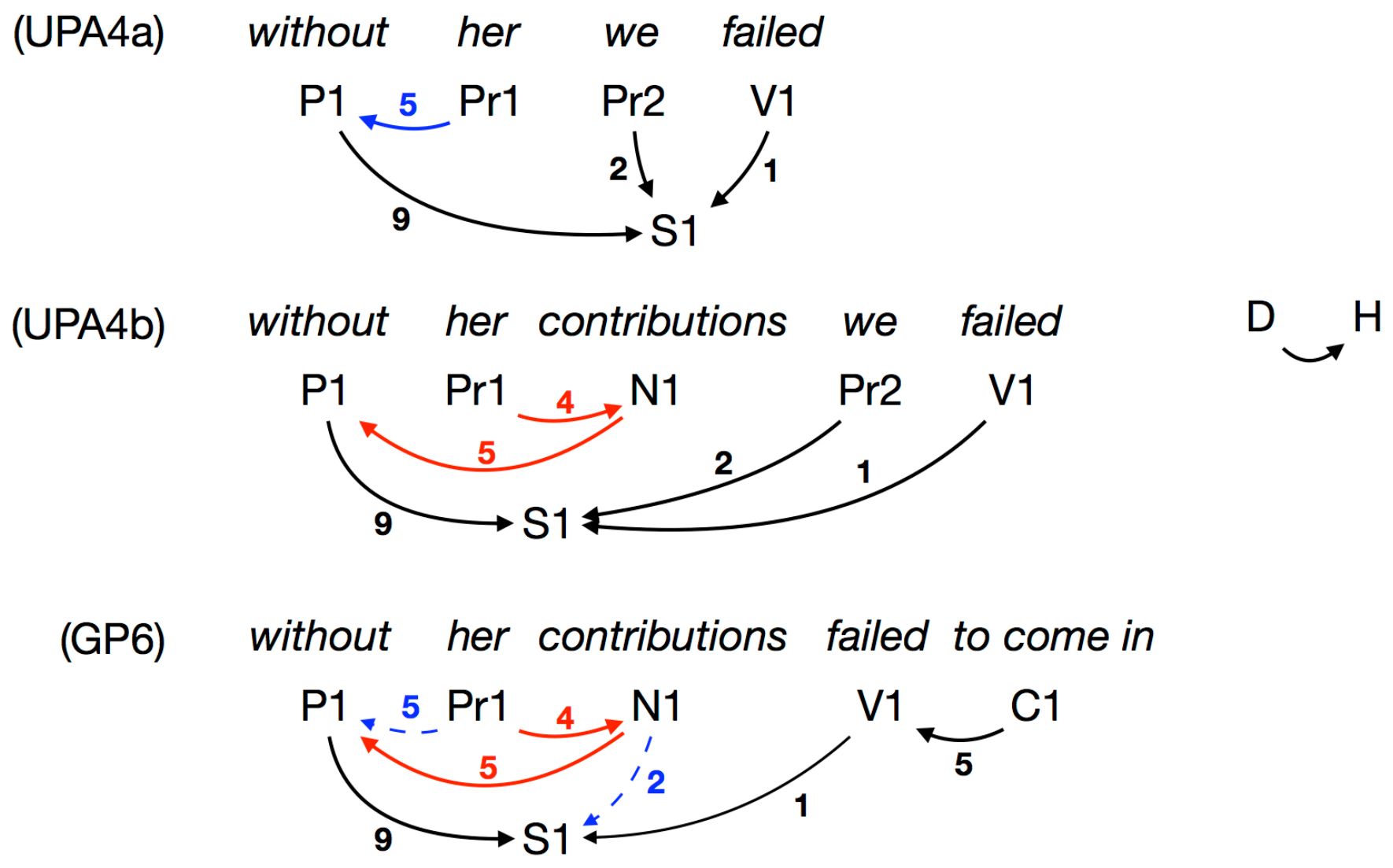

**Figure S26. Unproblematic ambiguity (UPA) and garden path (GP)**. The HD-NBA dependency structures of UPA4a, UPA4b and GP6. LMAs and SGs in Table S11. The dashed connections represent conflicting bindings that are not realized. Dependents (D) point to heads (H). Sentence labels based on Lewis (1993).

However, not all ambiguities are unproblematic. Some of them cause parsing difficulties know as garden paths (GPs). An illustration is given with the next triplet of sentences. The first two sentences form a UPA (Pritchett, 1988) and the third sentence is a GP (Frazier, 1978; Pritchett, 1988). They are labeled as UPA4 and GP6 in Lewis (1993):

| | |
|---|---|
| UPA4a | *Without her we failed*. |
| UPA4b | *Without her contributions we failed*. |
| GP6 | *Without her contributions failed to come in*. |

Figure S26 illustrates the HD-NBA structures of these sentences. The unproblematic ambiguities UPA4a and UPA4b could be solved in the same manner as UPA1a and UPA1b, illustrated in Figure S25. So, this results in the binding of *her* and *without* with SG5-Compl in UPA4a (blue connection), and the binding of *her* and *contributions* with SG4-Det in UPA4b (red connection).

But then, *her* and *contributions* would also bind with SG4-Det in GP6. Yet, in this sentence *her* should bind to *without* with SG5-Compl, as in UPA4a. However, due to the binding of *her* and *contributions* with SG4-Det, this binding is no longer possible, which results in the garden path character of this sentence (indicated with the blue dashed connections).

The garden path nature of a sentence could result from the moment at which a binding is made, as in Figure S21D. This can be further illustrated with the contrast between a classical GP (Bever, 1970) and a related UPA (Gibson, 1991, 1998; Pritchett, 1992), listed as GP14 and UPA13 in Lewis (1993):

| | |
|---|---|
| GP14 | *The horse raced past the barn fell.* |
| UPA13a | *The bird found in the room died.* |
| UPA13b | *The bird found in the room enough debris to build a nest.* |

Figure S27 shows the HD-NBA dependency structures of each sentence (ignoring *the*). For the UPA sentences, there are two options (up to the word *room*) for binding the word *found*. Both options can be kept open by activating the SGs involved (SG1-Verb and SG5-Compl) in the related head and dependent rows, without yet activating their bindings in the connection matrix. Then, a selection can be made after the word *room*, resulting in the binding of V1-*found* with SG5-Compl to N1-*Bird* in UPA13a (blue connection) and the binding of V1-*found* with SG1-Verb to S1 in UPA13b (red connection). The dashed connections in these sentences illustrate the competing bindings that are inhibited in the dependent rows. The binding of P1-N2 with SG5-Compl does not interfere with the (potential or later) binding of N1-V1 with SG5-Compl (see Figure S17).

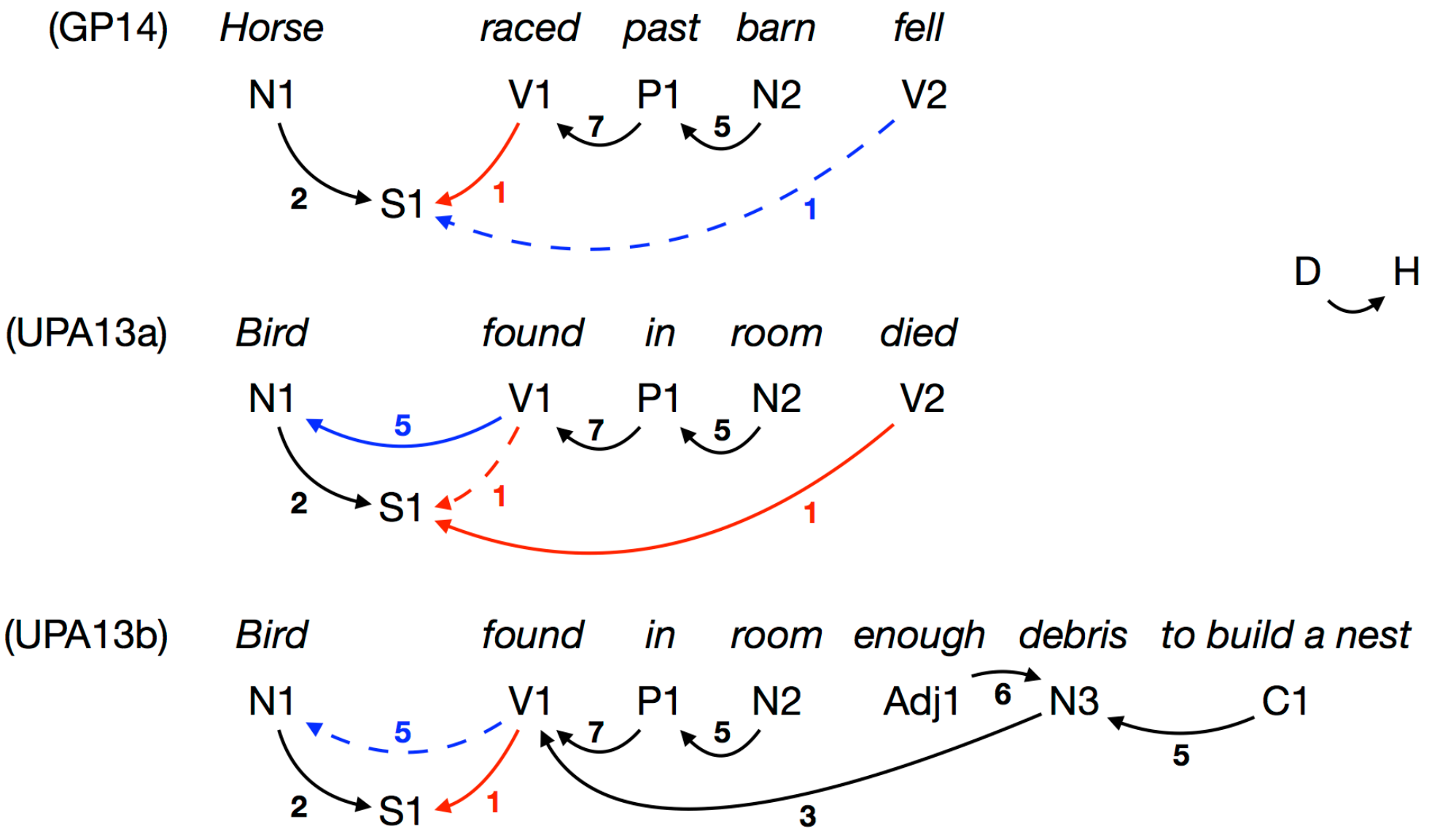

**Figure S27. Garden path (GP) versus unproblematic ambiguity (UPA).** The HD-NBA dependency structures of GP14, UPA13a and UPA13b. LMAs and SGs in Table S11. The dashed connections represent conflicting bindings that are not realized. Dependents (D) point to heads (H). Sentence labels based on Lewis (1993).

UPA13a and UPA13b again illustrate the possibility of backward processing without explicit backtracking in the HD-NBA. But they also raise the question of why that does not seem to occur with GP14, which is structurally similar to UPA13a. In the HD-NBA, the garden path character of GP14 occurs when V1-*raced* is bound directly as verb to S1 (red connection). In that case, the binding between V2-*fell* and S1 is blocked, as indicated with the dashed blue connection. The binding S1-V2 is blocked because the subassembly for SG1-Verb in the head row of S1 is inhibited due to the binding S1-V1 (see Figure S16).

Hence, in the framework of the HD-NBA, a difference as between GP14 and UPA13 results from the moment of making a binding choice, here for the first verb (V1). If the choice is made directly, a garden path occurs, as in GP14. If it is postponed (to obtain more information) a garden path could be avoided, as in UPA13. In turn, the moment of making the binding choice would be influenced by the semantic relations between the words involved, which illustrates the influence of semantics on the binding process in the HD-NBA.

Also, a direct binding reduces activity (hence energy), but only if the resulting structure is 'reasonably' certain. Otherwise, a wrong structure could follow, which would increase the energy required even more because of the re-analysis needed. This illustrates the interaction between dynamics and information processing in the HD-NBA, which clearly has an evolutionary basis (here, a trade-off between use of energy and correctness of analysis).

### S8.2. Multiple embeddings and performance

Performance effects in the HD-NBA will also occur when (more) bindings cannot be achieved directly but have to be kept in memory during sentences processing. Two examples are presented in Figure S28. The first one is a sentence listed as a parsing breakdown (PB), in particular PB1, in Lewis (1993). It is a center-embedded object-relative sentence (Miller and Chomsky, 1963; Miller and Isard, 1964). Sentences of this kind are indeed very hard to process (Fodor and Garrett, 1967). The second sentence is a post-verbal center-embedded object-relative sentence (Eady & Fodor, 1981), listed as an acceptable embedding (AE), in particular AE6, in Lewis (1993):

PB1 *The man that the woman that the dog bit likes eats fish.*
AE6 *I saw the man that the woman that the dog bit likes.*

Both sentences share the embedded object-relative clauses *the man that the woman that the dog bit likes.* In AE6, they follow the main verb *saw*, in PB1 they precede the main verb *eats.*

Accounts for parsing breakdowns typically assume that they result from limited computational resources, either as specific architectural limitations (e.g., Lewis, 1993) or in terms of a more abstract metric. For example, Gibson (1998) presented a theory in which parsing breakdown results from the number of 'open slots' that occur during processing. When that metric reaches a certain limit (5 open slots) a breakdown of parsing occurs.

Figure S28 shows the HD-NBA dependency structures of PB1 and AE6 (ignoring *the*). The (red) numbers and labels in the dependency connections indicate the bindings that cannot be achieved directly. Here, they concern the bindings with SG1-Verb and with SG11-GAP.

AE6 is an acceptable embedding in terms of processing in the HD-NBA. To achieve all bindings needed, SG11-GAP of N1 and N2 need to be pre-activated as dependent because N3 is activated before these bindings can occur. Likewise, SG1-Verb of C1 needs to be pre-activated as head because C2 is activated before this binding can occur. When two SGs of the same type are activated (in different rows), the competition is decided in favour of the last activated SG (Figure S17). When that SG has resulted in binding, the first SG can also produce a binding. In this way, N2 will first bind to V1 with SG11-GAP, and N1 will then bind to V2 with SG11-GAP. Similarly, V1 will first bind to C2 (as dependent) with SG1-Verb, and V2 will then bind to C1 (as dependent) with SG1-Verb.

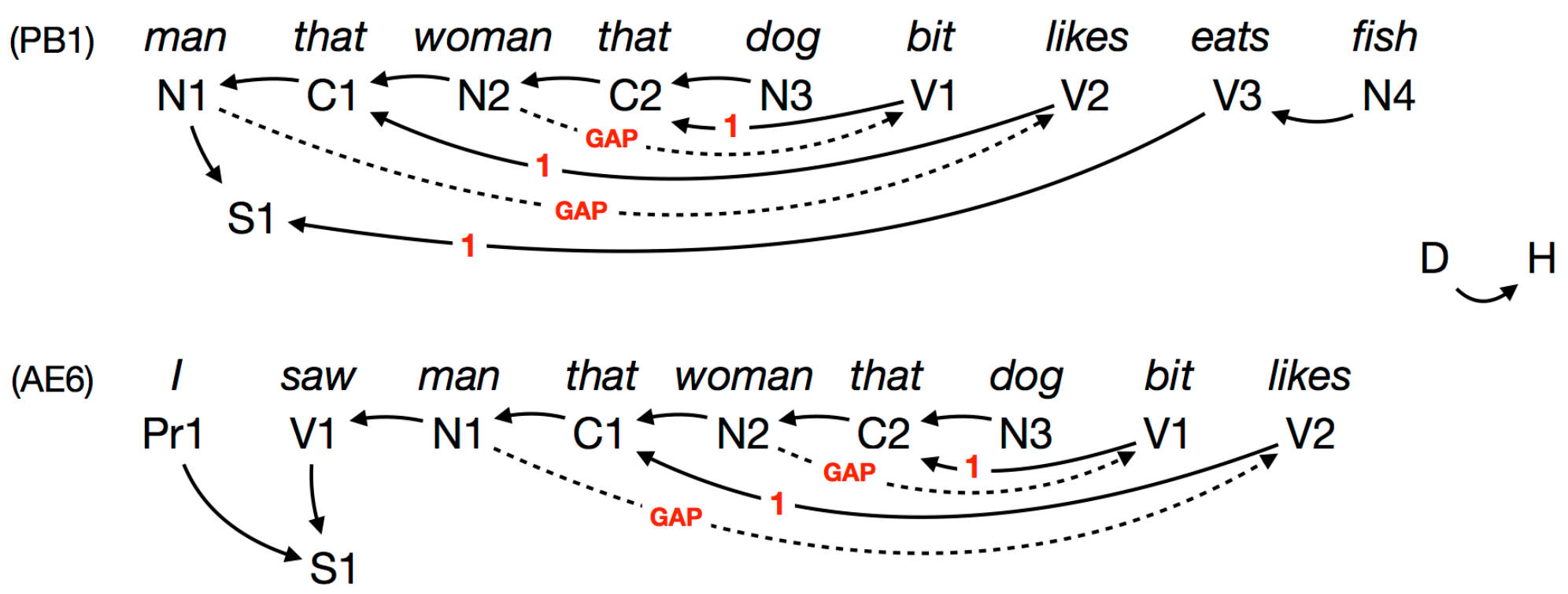

**Figure S28. Parsing breakdown (PB) and acceptable embedding (AE).** The HD-NBA dependency structures of PB1 and AE6. LMAs and SGs in Table S11. Dotted connections represent bindings with SG11-GAP. Dependents (D) point to heads (H). Sentence labels based on Lewis (1993).

In PB1, the binding of the main verb to S1 will intervene with the binding of the verbs as dependent with SG1-verb as outlined for AE6. This will occur specifically when SG1-Verb of S1 is pre-activated (as head). Then, there will be an active SG1-Verb in three head rows when the first verb appears. This one will bind with C2, based on the competition illustrated in Figure S17. But after that, there will be an active SG1-Verb in two head rows, causing binding problems for the other verbs. To eliminate this problem, the activation of SG1-Verb for S1 should be postponed until the last verb appears (being the verb of the main sentence). But this will put an additional burden on parsing control. So, the HD-NBA account of the processing differences between PB1 and AE6 is based on architectural constraints and matches the metric proposed by Gibson (1998).

However, there is another reason why processing of PB1 is more difficult than processing of AE6 in the HD-NBA. This concerns the requirement that only one word population is active in the intersection between the phonological and the sentence neural blackboards (representing the current word). The appearance of the next word in the sentence will initiate the activation of a new word population. So, the word population of the current word needs to bind with an LMA in the sentence neural blackboard before the next word initiates the activation of a new word population.

Moreover, when the next word is of the same type (e.g., verb) as the current word, SGs need to be selected in the head or dependent rows of the current word before the LMA of the next word is active. Otherwise, this LMA will inhibit the previous LMA of the same type, and the control and binding signals that initiate SG activation will operate for the LMA of the next word, instead of for LMA of the current word.

So, control and binding problems can arise when there is a limited time available in which the word population can bind to an LMA, and SGs can be selected in the head or dependent rows of that LMA. This occurs in PB1 for the verbs V1 to V3. First, V1 and V2 each require two bindings of the same kind (SG1-Verb and SG11-GAP), in which they have different roles (dependent and head, respectively). This requires that these bindings be activated sequentially, which takes time. Secondly, the activation of V2 and V3 require the inhibition of the previous Verb LMA, which also takes time.

Simulations, using Wilson-Cowan (1972) population dynamics, show that problems indeed arise for PB1 when there is a period of around 300 milliseconds available for the activation of a word population (in line with the processing of 3-4 words per second, Rayner & Clifton, 2009). In that case, the binding process of V2 'spills over' to V3, which results in erroneous bindings in the sentence structure when V3 is activated. This 'spill over' also occurs in AE6, but it has no effect in this sentence because V2-*likes* is the last word.

So, next to architectural and memory limitations (as given by a decline of sustained activation), the HD-NBA also offers another account for parsing problems, given by timing constraints on the binding process in the architecture. This would suggest that PB1 becomes more acceptable when pauses are introduced in the binding process, in particular after V2-*likes*.

## S9. Discussion

The HD-NBA is based on a number of core assumptions (S2.1). One of them is the notion of content-addressability, which was one of the key factors in the development of connectionism (Bechtel and Abrahamsen, 1991). The HD-NBA combines content-addressability of words (concepts) in sentence structures with combinatorial productivity. This sets the HD-NBA apart from other architectures of sentence representation and processing, both symbolic and connectionist (neural) ones.

In the HD-NBA, words (in-situ concept structures) are accessible in sentence structures (connection paths). The accessibility of 'constituents' (e.g., words) in structures such as sentences is often seen as a defining characteristic of symbol manipulation (e.g., Fodor and Pylyshyn, 1988; Newell, 1990). However, this does entail that the HD-NBA is just an implementation of a symbol manipulating architecture based on a form of digital computing, as described by Fodor and Pylyshyn (1988) and Newell (1990). The HD-NBA does not rely on neural 'codes' (implementing symbols) and neural 'registers' (implementing memory locations to store these codes). Instead, the HD-NBA is based on a form of 'in-situ' computing, which relies on creating connection paths in a 'fixed' connection structure. This form of computing is Turing equivalent (van der Velde, 1997), but it does not depend on neural 'codes' or 'registers'.

In the HD-NBA, words (constituents) are accessible in sentence structures because they are content-addressable. This is not the case in symbol manipulation architectures. Here, the symbol is used as a token to obtain more information elsewhere in the architecture when needed (Newell, 1990). But when a symbol is stored in an arbitrary register, located somewhere in the architecture, it is not directly accessed or activated when a copy of that symbol is activated somewhere else in the architecture. To access or retrieve a symbol, a deliberate search process has to be initiated based on the identity of the symbol or the address code of the register in which it is stored. Similarly, when new information related to a symbol (e.g., the object it stands for) emerges, it needs to be linked deliberately to the existing information available about the symbol, again using symbol identity or address code.

In this way, the HD-NBA is also different from neural accounts of combinatorial (sentence) processing that do store neural codes representing words or address codes in dedicated registers (Kriete et al., 2013; Müller et al., 2020). Similarly, the HD-NBA is different from neural (connectionist) accounts in which sentences are encoded in such a way that its constituents (e.g., words, clauses) are not directly identifiable, and hence not content-addressable, in the sentence structure. Examples are tensor forms of binding based on the vector representations of words (e.g., Smolensky, 1990) or reduced vector representations of sentences (e.g., Plate, 1995; Eliasmith, 2015).

The difference between the HD-NBA and these architectures is again found in the role of content-addressability, which is lacking in these architectures. As discussed in the main article, content-addressability would have had major benefits over more indirect and hence slower and more error-prone approaches in retrieving information from sentence structures.

Viewed in this way, 'in-situ' architectures such as the HD-NBA are more related to motor control, in which the sequential activation of muscles (which are in-situ by construction) is controlled by connections from (potentially distant) processing networks, not by transporting information to them.

In its current version, the HD-NBA is hand-designed. Future research would have to account for the way an architecture like the HD-NBA could develop in the brain, presumably based a a combination of genetic and learning influences.

One could perhaps argue that it is better to start with learning architectures straight away. However, many learning architectures are also hand-designed. They learn the weights of their connections, but not the structure of the model itself. Moreover, as research on language learning with chimpanzees and bonobos suggests, even highly sophisticated network structures like the brains of these primates are not capable of learning human language in its full potential. So, just starting with architectures without knowing their capabilities beforehand could entail that, even after some initial successes, the architectures would not be able to learn human language in full.

Hence, a combination of approaches could be a better solution, in which architectures are developed with full language capabilities and learning methods are developed that could show how such architectures could develop from simpler versions, based by learning and development. This is the approach advocated here.

## S10. References

Amit, D. (1989). *Modeling Brain Function*. Cambridge, MA: Cambridge University Press.

Andor, D., Alberti, C., Weiss, D., Severyn, A., Presta, A., Ganchev, K., Petrov, S., & Collins, M. (2016). Globally normalized transition-based neural networks. *arXiv: 1603.06042v2*.

Baars, B. J. (1988). *A Cognitive Theory of Consciousness*. Cambridge University Press, Cambridge.

Bastos, A. M., Loonis, R., Kornblith, S., Lundqvist, M. and Miller E. K. (2018). Laminar recordings in frontal cortex suggest distinct layers for maintenance and control of working memory. *PNAS, 115(5),* 1117–1122.

Bechtel, W. & Abrahamsen, A. (1991). *Connectionism and the Mind*. Cambridge, MA: Blackwell.

Bever, T. G. (1970). The cognitive basis for linguistic structures. In Hayes, J. R. (ed.), *Cognition and the Development of Language*. New York, Wiley.

Bloom, P. (2000). *How children learn the meaning of words*. Cambridge, MA: MIT Press.

Brennan, J. R., Stabler, E. P., Van Wagenen, S. E., Luh, W.-M., and Hale, J. T. (2016). Abstract linguistic structure correlates with temporal activity during naturalistic comprehension. *Brain and Language, 157-158*, 81-94.

Culicover, P.W. & Jackendoff, R. S. (2005). *Simpler Syntax*. Oxford: Oxford University Press.

Dehaene, S., Meynielm F., Wacongne, C., Wang, L., & Pallier, C. (2015). The neural representation of sequences: From transition probabilities to algebraic patterns and linguistic trees. *Neuron, 88*, 2-19.

Doupe, A.J. & Kuhl, P.K. (1999). Birdsong and human speech. *Annual Review Neuroscience, 22,* 567–631.

Eady, J. & Fodor, J. D. (1981). *Is center embedding a source of processing difficulty?* Presented at the Linguistic Society of America Annual Meeting (cited in Lewis, 1993).

Eliasmith, C. (2015). *How to Build a Brain: A Neural Architecture for Biological Cognition.* Oxford: Oxford University Press.

Feldman, J. (2013). The neural binding problem(s). *Cognitive Neurodynamics. 7*, 1–11.

Ferreira, F. & Henderson, J. M. (1990). Use of verb information in syntactic parsing: Evidence from eye movements and word-by-word self-paced reading. Journal of Experimental Psychology: *Learning, Memory, and Cognition, 16*, 555–568.

Fodor, J. A. & Garrett, M. (1967). Some syntactic determinants of sentential complexity. *Perception and Psychophysics, 2(7),* 289–296.

Fodor, J. A. & Pylyshyn, Z. W. (1988). Connectionism and cognitive architecture: A critical analysis. *Cognition, 28*, 3-71.

Frazier, L. (1978). *On comprehending sentences: Syntactic parsing strategies*. PhD thesis, University of Connecticut (cited in Lewis, 1993).

Gibson, E. (1991). *A Computational Theory of Human Linguistic Processing: Memory Limitations and Processing Breakdown*. PhD thesis, Carnegie Mellon.

Gibson, E. (1998). Linguistic complexity: locality of syntactic dependencies. *Cognition, 68*, 1-76.

Hadley, R. F. (2006). Neural circuits, matrices and conjunctive binding. *Behavioral and Brain Sciences, 29*, 80.

Huddleston, R. & Pullum, G. K. (2002). *The Cambridge Grammar of the English Language*, Cambridge: Cambridge University Press.

Huth, A. G., de Heer, W. A., Griffiths, T. L., Theunissen, F. E., & Gallant, J. L. (2016). Natural speech reveals the semantic maps that tile human cerebral cortex. *Nature, 532 (7600),* 453-458.

Jackendoff, R. (2002). *Foundations of Language*. Oxford: Oxford University Press.

James, W. (1890, 1950). *The Principles of Psychology*. Volume I and II. Dover publications,

USA.
Just, M. A., & Carpenter, P. A. (1992). A capacity theory of comprehension: Individual differences in working memory. *Psychological Review, 99(1),* 122–149
Just, M. A., Carpenter, P. A., Keller, T. A., Eddy, W. F. & Thulborn, K. R. (1996). Brain activation modulated by sentence comprehension. *Science, 274*, 114-116.
Kimball, J. (1973). Seven principles of surface structure parsing in natural language. *Cognition, 2*, 15–47.
Lambon-Ralph, M. A., Jefferies, E., Patterson, K. & Timothy T. Rogers, T. T. (2017). The neural and computational bases of semantic cognition. *Nature Reviews Neuroscience 18*, 42–55.
Lewis, R. L. (1993). *An Architecturally-based Theory of Human Sentence Comprehension.* Thesis Carnegie Mellon University, Pittsburgh, PA.
Kriete, T., Noelle, D. C., Cohen, J. D., & O'Reilly, R. C. (2013). Indirection and symbol-like processing in the prefrontal cortex and basal ganglia. *PNAS, 110 (41)* 16390-16395.
Letzkus, J. J., Wolff, S. B. E., & Lüthi, A. (2015). Disinhibition, a Circuit Mechanism for Associative Learning and Memory, *Neuron, 88*, 264-276.
Marslen-Wilson, W. (1973). Linguistic structure and speech shadowing at very short latencies. *Nature, 244*, 522–523.
Mason, R.A., Just, M.A., Keller, T. A. & Carpenter, P.A. (2003). Ambiguity in the brain: what brain imaging reveals about the processing of syntactically ambiguous sentences. *Journal of Experimental Psychology: Learning, Memory and Cognition, 29(6),* 1319-38.
McClelland, J. L. & Rogers, T. T. (2003). The parallel distributed processing approach to semantic cognition. *Nature Reviews Neuroscience, 4*, 310-322.
Miller, G. A. & Chomsky, N. (1963). Finitary models of language users. In Luce, D. R., Bush, R. R. and Galanter, E. (eds.). *Handbook of Mathematical Psychology, volume II.* New York, John Wiley.
Miller, G. A. & Isard, S. (1964). Free recall of self-embedded English sentences. *Information and Control, 7*, 292–303.
Misner, C. W., Thorne, K. S. & Wheeler, J. A. (1973) *Gravitation.* Freeman, San Francisco.
Müller, M. G., Papadimitriou, C. H., Maass, W., & Legenstein, R. (2020). A Model for Structured Information Representation in Neural Networks of the Brain. *eNeuro. 29,* 7(3).
Nelson, M. J., El Karouib, I., Giberc, K., Yang, X., Cohen, L., Koopman, H., Cashc, S. S., Naccache, L., Haleg, J. T., Pallier, C., & Dehaene, S. (2017). Neurophysiological dynamics of phrase-structure building during sentence processing, *PNAS, E3669–E3678*.
Newell, A. (1990). *Unified Theories of Cognition*. Cambridge, MA: Harvard University Press.
Nivre, J. (2008). Algorithms for deterministic incremental dependency parsing. *Computational Linguistics, 34 (4),* 513-553.
Plate, T. A. (1995). Holographic reduced representations. *IEEE Transactions Neural Networks, 6*, 623-641.
Pritchett, B. L. (1988). Garden path phenomena and the grammatical basis of language processing. *Language, 64*, 539–576.
Pritchett, B. L. (1992). *Grammatical Competence and Parsing Performance*. Chicago: University of Chicago Press.
Pylkkänen, L. (2019). The neural basis of combinatory syntax and semantics. *Science, 366(6461),* 62-66.
Rayner, K. and Clifton, C. J. (2009). Language processing in reading and speech perception is fast and incremental: Implications for Event Related Potential research. *Biological Psychology, 80*, 4–9.
Shanahan. M. (2010). *Embodiment and the Inner Life*. Oxford: Oxford University Press.
Smolensky, P. (1990). Tensor product variable binding and the representation of symbolic structures in connectionist systems. *Artificial Intelligence, 46,* 159-216.

Tanenhaus, M. K., Spivey-Knowlton, M. J., Eberhard, K. M., and Sedivy, J. C. (1995). Integration of visual and linguistic information in spoken language comprehension. *Science, 268 (5217),* 1632–1634.

Trask, R. L. (1993). *A dictionary of grammatical terms in linguistics*. London: Routledge.

Van der Velde, F. (1997). On the use of computation in modelling behaviour. *Network: Computation in Neural Systems, 8*, 1-32.

Van der Velde, F. (2015a). Communication, concepts and grounding. *Neural Networks, 62*, 112-117.

Van der Velde, F. (2015b). Computation and dissipative dynamical systems in neural networks for classification. *Pattern Recognition Letters, 64*, 44-52.

Van der Velde, F. & de Kamps, M. (2006). Neural blackboard architectures of combinatorial structures in cognition (target article). *Behavioral and Brain Sciences, 29*, 37-70.

Van der Velde, F & de Kamps, M. (2010). Learning of control in a neural architecture of grounded language processing. *Cognitive Systems Research, 11*, 93–107.

Van der Velde, F & de Kamps, M. (2011). Compositional connectionist structures based on in situ grounded representations. *Connection Science, 23*, 97–107.

Van der Velde, F. & de Kamps, M (2015). The necessity of connection structures in neural models of variable binding. *Cognitive Neurodynamics, 9,* 359–370.

Van Valin, R. (2001). *An introduction to syntax*. Cambridge, UK: Cambridge University Press.

Watts D. J. & Strogatz, S. H, (1998). Collective dynamics of 'small-world' networks. *Nature, 393*, 440–442.

Wiggins, G. A. (2012). The mind's chorus: creativity before consciousness. *Cognitive Computation. 4(3),* 306–319.

Wilson, H. R., Cowan, J. D. (1972). Excitatory and inhibitory interactions in localized populations of model neurons. *Biophysical Journal, 12*, 1–24.

Wilson, H. R., Cowan, J. D. (2021). Evolution of the Wilson-Cowan equations. *Biological Cybernetics, 115(6),* 643-653.

## S11. List of abbreviations

*Literature:*

| | |
|---|---|
| Cambridge Grammar | Huddleston and Pullum (2002). |
| CG | Cambridge Grammar. |

*General:*

| | |
|---|---|
| AE | Acceptable Embedding. |
| D | Dependent |
| GP | Garden Path. |
| H | Head |
| HD-NBA | Head-Dependent NBA |
| LMA | Lexical Main Assembly. |
| NBA | Neural Blackboard Architecture. |
| PB | Parsing Breakdown. |
| SA | Subassembly. |
| SG | Structure Group. |
| UPA | Un-Problematic Ambiguity. |
| W | Word population. |
| WM | Working Memory. |

*Lexical Main Assembly types:*

| | |
|---|---|
| Adj | Adjective. |
| Adv | Adverb. |
| C | Clause. |
| Co | Coordinator |
| Det | Determiner |
| N | Noun. |
| P | Preposition. |
| Pr | Pronoun. |
| S | Sentence. |
| V | Verb. |

*Structure Group:*

| | |
|---|---|
| Compl | Complement. |
| Coord | Coordination. |
| Det | Determiner. |
| ExMod | External Modifier |
| Mod | Modifier. |
| PC | Predicate Complement. |